\documentclass{article}

\PassOptionsToPackage{numbers, compress}{natbib}
 \usepackage[preprint]{neurips_2026}

\usepackage[utf8]{inputenc}             
\usepackage[T1]{fontenc}                
\usepackage{hyperref}                   
\usepackage{url}                        
\usepackage{booktabs}                   
\usepackage{amsfonts}                   
\usepackage{nicefrac}                   
\usepackage{microtype}                  
\usepackage[dvipsnames]{xcolor}         
\usepackage{amsmath}                    
\usepackage{xcolor}
\usepackage{graphicx}                   
\usepackage{subcaption}
\usepackage{float}                      

\usepackage{multirow}                   
\usepackage{booktabs}                   

\usepackage{soul}                       

\usepackage[inline]{enumitem}

\newcommand{\olmoThree}{OLMo3}
\newcommand{\olmoTwo}{OLMo2}

\title{Tracing Morphology, Semantics, and Divergence in Language Acquisition}

\author{%
  Forough Shirin Abkenar, Haoshuo Fu, Liang Feng, Ning Yan, Masood S. Mortazavi\\
  Futurewei Technologies, USA\\
  \texttt{\{fshirin, hfu, lfeng, yan.ningyan, masood.mortazavi\}@futurewei.com}
}

\begin{document}

\maketitle


\begin{abstract}
  Understanding the dynamics of language acquisition in Large Language Models (LLMs) is crucial for uncovering how statistical representations evolve into structured linguistic capabilities. However, most existing studies focus primarily on fully trained models, leaving the developmental trajectory across intermediate training stages largely underexplored. In this work, we leverage intermediate checkpoints from open-source models (e.g., Pythia and OLMo) under teacher-forcing evaluation to systematically investigate language acquisition across training steps and model scales. We focus on four core characteristics: morphological structure acquisition, embedding drift, repetition dominance, and co-occurrence effects. Our findings reveal that models acquire morphological structures well before semantic embeddings stabilize. Furthermore, we show that training dataset size has a positive impact on divergence recognition, while co-occurrence during pre-training and repetition during inference have a negative impact on divergence recognition and associative link generation.
\end{abstract}


\section{Introduction}\label{sec:intro}
Linguists and cognitive scientists have developed multiple explanations and theories for ``Plato's Problem''---how human beings come to acquire and then creatively use a native language (and other cognitive faculties) having been exposed to minimal experience at an early age~\citep{vygotsky1962thought, chomsky1986knowledge, ogrady2005how, evans_levinson_2009}. 
Language acquisition includes learning morphological structures and semantic representations \citep{fodor1988, llms_vs_human, morphology_generalization, katz-fodor, framenet2003, turney2010frequency, mikolov2013word2vec, pennington2014glove}. Statistical induction of morphological inflections helps LLMs handle syntactic requirements, referential agreements,  complex vocabularies, meaning generalization (to rare or invented words), potential translation or grammar errors, and semantic similarity of words \citep{impact, peng2025leveraging}.

Unlike humans, who acquire linguistic skills from real-life experience, large language models (LLMs) learn to generate text through statistical-learning (e.g., next-token prediction) on massive text corpora. However, the evolution of linguistic capabilities across time (training steps) and scale (model size) remains underexplored. To investigate the structure of language acquisition by LLMs, we study and analyze four underlying characteristics:  
\begin{enumerate*}[label=(\roman*)]
\item
morphological inflection \cite{morphology};
\item
embedding drift \citep{tracking_feature_dynamics,weight_tying};
\item 
recall and repetition \cite{JohnHopkins};
and \item 
co-occurrence \cite{smi}. 
\end{enumerate*}

We explore how these characteristics influence two linguistic creativity tests--Remote Associates Test (RAT) \citep{rat} and a variation on Divergent Association Task (DAT) \citep{dat_main}. Our modified DAT  measures the competence to recognize semantic divergence. RAT measures the competence to recall a single linking word that associates with each member of a set of diverging ones. Although there may be disagreements about how to define or measure linguistic and other forms of creativity \citep{ghiselin1985creative}, we investigate related and prerequisite competencies and how they evolve during language acquisition:

\begin{itemize}
    \item We evaluate the co-evolution of morphological structures and embedding drift for Pythia suite \citep{pythia} and OLMo3 \citep{olmo3} \footnote{To save space, similar results for GPT-2 \citep{gpt2_git, gpt2_checkpoints} and OLMo2 \cite{olmo2} have been placed in App. \ref{sec:appx_morph}-\ref{sec:appx_assoc}.}. To our knowledge, the \textit{evolution} of morphological competence across time and scale has not been studied earlier.
    \item We show that, in the open-source language models we have studied, the onset and maturity of morphological competence occur before the onset and maturity of semantic competence. 
    \item We investigate the evolution of linguistic creativity during training based on RAT (for link generation) and our modified version of DAT (for divergence recognition). To date, no previous studies have explored the evolution of creativity measures or how the acquisition of other linguistic competencies or characteristics might impact them.
    \item We also discover that although larger volumes of language-acquisition data can positively impact divergence recognition, repetition and co-occurrence have a negative impact on both of our creativity measures. 
\end{itemize}


\section{Related Work}\label{sec:intro_litrvu}

\subsection{Morphology Acquisition}\label{sec:intro_litrvu_morph}
Morphemes are the smallest semantic units in any natural language. Their inflections produce colors of added meaning in sentential contexts. \citet{impact} investigated morphological induction competence in LLMs with rich morphologically structures focusing on tense, gender, number, and case.
\citet{morphology_generalization} showed that LLMs struggle with these abilities (more so in agglutinative languages) because they rely heavily on memorized lexical associations. The authors in \citep{morphBPE} and \citep{SKMT} showed Byte Pair Encoding (BPE) \cite{bpe} disregards meaningful morpheme boundaries resulting in suboptimal and inconsistent segmentation, ambiguity, and disrupted semantic coherence (more so in morphologically rich languages). BPE  is a fundamental tokenizer, widely used in different foundation models, such as Pythia \citep{pythia} and OLMo \cite{olmo3,olmo2}.

\subsection{Semantic Acquisition}\label{sec:intro_litrvu_embed}
\citet{peng2025leveraging} investigate LLMs' vector embeddings for semantic similarity computation and analysis, observing that stronger LLMs display better performance in semantic similarity tasks.
\citet{tracking_feature_dynamics} observed three distinct  stages in semantic feature evolution: \begin{enumerate*}[label=(\roman*)]
\item Initialization and Warmup Phase---behaving as random noise with  activations lacking clear semantic structure; 
\item Emergent Phase---experiencing  ``geometric convergence'' where similar data points cluster together into distinct regions, and gaining recognizable semantic meaning (e.g., separating verbs from nouns); 
and \item Convergent Phase---entering a consistent period wherein semantic meaning does not change rapidly and the internal representation reaches its mature structure.
\end{enumerate*}

\subsection{Repetition and Co-occurrence}\label{sec:intro_litrvu_rep}
\citet{JohnHopkins} underscored the effect of repetition and short-term memory (STM) across time and scale. Employing verbatim in-context retrieval as a measure of artificial STM, they demonstrated that transformers achieve STM early in their training, and that STM  correlates positively with an array of LLM performance metrics. Token co-occurrence reinforces the link between two tokens based on their contextual co-occurrences. Observing that larger LLMs are more capable of learning more complex token patterns than smaller ones, \citet{token_cooccurrence} explored token reinforcement as models seek to maximize next-token likelihoods.


\section{Model Selection}\label{sec:models}
We present our findings for LLMs  with publicly available language-training checkpoints\footnote{Due to space limitations, evaluations for other models, i.e., \olmoTwo\ and GPT-2, can be found in Appendix \ref{sec:appx_extend}, and extra information about the models' architecture and training checkpoints are available in Appendices \ref{sec:appx_arch} and \ref{sec:appx_checkpoint}.}:
\begin{enumerate*}[label=(\roman*)]
    \item Pythia \citep{pythia} is a suite of decoder-only language models ranging from 14M to 12B parameters, including 154 intermediate checkpoints. The models were trained on the Pile dataset \citep{pile}, a 300B-token English-only corpus containing texts from 22 sources. We evaluate \{70M, 160M, 410M, 1B, 6.9B, 12B\} model sizes at 27 training checkpoints \footnote{To save space, we provide results for 70M and 12B. Results for other scales are available in Appendix.}. All checkpoints were accessed through the HuggingFace Transformers library \citep{transformers}; and
    \item Open Language Model (OLMo)--version 3 (\olmoThree) \citep{olmo3} comes in two sizes (7B and 32B) and four core variants (Base, Instruct, Think, and RL Zero). We evaluate OLMO3-7B-Base
\end{enumerate*}

trained on the 5.9T-token Dolma-3 dataset (mix of web text and academic data) through three stages---language acquisition, reasoning acquisition, and long-context extensions.


\section{Tracing Morphological Preferences}\label{sec:morphology}
Among common lexical categories\citep{essential_of_linguistics}---nouns, verbs, adjectives, adverbs, determiners, auxiliary verbs, prepositions, conjunctions, pronouns, and interjections---we focus on three: 
\begin{enumerate*}[label=(\roman*)]
\item Verbs: tense inflections and modal auxiliaries; 
\item Nouns: derivational affixes, singular and plural forms; and
\item Pronouns: possessive and gender agreements. 
\end{enumerate*}
For each of these categories, we design a number of unit tests\footnote{Designing a comprehensive test for LLMs to learning the morphological structures is beyond the scope of this paper. The simple test here covers the defined lexical categories. Also, our teacher-forcing setup cannot be employed to investigate morphological learning profiles of certain lexical categories, such as personal pronouns, as they start the incipit prompts.}. 

\begin{figure}[h]
    \centering
    \begin{subfigure}{.45\textwidth}
          \centering
          \includegraphics[width=\linewidth]{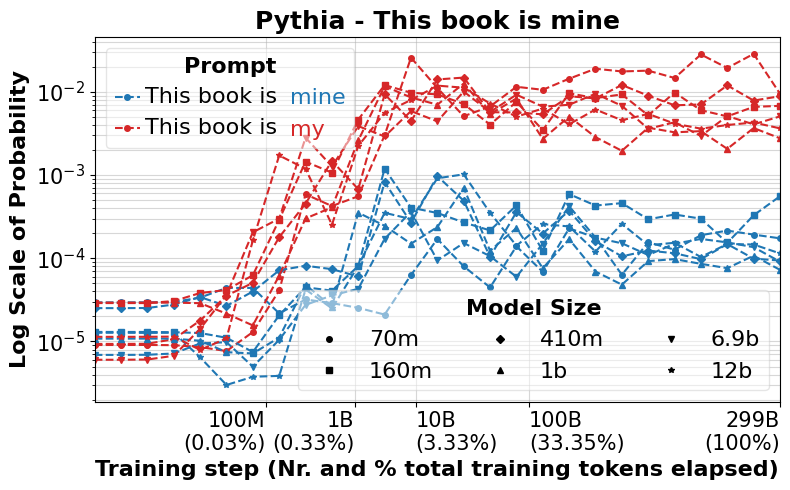}
          \caption{Pythia--Without closing token.}
          \label{fig:morph_wo_dot_pythia}
    \end{subfigure}%
    \begin{subfigure}{.43\textwidth}
          \centering
          \includegraphics[width=\linewidth, trim={1cm 0 0 0}, clip]{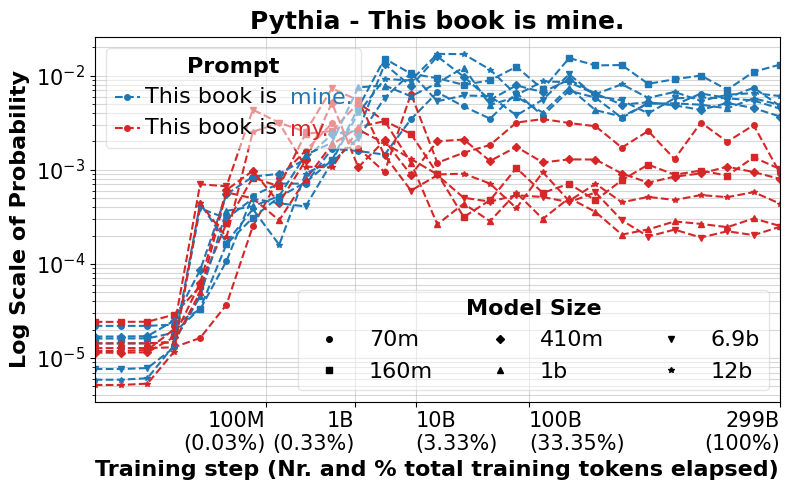}
          \caption{Pythia--With closing token ".".}
          \label{fig:morph_w_dot_pythia}
    \end{subfigure}\\
    
    \caption{Without a closing token, target probabilities shift due to valid alternate continuations after the presumably unfavorable inflection.}
    \label{fig:morph_w_wo_dot_pythia}
\end{figure}

Each case includes an \textit{incipit} prompt followed by a \textit{target}--an inflection of a morpheme followed by one or more \textit{closing} tokens. Each incipit prompt is provided to the LLM, and the generation probability of the target is computed via teacher forcing \citep{teacher_forcing}. The closing tokens would ensure the proper exclusion (or attenuation) of alternate continuations (see Fig. \ref{fig:morph_w_wo_dot_pythia}).
Using simple grammatical rules, we construct many  (\textit{favorable}, \textit{unfavorable}) target pairs---for both of which we track changes in generation probabilities over training steps, expecting the favorable target generation probability to surpass unfavorable one. For example, for the incipit prompt ``I love'', we may have favorable and unfavorable targets ``books.'' and ``book.''. Whenever target $t$ is broken into $N_t$ tokens, the probability of $t$, $P_{t}$, is the product of its token probabilities $p_{i}$ conditioned on the corresponding prior token sequence $x_i$ that starts with the incipit $x$, $log(P_{t}|x) = \sum_{i=1}^{N_{t}}\log(p_{i}|x_i)$. As language acquisition (training) progresses, we expect increasingly higher probabilities for favorable target morphological inflections compared to unfavorable ones.

\subsection{Morphological Structures per Lexical Category}\label{sec:morph_per_lex}
\begin{figure}[h]
    \centering
    \begin{subfigure}{.34\textwidth}
          \centering
          \includegraphics[width=\linewidth]{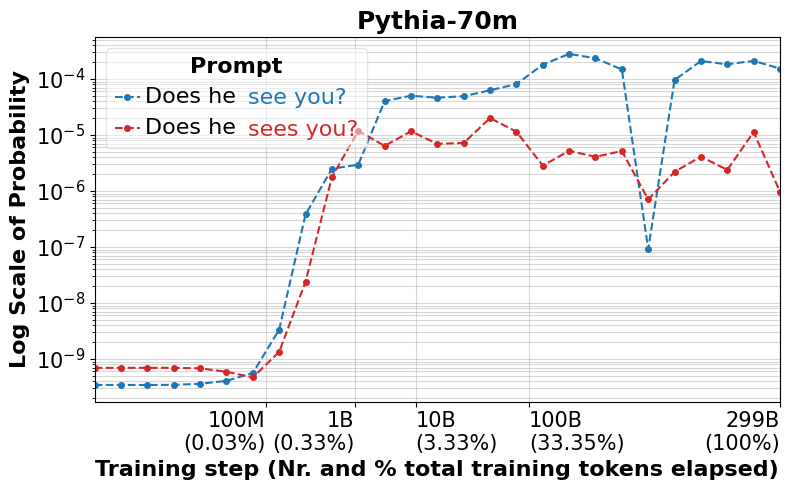}
          \label{fig:morph_eg_pythia_70m_v10}
    \end{subfigure}%
    \begin{subfigure}{.325\textwidth}
          \centering
          \includegraphics[width=\linewidth, trim={1cm 0 0 0}, clip]{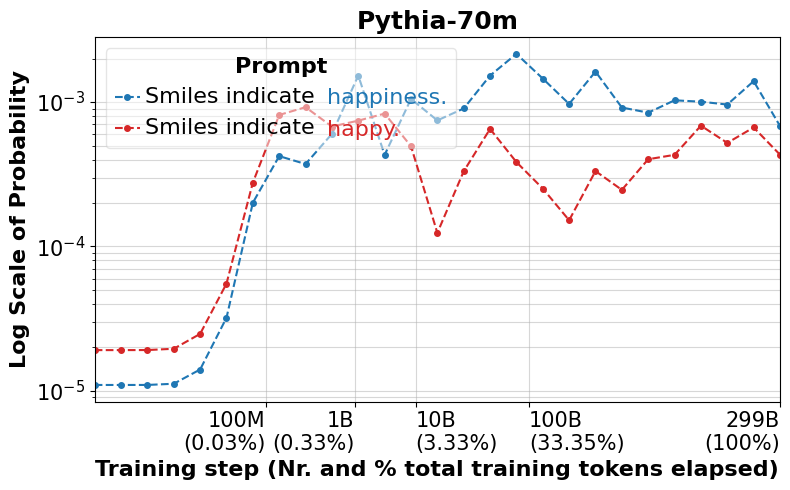}
          \label{fig:morph_eg_pythia_70m_n20}
    \end{subfigure}%
    \begin{subfigure}{.325\textwidth}
          \centering
          \includegraphics[width=\linewidth, trim={1cm 0 0 0}, clip]{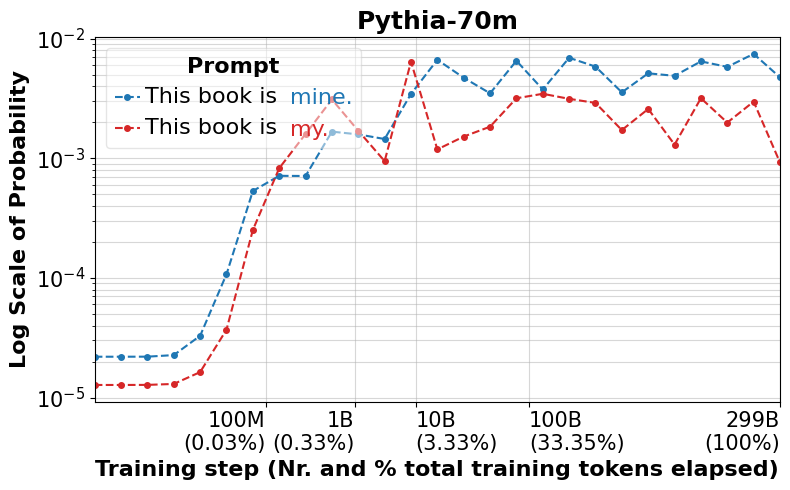}
          \label{fig:morph_eg_pythia_70m_p2}
    \end{subfigure}
    \\
    \vspace{-0.4cm}
    \begin{subfigure}{.34\textwidth}
          \centering
          \includegraphics[width=\linewidth]{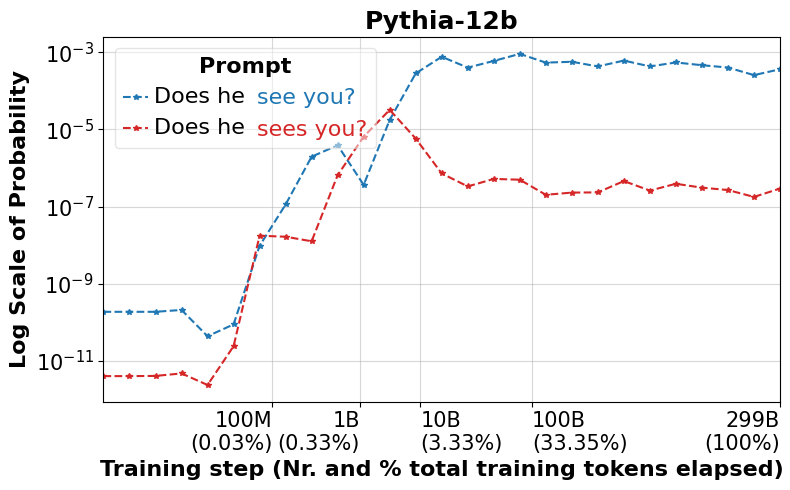}
          \label{fig:morph_eg_pythia_12b_v10}
    \end{subfigure}%
    \begin{subfigure}{.325\textwidth}
          \centering
          \includegraphics[width=\linewidth, trim={1cm 0 0 0}, clip]{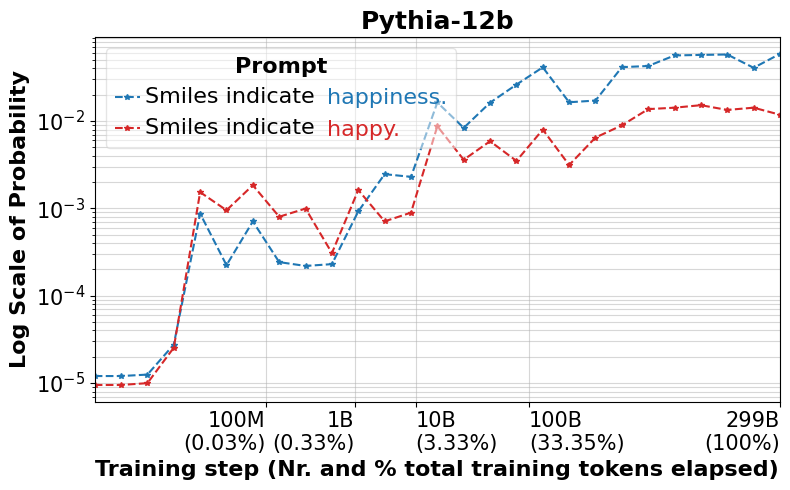}
          \label{fig:morph_eg_pythia_12b_n20}
    \end{subfigure}%
    \begin{subfigure}{.325\textwidth}
          \centering
          \includegraphics[width=\linewidth, trim={1cm 0 0 0}, clip]{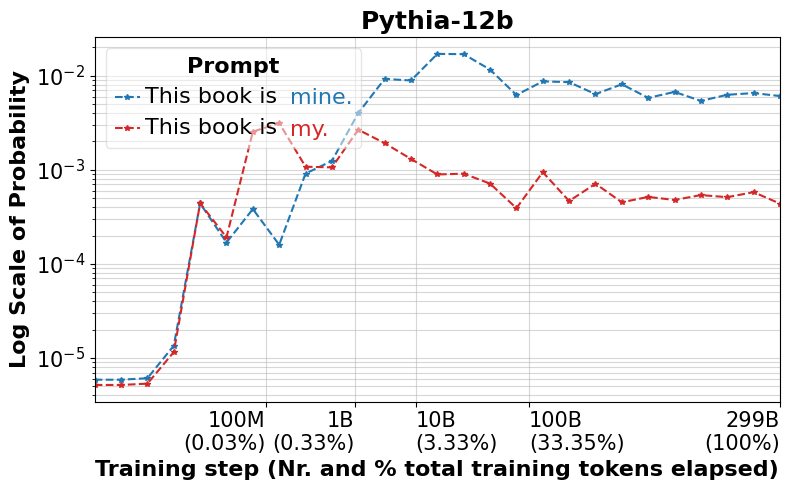}
          \label{fig:morph_eg_pythia_12b_p2}
    \end{subfigure}
    \\
    \vspace{-0.4cm}
    \begin{subfigure}{.34\textwidth}
          \centering
          \includegraphics[width=\linewidth, height=1in]{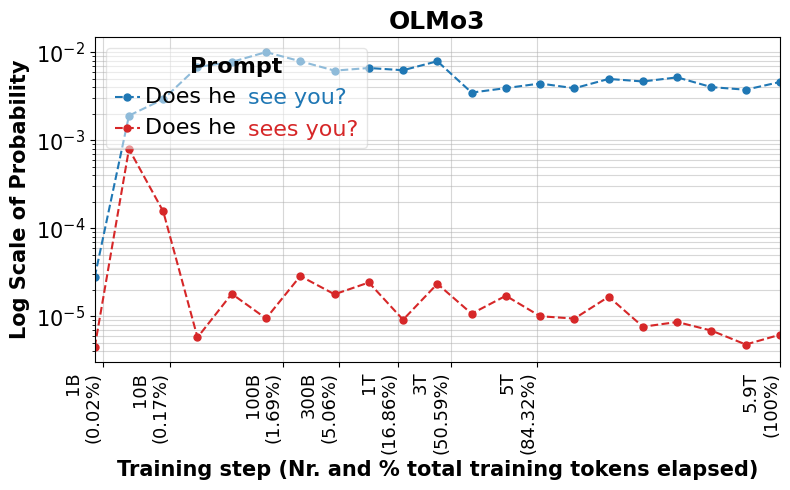}
          \caption{Does he \textcolor{blue}{see you?}}
          \label{fig:morph_eg_olmo3_v10}
    \end{subfigure}%
    \begin{subfigure}{.325\textwidth}
          \centering
          \includegraphics[width=\linewidth, height=1in, trim={1cm 0 0 0}, clip]{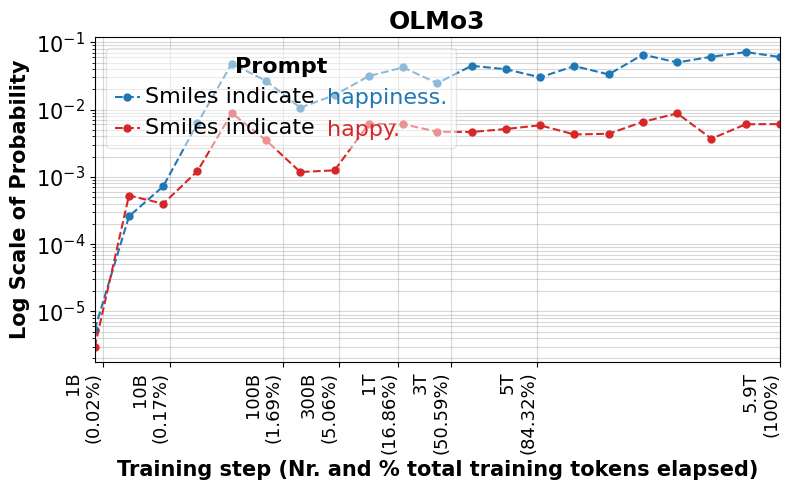}
          \caption{Smiles indicate \textcolor{blue}{happiness.}}
          \label{fig:morph_eg_olmo3_n20}
    \end{subfigure}%
    \begin{subfigure}{.325\textwidth}
          \centering
          \includegraphics[width=\linewidth, height=1in, trim={1cm 0 0 0}, clip]{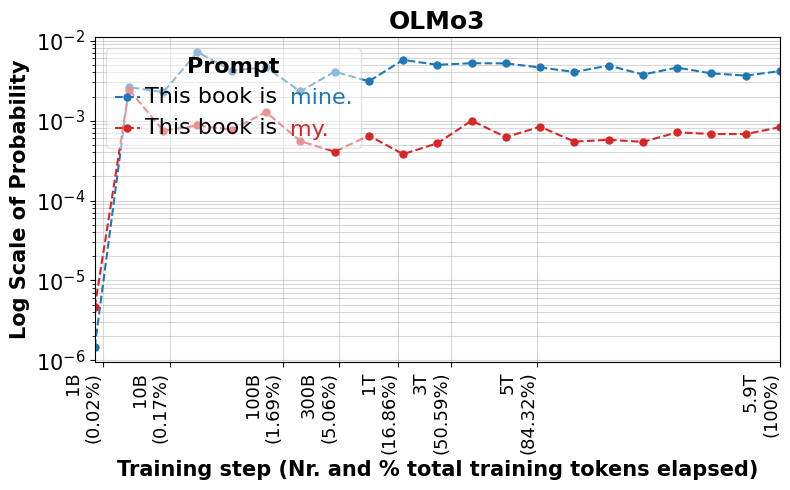}
          \caption{This book is \textcolor{blue}{mine.}}
          \label{fig:morph_eg_olmo3_p2}
    \end{subfigure}
    
    \caption{Morphological inflections across time for Pythia-70M (top), Pythia-12B (middle), and \olmoThree-Stage1 (bottom). 
    The horizontal axis represents the number and percentage of total training tokens elapsed in the training steps; the vertical axis shows the log scale of target words' probabilities.}
    \label{fig:morph_eg}
    \vspace{-0.2cm}
\end{figure}

Figure \ref{fig:morph_eg} shows the acquisition trajectories for one representative case in each of the selected lexical categories. The blue and red graphs show how generation probabilities for favorable and unfavorable targets evolve, respectively.

For Pythia, the blue curves exhibit a ``first turning point'' that marks a transition from an initially steady state to onset of maturation---before or at about 100M tokens for both Pythia-12B and Pythia-70M. With rare exceptions, the probability of favorable targets settle at a higher level than unfavorable targets. We observe that the onset of learning morphological structures---starting at the ``first turning point''-- usually by or before 100M training tokens precedes the onset of semantic maturation of the models (See Sec. \ref{sec:embedding}). Morphological maturation occurs in both Pythia and \olmoThree\ models between 1B and 100B tokens, after a ``second turning point''.

\begin{figure}[!t]
    \centering
    \begin{subfigure}{.34\textwidth}
          \centering
          \includegraphics[width=\linewidth]{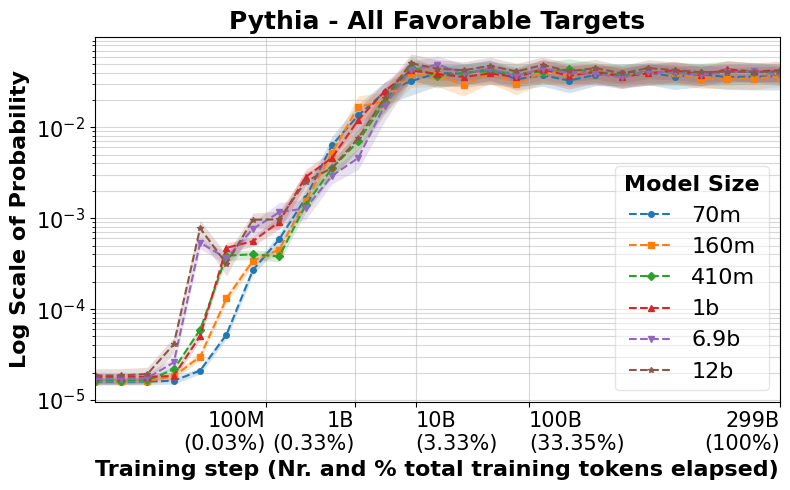}
          \caption{Pythia\textemdash Favorable Targets}
          \label{fig:morph_main_pythia_correct}
    \end{subfigure}%
    \begin{subfigure}{.325\textwidth}
          \centering
          \includegraphics[width=\linewidth, trim={1cm 0 0 0}, clip]{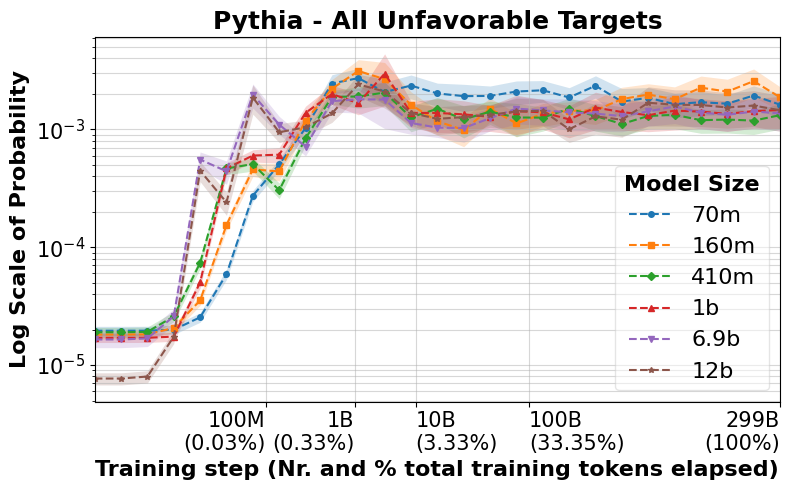}
          \caption{Pythia\textemdash Unfavorable Targets}
          \label{fig:morph_main_pythia_incorrect}
    \end{subfigure}%
    \begin{subfigure}{.325\textwidth}
          \centering
          \includegraphics[width=\linewidth, trim={1cm 0 0 0}, clip]{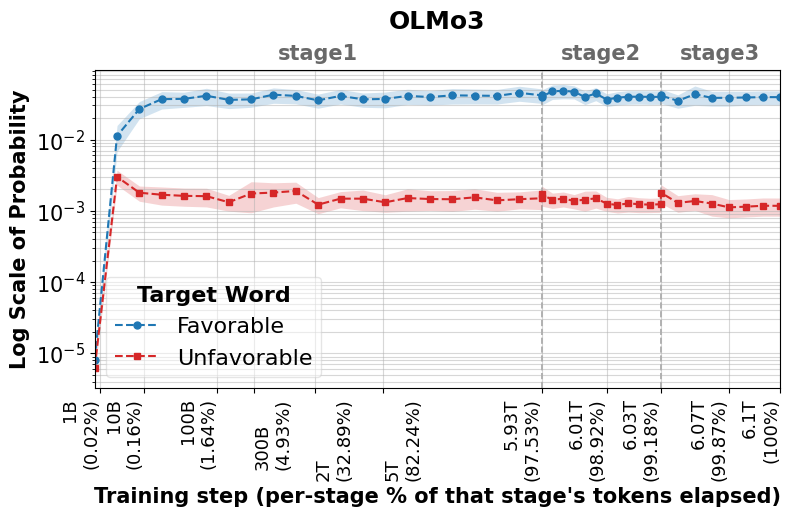}
          \caption{\olmoThree}
          \label{fig:morph_main_olmo3}
    \end{subfigure}%
    
    \caption{Morphological inflections in Pythia suite and \olmoThree\ w.r.t. all unit-test cases. Shadows represent the standard deviation of results.}
    \label{fig:morph_main}
\end{figure}

\subsection{Three Phases in Acquiring Morphological Competence}\label{sec:morph_threephase}
While the examples discussed in Sec. \ref{sec:morph_per_lex} illustrate the details of morphological acquisition, Fig. \ref{fig:morph_main} provides a comprehensive overview aggregated across all unit-test cases (the full list is available in Table \ref{tab:appx_all_morph_eg}) for various model scales over time. As shown in Figs. \ref{fig:morph_main_pythia_correct} and \ref{fig:morph_main_pythia_incorrect}, Pythia models exhibit a characteristic three-phase trajectory: \textit{Initialization:} a steady state, from the earliest steps to the first turning point, which occurs before 100M tokens; \textit{Accelerated Learning:} an increasing trend between the first turning point and second turning point, which occurs at 1B to 10B tokens; and \textit{Maturity Plateau:} a relatively stabilized state after the second turning point. In Fig. \ref{fig:morph_main_olmo3}, \olmoThree\ appears to follow the same three-phase behavior. Unfortunately, \olmoThree\ has not released  early phase checkpoints near the first turning point, which represents the onset of morphological acquisitions. At settlement, after the second turning point, both Pythia and \olmoThree\ prefer to predict favorable morphological forms by a factor of roughly 30 to 35.


\section{Tracing Semantic Stabilization}\label{sec:embedding}
We further explore how embedding drift characterizes semantic acquisition of models over time. \citet{tracking_feature_dynamics} measure the progression of cosine similarity of decoder vectors among intermediate checkpoints and their final state, observing a drift. We measure the cosine similarity of target words' embeddings at any intermediate checkpoints with respect to the final checkpoints' embedding per incipit prompt per model scale.

\begin{figure}[!t]
    \centering
    \setlength{\tabcolsep}{0pt}
    \begin{tabular}{cl}
        \centering
        {
        \renewcommand{\arraystretch}{0.6} 
        \begin{tabular}{@{}l@{}}
            \vspace{-0.4cm}
             \begin{subfigure}{.75\textwidth}
                  \centering
                  \includegraphics[width=\linewidth, trim={0 9.5cm 0 0}, clip]{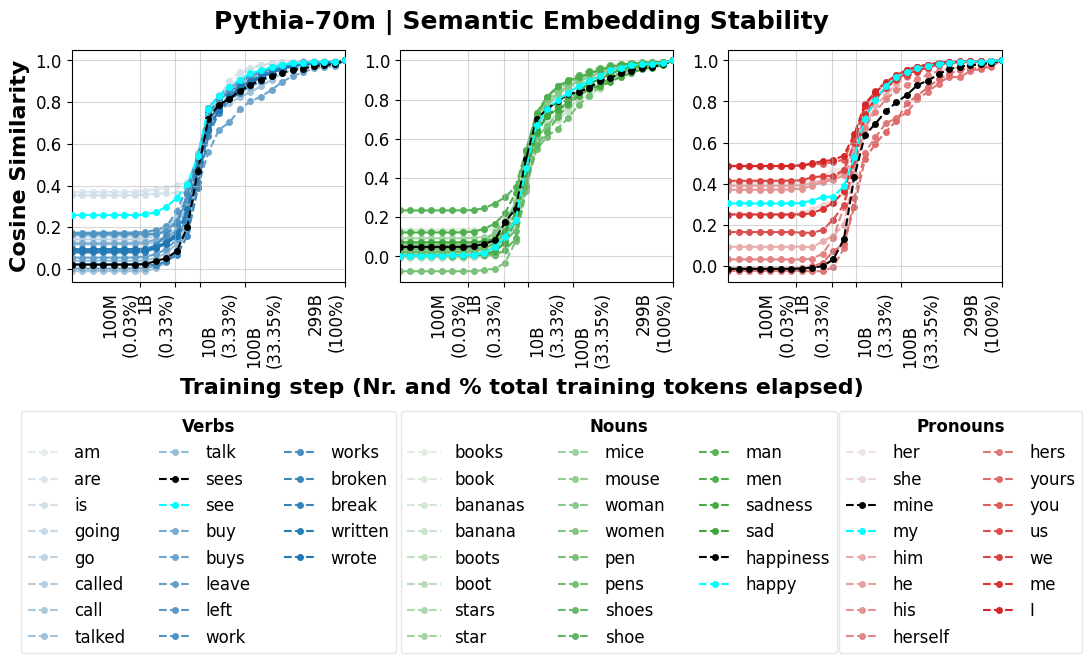}
                  \label{fig:embedding_pythia_70m}
            \end{subfigure}\\
            \vspace{-0.48cm}
             \begin{subfigure}{.75\textwidth}
                  \centering
                  \includegraphics[width=\linewidth, trim={0 7.2cm 0 0.2cm}, clip]{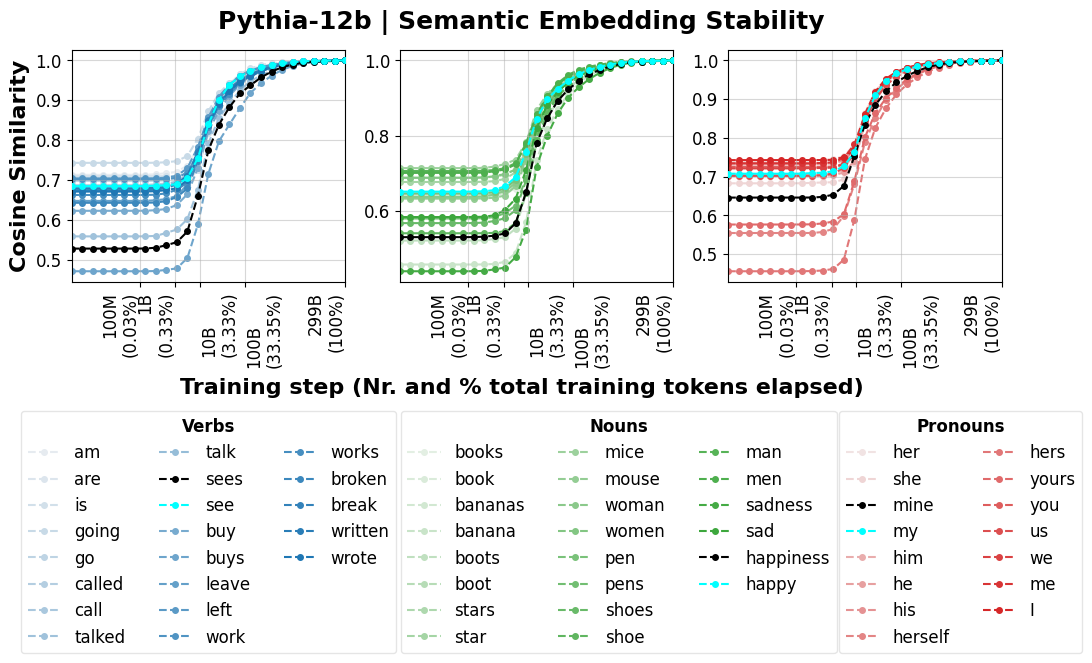}
                  \label{fig:embedding_pythia_12b}
            \end{subfigure}\\
            \cline{1-1}
            \hspace{-.04cm}
            \begin{subfigure}{.75\textwidth}
                  \centering
                  \includegraphics[width=\linewidth, , trim={0 6.5cm 0 0.2cm}, clip]{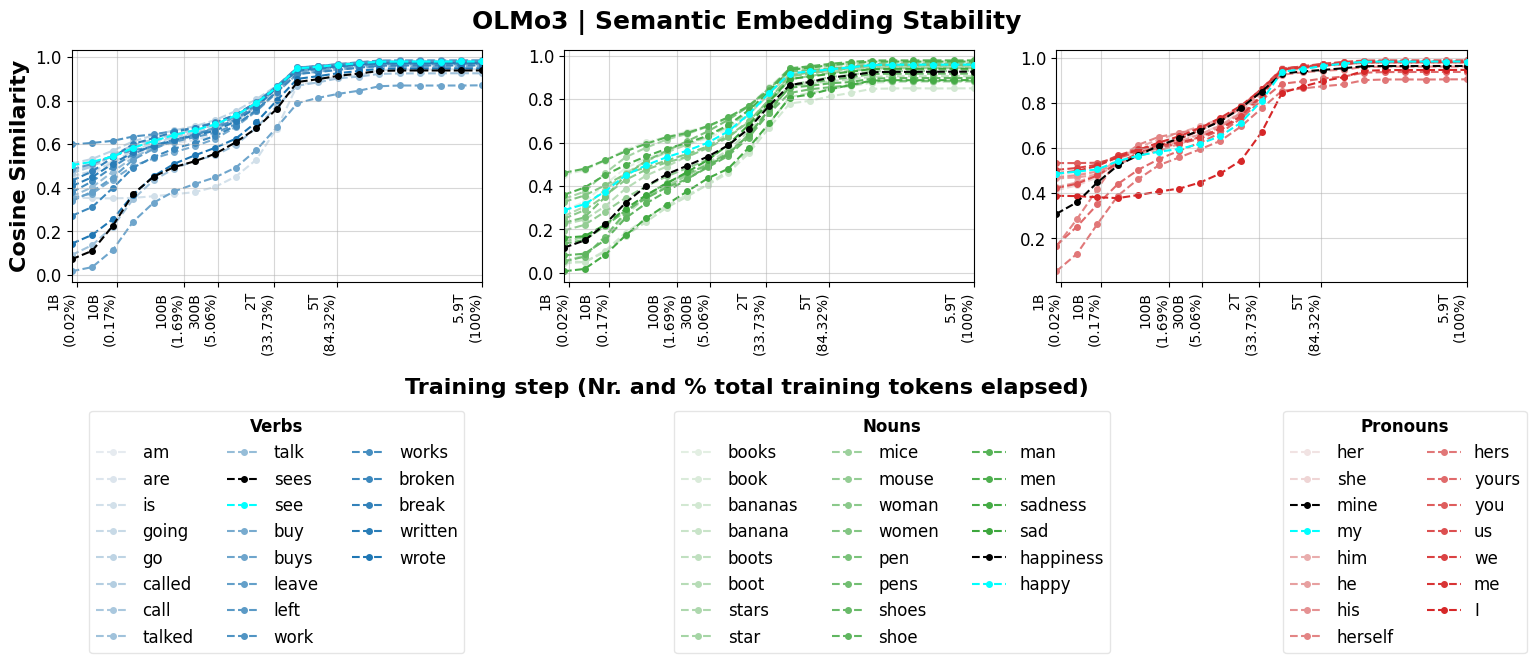}
                  \label{fig:embedding_olmo3}
            \end{subfigure}
        \end{tabular}
        }
        &
        {
        \renewcommand{\arraystretch}{0.6} 
        \begin{tabular}{@{}l@{}}
            \hspace{-1.1cm}
            \begin{subfigure}{.27\textwidth}
                  \centering
                  \includegraphics[width=\linewidth, trim={0 0cm 17.6cm 10.4cm}, clip]{images/embedding/pythia_70m_embed_sim.png}
            \end{subfigure} \\
            \hspace{-0.9cm}
            \begin{subfigure}{.27\textwidth}
                  \centering
                  \includegraphics[width=\linewidth, trim={10.15cm 0cm 6.5cm 10.4cm}, clip]{images/embedding/pythia_70m_embed_sim.png}
            \end{subfigure}\\
            \hspace{-0.6cm}
            \begin{subfigure}{.2\textwidth}
                  \centering
                  \includegraphics[width=\linewidth, trim={21.25cm 0cm 0cm 10.4cm}, clip]{images/embedding/pythia_70m_embed_sim.png}
            \end{subfigure}
        \end{tabular}
        }
    \vspace{-0.5cm}
    \end{tabular}
    \caption{Semantic embedding stability for Pythia-70M (top), Pythia-12B (middle), and \olmoThree-Stage1 (bottom): left (Verbs), middle (Nouns), and right (Pronouns).}
    \label{fig:embedding_all}
\end{figure}

Figure \ref{fig:embedding_all} shows the embedding drifts for Pythia-70M (top row), Pythia-12B (middle row), and \olmoThree\ (bottom row) w.r.t. different lexical categories: Verbs (blue), Nouns (green), and Pronouns (red)\footnote{The results for other Pythia scales, GPT-2, and OLMo2 are available in App. \ref{sec:appx_embedd}.}in left, middle, and right, respectively. The figure adopts a word-centric view, illustrating how the similarity of a specific word with its final state evolves across all training checkpoints. This visualization shows that the similarity initially remains relatively stable, then experiences a sudden increase during training, and gradually transitions into a plateau as it approaches the end. These trends align with the three phases we previously identified in Sec. \ref{sec:morphology}: initialization (a steady phase of high noise and no established semantics in the model's internal activations), onset of accelerated learning (after the first turning point with embeddings moving to their semantic settlement), and maturity plateau (a steady phase where embeddings settling along their final directions). The three-phase structure is consistent with phases in \citep{tracking_feature_dynamics}.

Based on the two top rows in Fig. \ref{fig:embedding_all}, Pythia's onset of semantic acquisition occurs well after 1B training tokens. Furthermore, comparing Figs. \ref{fig:morph_main} and \ref{fig:embedding_all}, we observe that the models experience the onset of morphology acquisition \textit{prior} to the onset of semantic acquisition. Finally, while Pythia experiences a plateau in morphology acquisition between 1B to 10B training tokens, it experiences its semantic-maturity plateaus much \textit{later}---after 100B training tokens. \olmoThree\ shows  similar trends with the onset of morphology plateau at 10B to 100B and semantic plateau at 2T training tokens. We conclude that the onset and maturity of morphological competence occur before the onset and maturity of semantic competence.


\section{Repetition and Co-occurrence}\label{sec:rept}
LLMs rely on the statistics of token patterns in their attention window and perform well on such patterns---e.g., on few-shot exemplars in sequential blocks \citep{llms_vs_human} or on retrieval from  recently listed sets \citep{JohnHopkins}. Following the latter, and to investigate whether semantic divergence in a list is recognized by LLMs, we devised a variation on DAT. DAT was originally designed to measure linguistic creativity \citep{dat_main, bellemarepepin2026divergent}. DAT asks subjects to produce $n$ semantically divergent words and computes a divergence score based on the average mutual cosine similarities in the produced set. Since models can hardly adhere to prompts in their initial training checkpoints and rarely generate appropriate target words, we modify the DAT task to recall-based DAT (rDAT), which emphasizes divergence recognition and recall. As indicated in the first row of Table \ref{tab:rept_prompts}, we prompt the model with a set of 10 words and expect $N_{tw}$ most divergent words to be selected. 
We set $N_{tw}=3$, referring to targets as ``triplets'', and adopt the DAT score to measure divergence in selected triplets. 

\begin{table}[!th]
    \centering
    \caption{Prompt formats for creativity tests, rDAT and RAT.}
    \label{tab:rept_prompts}
    \resizebox{\linewidth}{!}{
        \begin{tabular}{r||l}
        \toprule
        \multicolumn{1}{c||}{\textbf{Test Type}} & \multicolumn{1}{c}{\textbf{Prompt Format}}\\
        \toprule
        \textbf{rDAT} &  ``````Pick {$N_{tw}$} very different words from the list below: \{list of 10 words\}:''''''\\
        \textbf{RAT} & ``````Given three remote stimulus words ``word\_1'', ``word\_2'', ``word\_3'', name the single word that connects all three.''''''\\
        \bottomrule
        \end{tabular}
    }
\end{table}

The total number of distinct triplets built for a 10-word list is equal to $\binom{10}{3}=120$. We sort the triplet in descending order of their DAT scores. For each triplet, we apply teacher forcing \citep{teacher_forcing}---to compute the joint probability of target token sequence conditioned on the incipit token sequence. If the tokenizer breaks a word $w$ (in the target triplet) into $N_w$ tokens, we compute the corresponding geometric mean probabilities as $GM_{w} = \exp{(\sum_{i=1}^{N_{w}}\log(p_{i})/N_{w}})$, where and $p_{i}$ is the conditional probability of $i$-th token in $w$. The  probability for each triplet arrangement is the joint conditional probability of the word-sequence in the triplet, i.e., $P_{j}=\prod_{k=1}^{3}GM_{jk}, j=1, \ldots, 120$. We compute the overall probability of each triplet as the average probability of all its $3!$ permutations. As basis for our study, we select 16 mid-range 10-word lists \footnote{We will refer to such a 10-word list as ``list''. Results using 32 (instead of 16) such lists can be found in App. \ref{sec:appx_div}.},  from the dataset released by \citep{bellemarepepin2026divergent}. Our selected lists have DAT-scores ranging from 60 to 70.

\begin{figure}[!th]
    \centering
    \begin{subfigure}{.34\textwidth}
          \centering
          \includegraphics[width=\linewidth]{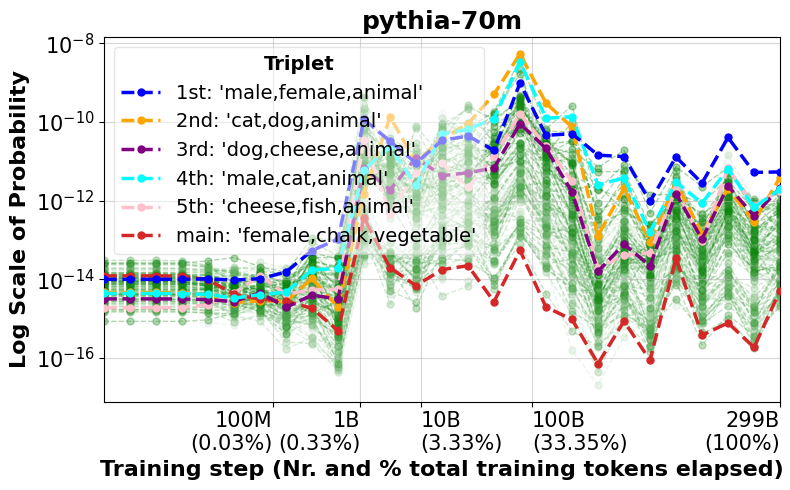}
          \caption{Pythia-70M}
          \label{fig:repet_dat_pythia_70m}
    \end{subfigure}%
    \begin{subfigure}{.325\textwidth}
          \centering
          \includegraphics[width=\linewidth, trim={1cm 0 0 0}, clip]{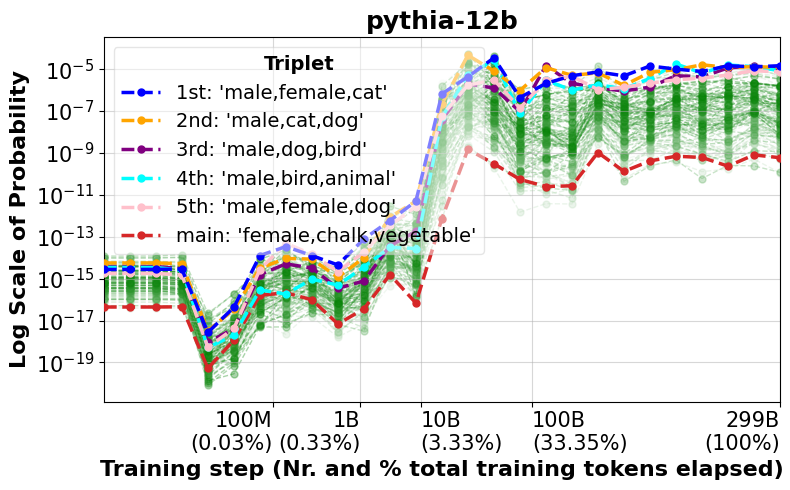}
          \caption{Pythia-12B}
          \label{fig:repet_dat_pythia_12b}
    \end{subfigure}%
    \begin{subfigure}{.325\textwidth}
          \centering
          \includegraphics[width=\linewidth, trim={1cm 0 0 0}, clip]{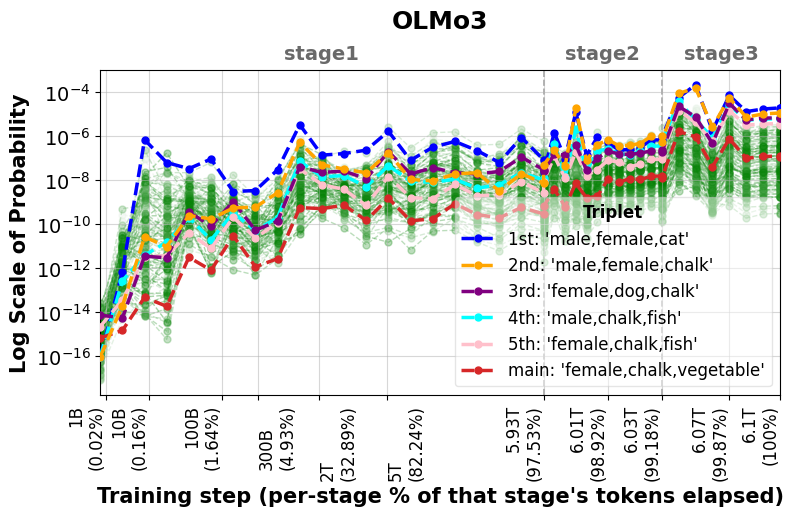}
          \caption{\olmoThree}
          \label{fig:repet_dat_olmo3}
    \end{subfigure}
    
    \caption{Our rDAT results across time for Pythia-70M, Pythia-12B, and \olmoThree with a reference 10-word list ``male, female, cat, dog, chalk, cheese, fish, bird, animal, vegetable''.}
    \label{fig:rept_dat}
\end{figure}

Figure \ref{fig:rept_dat} shows the results for an example rDAT test with the list ``male, female, cat, dog, chalk, cheese, fish, bird, animal, vegetable''. Among its sub-triplets, ``female, chalk, vegetable'' (colored red and labeled ``main'') has the highest DAT score. In addition, we exclude ``main'' and color five triplets whose generation probability at the last checkpoint is greater than other triplets in the test, labeling them as ``1st'' to ``5th''. Figure \ref{fig:rept_dat}, leads to three observations. First, the general rising production probabilities for all 120 triplets confirm the evolving STM, repetition, and recall skills \citep{JohnHopkins}. Second, all three models express accumulated probabilities in triplets with greatest chances of co-occurrence in the training corpus. Third, the increasing main triplet's production probability in \olmoThree\ clearly expresses a growing prompt and divergence recognition capability as it consumes roughly 19 folds more training tokens than either Pythia-70M or Pythia-12B---both of which fails to show this increase in divergence recognition. Larger volumes of training tokens lead to better prompt and divergence recognition.


\section{Semantic Divergence and Phrasal Association Links}\label{sec:creativity}
As another procedural test of creativity, we use RAT. \citet{rat} use a version of RAT which consists of 144 questions. Subjects are asked to find a single word that connects three diverging stimulus words. The prompt format is shown in the second row of Table \ref{tab:rept_prompts}. Based on the resolution rate, we define three categories: Easy, Medium, and Hard, answered by over 60\%, 20\% to 60\%, and under 20\% of participants, respectively. Three sample RAT cases are Easy: ``rocking, wheel, high'' $\rightarrow$ ``chair''; Medium: ``print, berry, bird'' $\rightarrow$ ``blue''; Hard: and ``way, ground, weather'' $\rightarrow$ ``fair''\footnote{Extra evaluations are available in App. \ref{sec:appx_assoc}.}. For each incipit, we measure all 144 target words' probabilities. Overall, we discover that morphological and semantic learning shapes and occurs earlier than prompt and divergence recognition (using rDAT) as well as associative link generation (using RAT). 

\subsection{Divergence Recognition}\label{sec:merc_div}
For each rDAT list, we sort the set of its sub-triplets $L$ based on their DAT scores and choose the top-10 subset\footnote{The results for top-1 and top-5 candidate triplets also show the same trends (see App. \ref{sec:appx_div}).} $L_t$ of triplets, and compute its 
probability sum at the intermediate checkpoint $c$, $S_{t,c} = \sum_{i=1}^{|L_{t}|}P_{i,c}$, (for $c=1, \ldots, C$). We divide this by the probability sum of all $\binom{10}{3}$ triplets either at the first ($S_{DR}=\sum_{i=1}^{|L|}P_{i,1}$) or at the last ($S_{MR}=\sum_{i=1}^{|L|}P_{i,C}$) checkpoint, and refer to these as Development Ratio (DR) and Maturity Ratio (MR), respectively.

\begin{figure}[!th]
    \centering
    \begin{subfigure}{.32\textwidth}
          \centering
          \includegraphics[width=\linewidth]{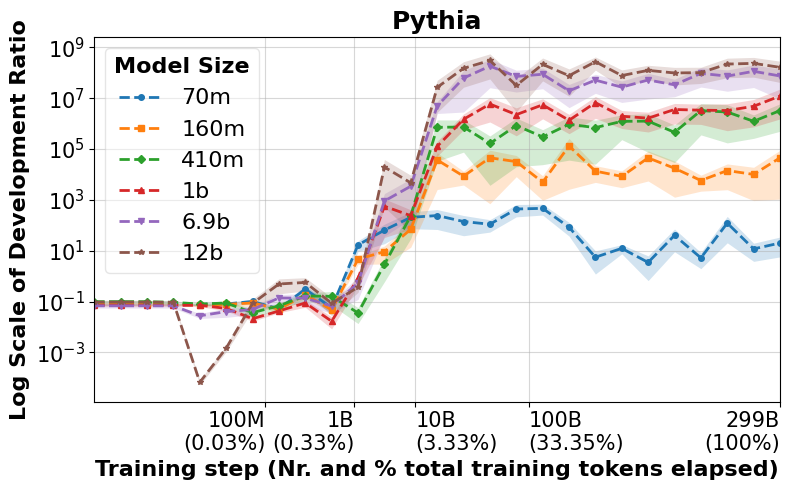}
          \caption{Pythia\textemdash Development Ratio}
          \label{fig:rept_dat_creat_pythia_dr_t10}
    \end{subfigure}%
    \begin{subfigure}{.32\textwidth}
          \centering
          \includegraphics[width=\linewidth]{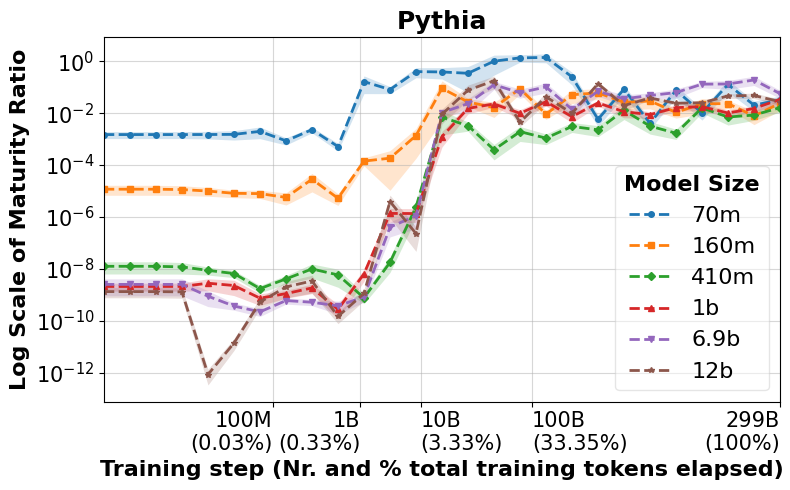}
          \caption{Pythia\textemdash Maturity Ratio}
          \label{fig:rept_dat_creat_pythia_mr_t10}
    \end{subfigure}%
    \begin{subfigure}{.36\textwidth}
          \centering
          \includegraphics[width=\linewidth]{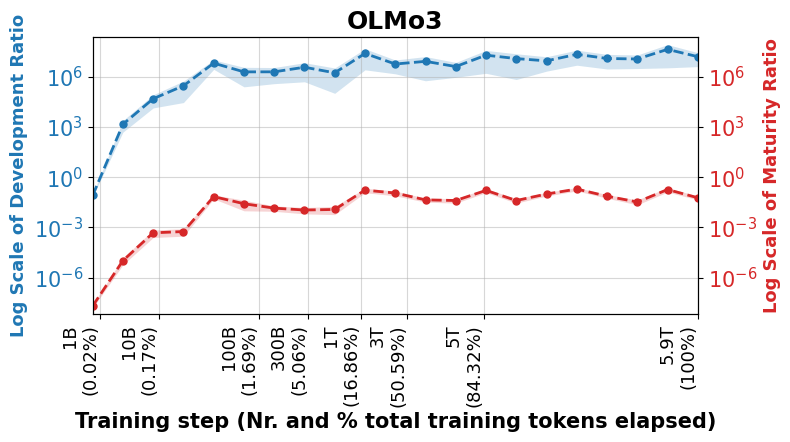}
          \caption{\olmoThree-Stage1}
          \label{fig:rept_dat_creat_olmo3_stage1_t10}
    \end{subfigure}%
    
    \caption{Performance of Pythia and \olmoThree: DR and MR for top-10 triplets of 16 lists in rDAT test.}
    \label{fig:rept_dat_creat}
\end{figure}

The results in Fig. \ref{fig:rept_dat_creat} indicate an approximate value just below $10^{-1}$ for both initial DR and final MR. This implies that both in the beginning, when the entropy is high, and in the end, when the models have matured, the models assign roughly equal probabilities to all triplets so that the DR (at the start) and MR (at the end) are almost equal to the ratio of the number of most divergent triplets to the total 120 possible triplets extracted from the ten words, $10/120$. Almost identical initial DR and final MR values indicate that, handling rDAT, Pythia and \olmoThree\ tend to repeat elements of a list more than they tend to recognize divergence in triplets of elements. This could be a due weakness in rDAT in comparison to RAT, which does not rely on repetition. Note that rDAT shifts the blame for low divergence recognition to the paucity of training data---see Sec. \ref{sec:rept}. The results in Fig. \ref{fig:rept_dat_creat}  underscore a balanced rise in all triplet probabilities, following the semantic plateauing in Fig. \ref{fig:embedding_all}. Analyzing the ratios in Figs. \ref{fig:rept_dat_creat_pythia_dr_t10} and \ref{fig:rept_dat_creat_pythia_mr_t10}, we note that divergent triplets' probability increases not only during training but also to varying degree across scales: Larger models reinforce the production probability more than smaller ones.

\subsection{Association Recognition}\label{sec:merc_assoc}
\begin{figure}[!t]
    \centering
    \begin{subfigure}{.34\textwidth}
          \centering
          \includegraphics[width=\linewidth]{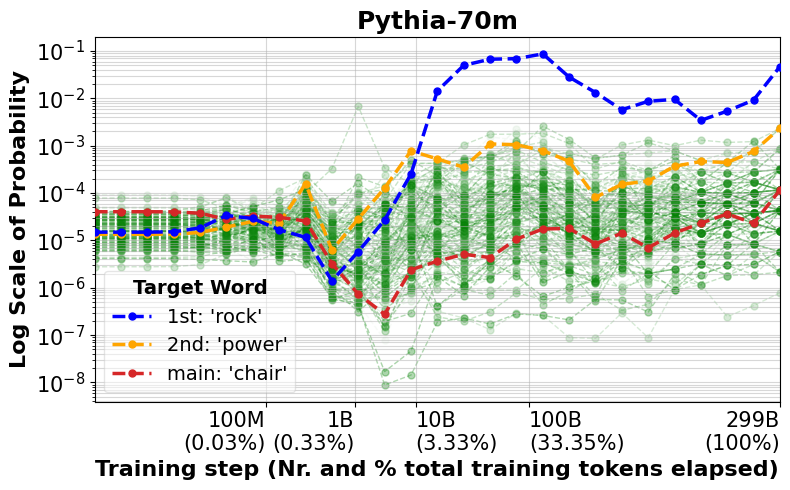}
          \label{fig:rept_rat_pythia_70m_easy_chair}
    \end{subfigure}%
    \begin{subfigure}{.325\textwidth}
          \centering
          \includegraphics[width=\linewidth, trim={1cm 0 0 0}, clip]{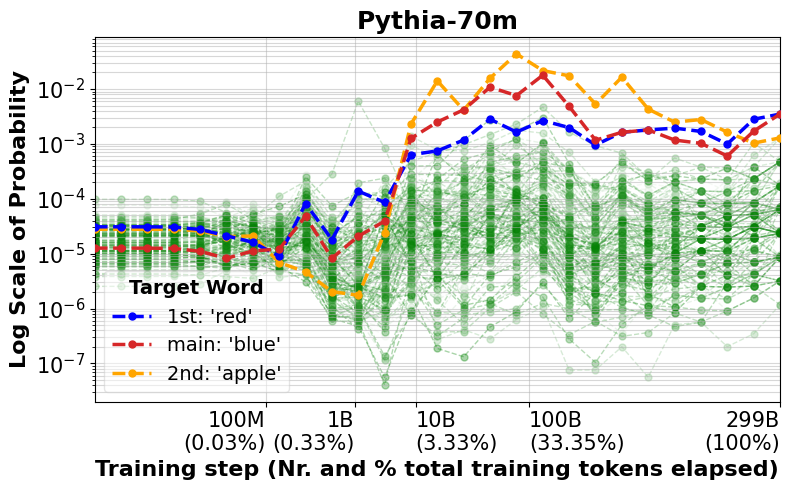}
          \label{fig:rept_rat_pythia_70m_medium_blue}
    \end{subfigure}%
    \begin{subfigure}{.325\textwidth}
          \centering
          \includegraphics[width=\linewidth, trim={1cm 0 0 0}, clip]{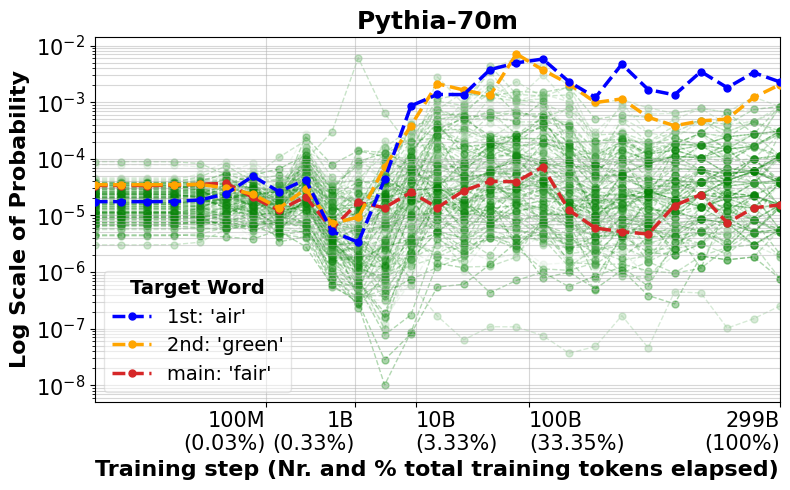}
          \label{fig:rept_rat_pythia_70m_hard_fair}
    \end{subfigure}
    \\
    \vspace{-0.3cm}
    \begin{subfigure}{.34\textwidth}
          \centering
          \includegraphics[width=\linewidth]{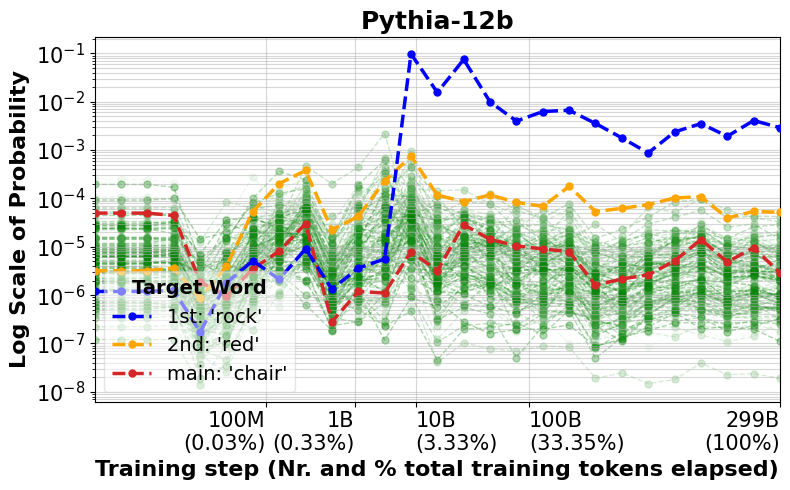}
          \label{fig:rept_rat_pythia_12b_easy_chair}
    \end{subfigure}%
    \begin{subfigure}{.325\textwidth}
          \centering
          \includegraphics[width=\linewidth, trim={1cm 0 0 0}, clip]{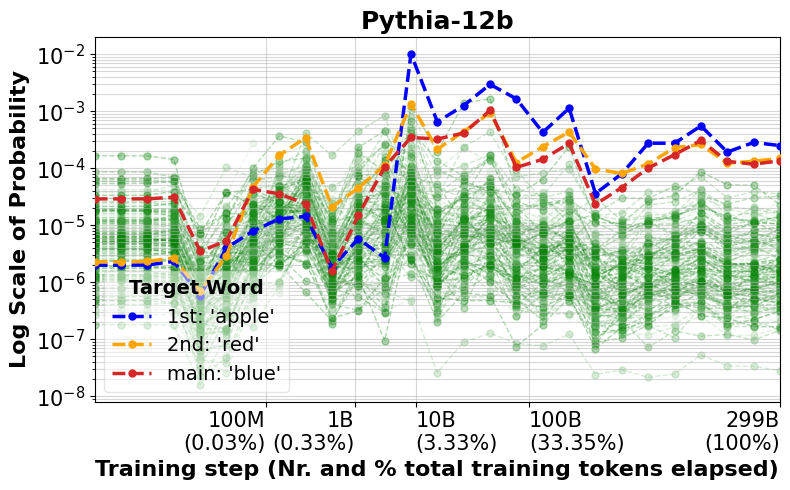}
          \label{fig:rept_rat_pythia_12b_medium_blue}
    \end{subfigure}%
    \begin{subfigure}{.325\textwidth}
          \centering
          \includegraphics[width=\linewidth, trim={1cm 0 0 0}, clip]{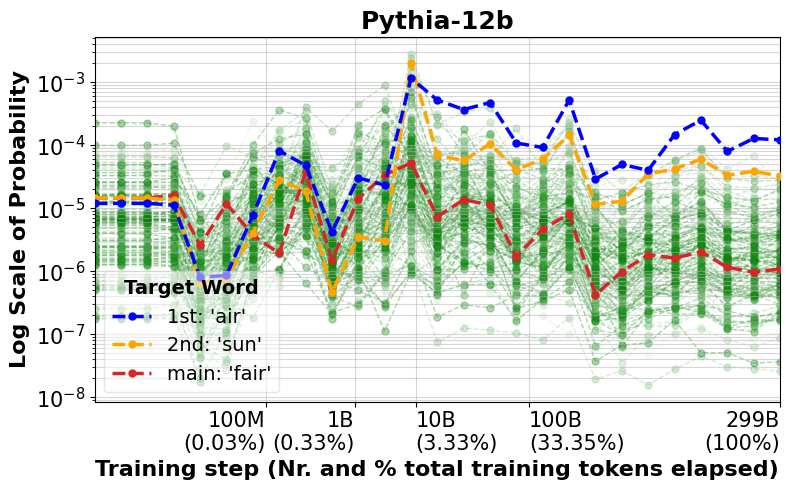}
          \label{fig:rept_rat_pythia_12b_hard_fair}
    \end{subfigure}
    \\
    \vspace{-0.3cm}
    \begin{subfigure}{.34\textwidth}
          \centering
          \includegraphics[width=\linewidth, height=1in]{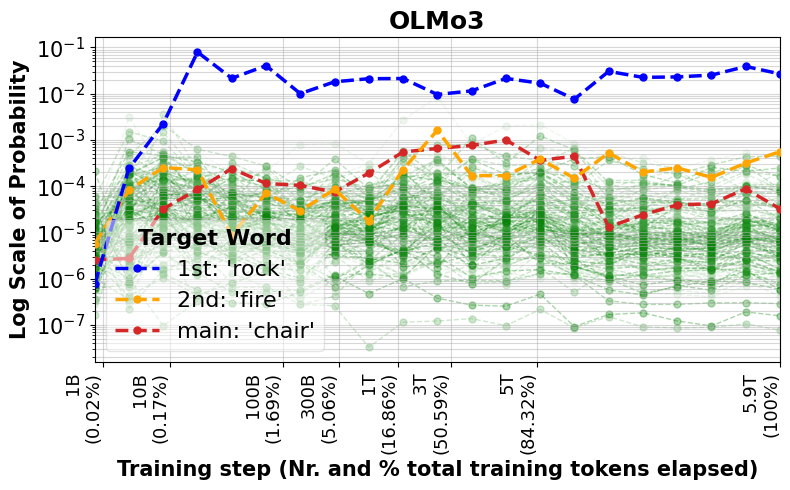}
          \caption{``rocking, wheel, high'' $\rightarrow$ ``chair''}
          \label{fig:rept_rat_olmo3_easy_chair}
    \end{subfigure}%
    \begin{subfigure}{.325\textwidth}
          \centering
          \includegraphics[width=\linewidth, height=1in, trim={1cm 0 0 0}, clip]{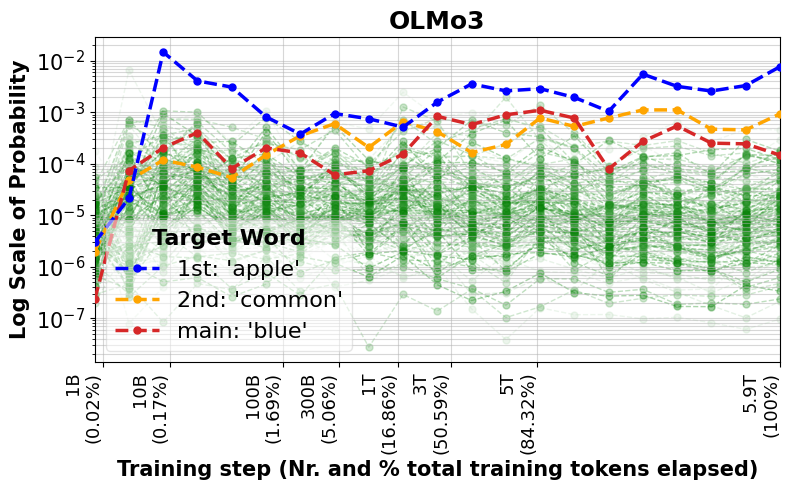}
          \caption{``print, berry, bird'' $\rightarrow$ ``blue''}
          \label{fig:rept_rat_olmo3_medium_blue}
    \end{subfigure}%
    \begin{subfigure}{.325\textwidth}
          \centering
          \includegraphics[width=\linewidth, height=1in, trim={1cm 0 0 0}, clip]{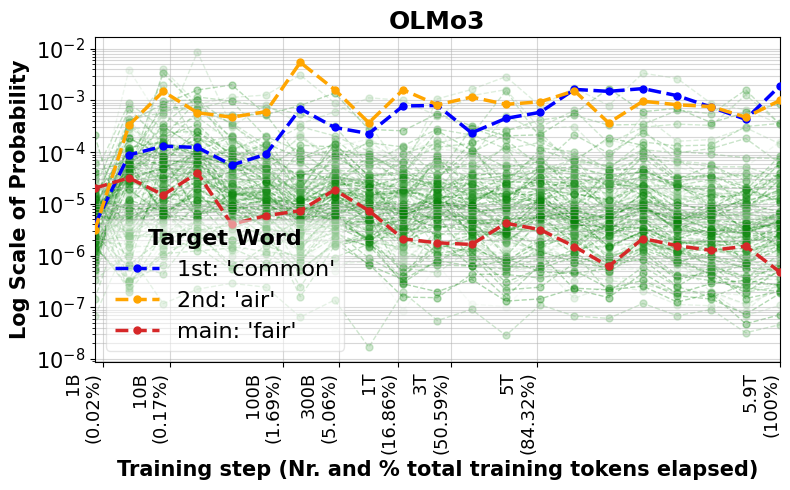}
          \caption{``way, ground, weather'' $\rightarrow$ ``fair''}
          \label{fig:rept_rat_olmo3_hard_fair}
    \end{subfigure}
    
    \caption{RAT results across time for Pythia-70M (top), Pythia-12B (middle), and \olmoThree\ (bottom): (a) Easy; (b) Medium; and (c) Hard cases.}
    \label{fig:rept_rat}
\end{figure}

Figure \ref{fig:rept_rat} shows Pythia-70M (top row), Pythia-12B (middle row), and \olmoThree\ (bottom row) behaviors w.r.t. Easy, Medium, and Hard (RAT) cases, at left, middle, and right columns, respectively. The correct answer is labeled ``main'' and colored red. We also show two targets with the highest probabilities at the last checkpoint. The results show the poor performance of the models in solving RAT-related puzzles. The linked words in the RAT test consist of two words (e.g., blue $+$ berry $\rightarrow$ blueberry in Fig. \ref{fig:rept_rat_olmo3_medium_blue}). This requires models to understand the structural chemistry of how words fuse to form new words. However, both Pythia and \olmoThree\ use BPE tokenizer (a sub-word algorithm to split text into smaller chunks) and default to surface-level thematic associations (fruit $\rightarrow$ apple $\rightarrow$ red).

Figure \ref{fig:rept_rat} shows that the models do not necessarily look for a single word that structurally attaches to all three words. Rather, they seem to prioritize words that frequently co-occur near them in training data. If models cannot find the associating word, they default to a word that has weak but broad links to the divergent triplet. For example, ``rock'' has the highest probability among all targets in the Easy example case (Fig. \ref{fig:rept_rat_olmo3_easy_chair}). We see ``rocking'' as the first word in the incipit prompt. This implies that repetition significantly affects the RAT score of the models, too. On the other hand, Fig. \ref{fig:rept_rat_olmo3_hard_fair} emphasizes the impact of co-occurrence where the tight relationship between ``ground'', ``whether'', ``air'', ``sun'', ``snow'', ``sky'', and ``green'' (all belong to the semantic category of nature, environment, and atmospheric phenomena) directs models to prefer these words over the target word ``fair''.

\section{Conclusion}\label{sec:conclusion}
Our experiments, using checkpoints of models in language training (Pythia and \olmoThree), show that LLMs acquire morphological and syntactic competence before acquiring stable semantic embeddings. Using two operational  measures for creative use of language (rDAT and RAT), we  observe that, in the absence of large enough training data-sets, models barely acquire divergent or associative linguistic competence. Instead, repetition and co-occurrence dominate their behavior. LLM language training progresses through three stages: \textit{Initialization} (embeddings experience high noise and lack any clear semantic structure), \textit{Accelerated Learning} (embeddings move towards semantic stability), and \textit{Maturity Plateau} (embeddings converge to their stable form). Semantic divergence recognition and associative link generation follow the same pattern. Co-occurrence in training corpus and repetition in inference negatively impact these competencies.

Future  possible research directions include: \begin{enumerate*}[label=(\roman*)]
\item 
expanding our morphological dataset;
\item analyzing the impact of model initialization;
    \item investigating the impact of training dataset---for example,  Pythia can also be trained on a larger training dataset, e.g., Dolma-3;
    \item 
    analyzing the behavior of Instruct and Think training\footnote{As a preliminary finding, \olmoTwo-7B-Instruct training has no impact on its morphological preferences. See App. \ref{sec:appx_extend}.};
    \item 
    increasing the number of the checkpoints of models such as \olmoThree\ during language training;
    \item
    devising new creativity measures that are possible to evaluate during language acquisition.
\end{enumerate*}

\clearpage
\bibliography{references}
\bibliographystyle{unsrtnat}

\clearpage
\appendix

\counterwithin{figure}{section}
\counterwithin{table}{section}

\section{Architectural Differences}\label{sec:appx_arch}
In Sec. \ref{sec:models}, we reviewed Pythia suite and \olmoThree. Here, we introduce GPT-2 \citep{gpt2_git} and \olmoTwo\ \citep{olmo2}.

\noindent\textbf{GPT-2:} Built upon the Hugging Face Transformers library \citep{transformers}, the Stanford CRFM team trained five GPT-2 Medium (345M-355M) and five GPT-2 Small (124M) models on the OpenWebText Corpus \citep{openweb} over 209B tokens. Each model corresponds to a unique seed. Only intermediate checkpoints are available for small models: Alias, Battlestar, Caprica, Darkmatter, and Expanse \citep{gpt2_git}.

\noindent\textbf{\olmoTwo:} Prior to \olmoThree, Allen Institute for AI developed \olmoTwo, including core model sizes of 1B, 7B, 13B, and 32B parameters. Each scale features raw base pre-trained models alongside post-trained instruct/chat variations adapted using T\"ulu alignment recipes \citep{olmo2}. Here, we employ \olmoTwo-7B-Base and Instruct, and for simplicity, we call them \olmoTwo\ and \olmoTwo-Instruct, respectively.

Table \ref{tab:appx_arch} indicates the architecture of the under-study LLMs, Pythia \citep{pythia}, GPT-2 \citep{gpt2_blog}, \olmoTwo\ \cite{olmo2}, and \olmoThree\ \cite{olmo3}. Unlike Pythia and GPT-2, which use classic pre-layer normalization, \olmoThree\ maintains the post-normalization structure inherited from \olmoTwo, with applying an additional RMSNorm layer to queries and keys before computing the attention. This prohibits attention logits to overgrow and hence enforces superior training stability at scale. Moreover, compared to Pythia and GPT-2 that are trained over 300B-token \citep{pythia} and 9B-token \citep{openweb} datasets respectively, \olmoTwo\ and \olmoThree\ are trained on the 5.9T-token \cite{olmo3,olmo2} dataset in their pretraining stage. For more information, the readers are referred to the corresponding references \citep{pythia,olmo3,gpt2_git,olmo2}.

\begin{table}[H]
    \vspace{-0.5cm}
    \centering
    \caption{Architectures of under-study LLMs}
    \label{tab:appx_arch}
    \begin{tabular}{lcccc}
        \toprule
        \multicolumn{1}{c}{\textbf{Model}} & \textbf{Size} & \textbf{Layers} & \textbf{Hidden Dimension} & \textbf{Heads}\\
        \midrule
        \multirow{6}{*}{\textbf{Pythia}} 
        & 70M  & 6  & 512  & 8 \\
        & 160M & 12 & 768  & 12\\
        & 410M & 24 & 1024 & 16\\
        & 1B   & 16 & 2048 & 8 \\
        & 6.9B & 32 & 4096 & 32\\
        & 12B  & 36 & 5120 & 40\\
        \midrule
        \textbf{GPT-2}  & All Seeds & 32 & 2560 & 20\\
        \midrule
        \textbf{\olmoTwo/3}  & 7B & 32 & 4096 & 32\\
        \bottomrule
    \end{tabular}
    \vspace{-0.5cm}
\end{table}

\section{Main Checkpoints}\label{sec:appx_checkpoint}
\textbf{Pythia.} We conduct experiments for the following checkpoints: \{0, 1, 2, 4, 8, 16, 32, 64, 128, 256, 512, 1000, 3000, 13000, 23000, 33000, 43000, 53000, 63000, 73000, 83000, 93000, 103000, 113000, 123000, 133000, 143000\}.

\textbf{GPT-2.} The following training checkpoints are used for GPT-2 small evaluations: \{0, 10, 50, 100, 150, 200, 500, 1000, 1500, 2000, 5000, 10000, 12000, 15000, 20000, 50000, 100000, 120000, 150000, 200000, 250000, 300000, 350000, 360000, 370000, 380000, 390000, 400000\}.

\textbf{\olmoTwo.} We choose the following checkpoints among all available checkpoints for \olmoTwo-7B-Base: \{1000, 2000, 4000, 8000, 16000, 32000, 64000, 128000, 256000, 512000, 600000, 700000, 800000, 900000, 910000, 920000, 928000, 928646, 'main'\}. We use all released checkpoints for \olmoTwo-7B-Instruct: \{60, 120, 180, 240, 300, 360, 'main'\}.

\textbf{\olmoThree.} The following checkpoints are used per stage:

\begin{itemize}[leftmargin=0.5cm]
    \item Stage 1: \{0, 1000, 2000, 4000, 8000, 16000, 32000, 64000, 128000, 256000, 512000, 1024000, 1100000, 1200000, 1300000, 1400000, 1410000, 1411000, 1412000, 1413000, 1413814\}.
    \item Stage 2: \{1000, 2000, 4000, 8000, 16000, 32000, 40000, 42000, 44000, 46000, 47000, 47600\}.
    \item Stage 3: \{1000, 2000, 4000, 8000, 10000, 11000, 11900, 11921, 'main'\}.
\end{itemize}

\section{Computational Resources}\label{sec:appx_comp}
Our computational framework utilizes NVIDIA H100 NVL GPUs, each featuring 94 GB of VRAM (from a total pool of 16 available nodes). Experiments are conducted on either a single-GPU setup for smaller configurations or a two-GPU data-parallel configuration for larger models. Evaluations are performed in couple of hours for all morphology, RAT, and rDAT tests.

\section{Extending Results}\label{sec:appx_extend}
In Sec. \ref{sec:models}, we reviewed Pythia suite and \olmoThree. Here, we first introduce GPT-2 \citep{gpt2_git,gpt2_checkpoints} and \olmoTwo\ \citep{olmo2}. We then present the extended evaluation results within each relevant section for Pythia suite (160M, 410M, 1B, 6.9B), GPT-2, \olmoTwo, and \olmoTwo-Instruct.

\begin{table}[!h]
    \vspace{-0.3cm}
    \centering
    \caption{List of unit-test cases in various lexical categories for some common morphological inflections. Here, \textvisiblespace{} shows a space character.}
    \label{tab:appx_all_morph_eg}
    \resizebox{\textwidth}{!}{%
        \begin{tabular}{ccc||cc||cc}
        \toprule
        \multirow{2}{*}{\textbf{Prompt Type}} & \multicolumn{2}{c}{\textbf{Verbs}} & \multicolumn{2}{c}{\textbf{Nouns}} & \multicolumn{2}{c}{\textbf{Pronouns}}\\
        \cmidrule(lr){2-3} \cmidrule(lr){4-5} \cmidrule(lr){6-7}
        & \multicolumn{1}{c}{\textbf{Incipit Prompt}} & \multicolumn{1}{c}{\textbf{Target Word}} & \multicolumn{1}{c}{\textbf{Incipit Prompt}} & \multicolumn{1}{c}{\textbf{Target Word}} & \multicolumn{1}{c}{\textbf{Incipit Prompt}} & \multicolumn{1}{c}{\textbf{Target Word}}\\
        \toprule
        \multicolumn{1}{l||}{\textbf{Favorable}} & \multirow{2}{*}{I} & \textcolor{blue}{am\textvisiblespace{}} & \multirow{2}{*}{I love} & \textcolor{blue}{books.} & \multirow{2}{*}{Every Wednesday, he visits} & \textcolor{blue}{her.}\\
        \multicolumn{1}{l||}{\textbf{Unfavorable}} &  & \textcolor{red}{are\textvisiblespace{}} &  & \textcolor{red}{book.} &  & \textcolor{red}{she.}\\
        \midrule
        \multicolumn{1}{l||}{\textbf{Favorable}} & \multirow{2}{*}{You} & \textcolor{blue}{are\textvisiblespace{}} & \multirow{2}{*}{I eat} & \textcolor{blue}{bananas.} & \multirow{2}{*}{This book is} & \textcolor{blue}{mine.}\\
        \multicolumn{1}{l||}{\textbf{Unfavorable}} &  & \textcolor{red}{is\textvisiblespace{}} &  & \textcolor{red}{banana.} &  & \textcolor{red}{my.}\\
        \midrule
        \multicolumn{1}{l||}{\textbf{Favorable}} & \multirow{2}{*}{I am} & \textcolor{blue}{going\textvisiblespace{}} & \multirow{2}{*}{He wears} & \textcolor{blue}{boots.} & \multirow{2}{*}{His father liked to talk to} & \textcolor{blue}{him.}\\
        \multicolumn{1}{l||}{\textbf{Unfavorable}} &  & \textcolor{red}{go\textvisiblespace{}} &  & \textcolor{red}{boot.} &  & \textcolor{red}{he.}\\
        \midrule
        \multicolumn{1}{l||}{\textbf{Favorable}} & \multirow{2}{*}{He has} & \textcolor{blue}{called.} & \multirow{2}{*}{He gazed at these} & \textcolor{blue}{stars.} & \multirow{2}{*}{She loves} & \textcolor{blue}{him.}\\
        \multicolumn{1}{l||}{\textbf{Unfavorable}} &  & \textcolor{red}{call.} &  & \textcolor{red}{star.} &  & \textcolor{red}{his.}\\
        \midrule
        \multicolumn{1}{l||}{\textbf{Favorable}} & \multirow{2}{*}{We had} & \textcolor{blue}{talked.} & \multirow{2}{*}{Most cats eat most} & \textcolor{blue}{mice.} & \multirow{2}{*}{She went out by} & \textcolor{blue}{herself.}\\
        \multicolumn{1}{l||}{\textbf{Unfavorable}} &  & \textcolor{red}{talk.} &  & \textcolor{red}{mouse.} &  & \textcolor{red}{hers.}\\
        \midrule
        \multicolumn{1}{l||}{\textbf{Favorable}} & \multirow{2}{*}{Does he} & \textcolor{blue}{see you?} & \multirow{2}{*}{He falls in love with every beautiful} & \textcolor{blue}{woman.} & \multirow{2}{*}{His dog ran to} & \textcolor{blue}{him.}\\
        \multicolumn{1}{l||}{\textbf{Unfavorable}} &  & \textcolor{red}{sees you?} &  & \textcolor{red}{women.} &  & \textcolor{red}{he.}\\
        \midrule
        \multicolumn{1}{l||}{\textbf{Favorable}} & \multirow{2}{*}{He} & \textcolor{blue}{sees you.} & \multirow{2}{*}{The boy wrote with this} & \textcolor{blue}{pen.} & \multirow{2}{*}{This book is} & \textcolor{blue}{yours.}\\
        \multicolumn{1}{l||}{\textbf{Unfavorable}} &  & \textcolor{red}{see you.} &  & \textcolor{red}{pens.} &  & \textcolor{red}{you.}\\
        \midrule
        \multicolumn{1}{l||}{\textbf{Favorable}} & \multirow{2}{*}{She can} & \textcolor{blue}{buy\textvisiblespace{}} & \multirow{2}{*}{She bought a new pair of} & \textcolor{blue}{shoes.} & \multirow{2}{*}{It depends on} & \textcolor{blue}{us.}\\
        \multicolumn{1}{l||}{\textbf{Unfavorable}} &  & \textcolor{red}{buys\textvisiblespace{}} &  & \textcolor{red}{shoe.} &  & \textcolor{red}{we.}\\
        \midrule
        \multicolumn{1}{l||}{\textbf{Favorable}} & \multirow{2}{*}{May I} & \textcolor{blue}{leave?} & \multirow{2}{*}{At the gathering, she met a handsome} & \textcolor{blue}{man.} & \multirow{2}{*}{It is all about} & \textcolor{blue}{me.}\\
        \multicolumn{1}{l||}{\textbf{Unfavorable}} &  & \textcolor{red}{left?} &  & \textcolor{red}{men.} &  & \textcolor{red}{I.}\\
        \midrule
        \multicolumn{1}{l||}{\textbf{Favorable}} & \multirow{2}{*}{It should} & \textcolor{blue}{work.} & \multirow{2}{*}{She was overwhelmed by a deep} & \textcolor{blue}{sadness.} &  & \\
        \multicolumn{1}{l||}{\textbf{Unfavorable}} &  & \textcolor{red}{works.} &  & \textcolor{red}{sad.} &  & \\
        \midrule
        \multicolumn{1}{l||}{\textbf{Favorable}} & \multirow{2}{*}{The radio is} & \textcolor{blue}{broken.} & \multirow{2}{*}{Smiles indicate} & \textcolor{blue}{happiness.} &  & \\
        \multicolumn{1}{l||}{\textbf{Unfavorable}} &  & \textcolor{red}{break.} &  & \textcolor{red}{happy.} &  & \\
        \midrule
        \multicolumn{1}{l||}{\textbf{Favorable}} & \multirow{2}{*}{The book is} & \textcolor{blue}{written.} & & & & \\
        \multicolumn{1}{l||}{\textbf{Unfavorable}} &  & \textcolor{red}{wrote.} & & & & \\
        \bottomrule
        \end{tabular}
    }
    \vspace{-0.5cm}
\end{table}
\begin{figure}[H]
    \centering
    \begin{subfigure}{.48\textwidth}
          \centering
          \includegraphics[width=\linewidth]{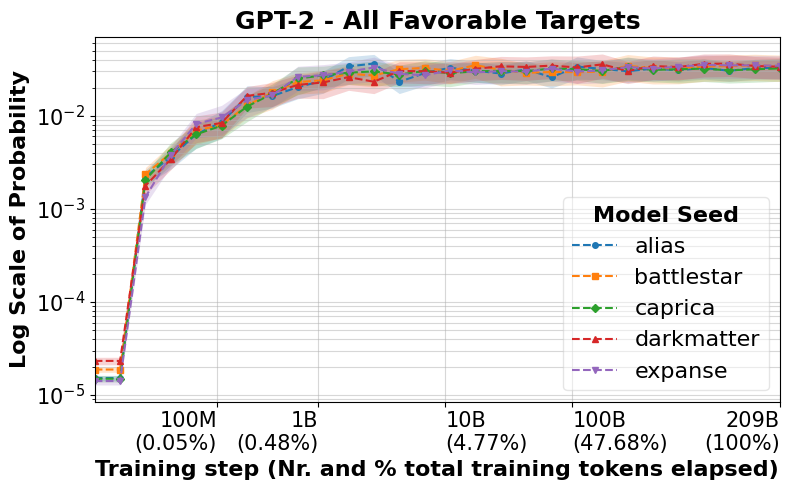}
          \caption{GPT-2\textemdash Favorable Targets}
          \label{fig:appx_morph_main_gpt2_correct}
    \end{subfigure}%
    \begin{subfigure}{.46\textwidth}
          \centering
          \includegraphics[width=\linewidth, trim={1cm 0 0 0}, clip]{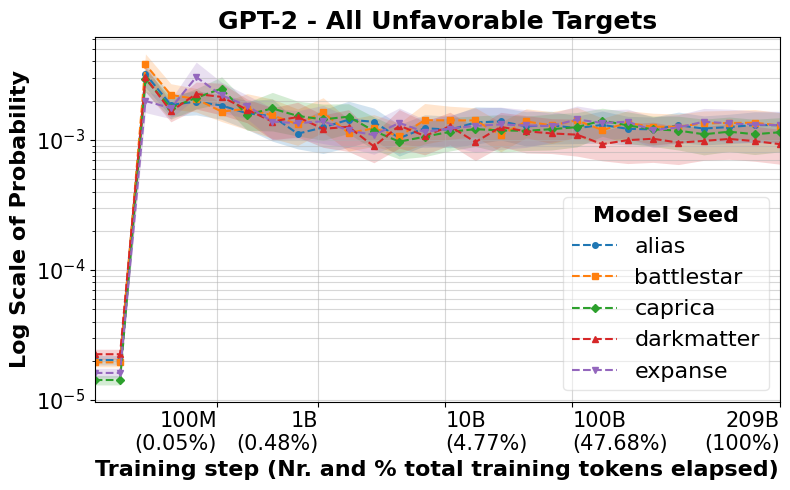}
          \caption{GPT-2\textemdash Unfavorable Targets}
          \label{fig:appx_morph_main_gpt2_incorrect}
    \end{subfigure}\\
    \begin{subfigure}{.48\textwidth}
          \centering
          \includegraphics[width=\linewidth, trim={0 0 0 0}, clip]{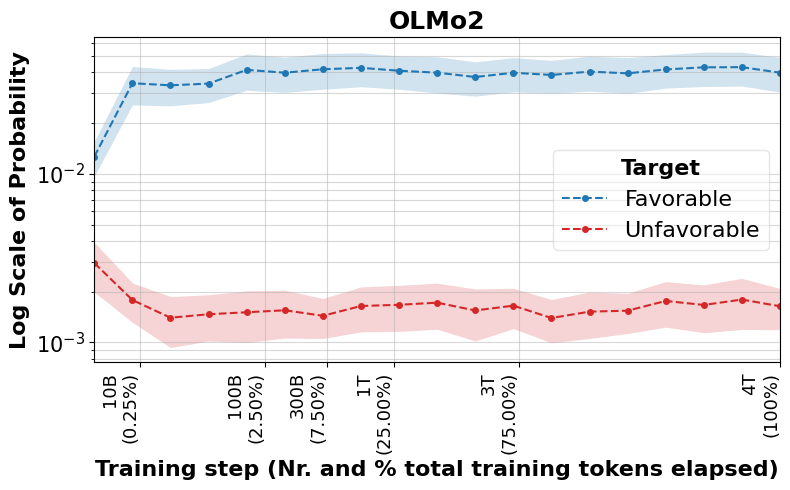}
          \caption{\olmoTwo}
          \label{fig:appx_morph_main_olmo2}
    \end{subfigure}%
    \begin{subfigure}{.46\textwidth}
          \centering
          \includegraphics[width=\linewidth, trim={1cm 0 0 0}, clip]{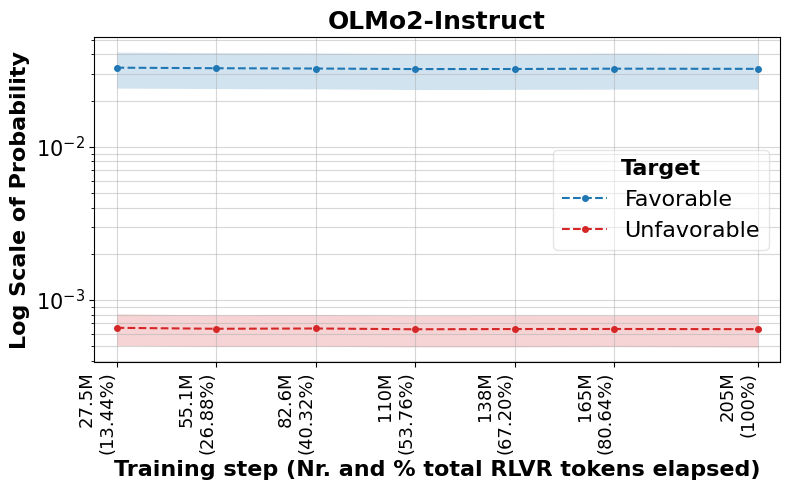}
          \caption{\olmoTwo-Instruct}
          \label{fig:appx_morph_main_olmo2_instruct}
    \end{subfigure}
    
    \caption{Morphological inflections in GPT-2 and \olmoTwo variants w.r.t. all unit-test cases.}
    \label{fig:appx_morph_main}
    \vspace{-0.5cm}
\end{figure}
\clearpage

\begin{figure}[H]
    \centering
    \vspace{-0.3cm}
    \begin{subfigure}{.34\textwidth}
          \centering
          \includegraphics[width=\linewidth]{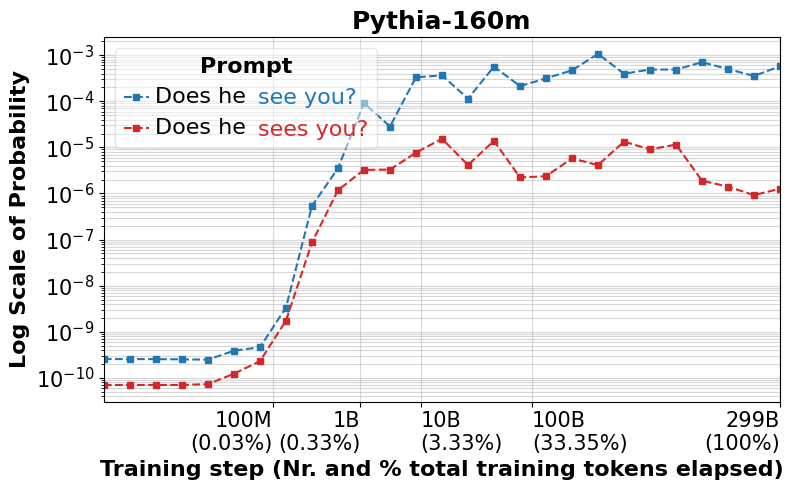}
          \label{fig:appx_morph_eg_pythia_160m_v10}
    \end{subfigure}%
    \begin{subfigure}{.325\textwidth}
          \centering
          \includegraphics[width=\linewidth, trim={1cm 0 0 0}, clip]{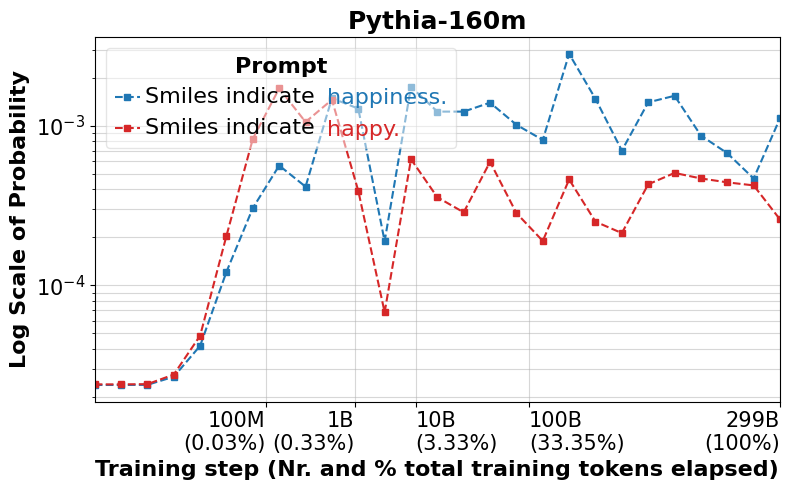}
          \label{fig:appx_morph_eg_pythia_160m_n20}
    \end{subfigure}%
    \begin{subfigure}{.325\textwidth}
          \centering
          \includegraphics[width=\linewidth, trim={1cm 0 0 0}, clip]{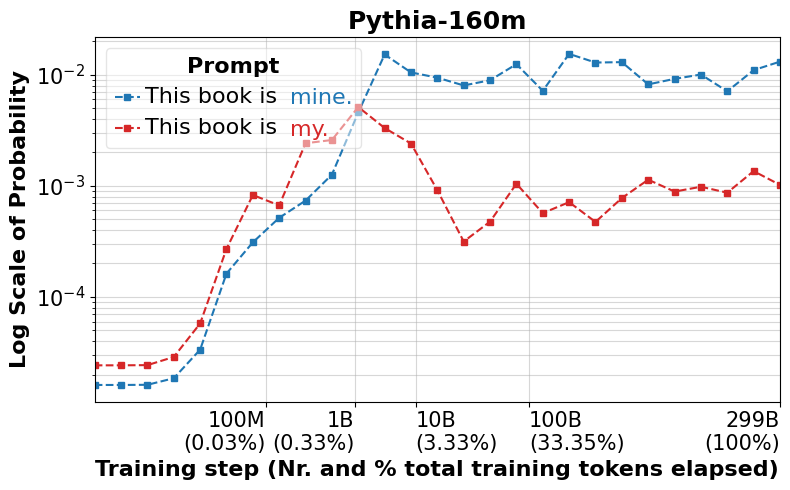}
          \label{fig:appx_morph_eg_pythia_160m_p2}
    \end{subfigure}
    \\
    \vspace{-0.3cm}
    \begin{subfigure}{.34\textwidth}
          \centering
          \includegraphics[width=\linewidth]{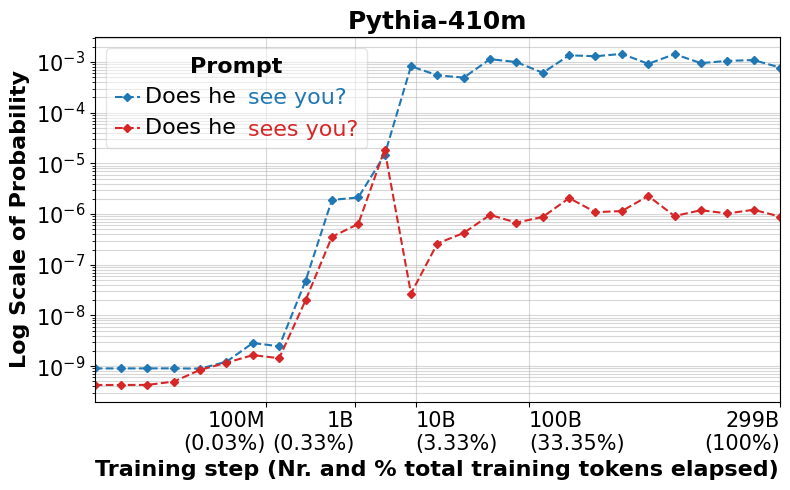}
          \label{fig:appx_morph_eg_pythia_410m_v10}
    \end{subfigure}%
    \begin{subfigure}{.325\textwidth}
          \centering
          \includegraphics[width=\linewidth, trim={1cm 0 0 0}, clip]{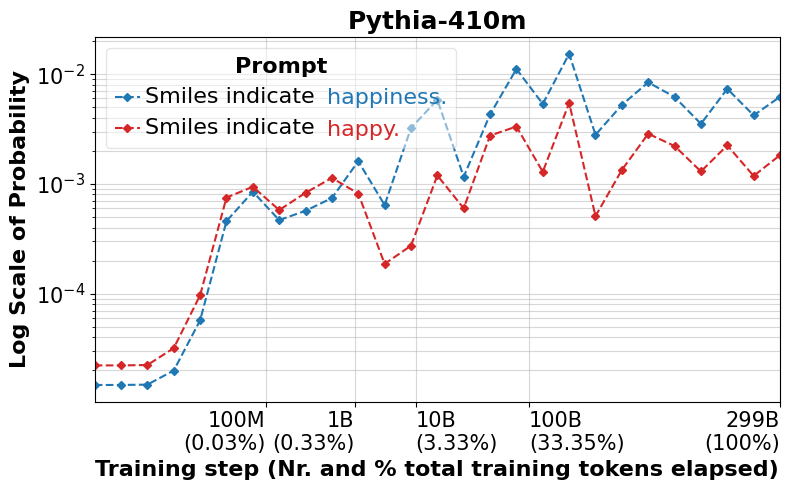}
          \label{fig:appx_morph_eg_pythia_410m_n20}
    \end{subfigure}%
    \begin{subfigure}{.325\textwidth}
          \centering
          \includegraphics[width=\linewidth, trim={1cm 0 0 0}, clip]{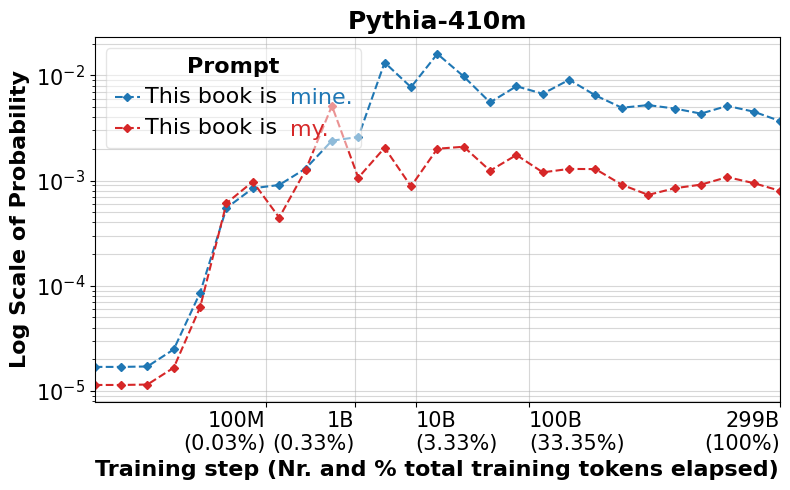}
          \label{fig:appx_morph_eg_pythia_410m_p2}
    \end{subfigure}
    \\
    \vspace{-0.3cm}
    \begin{subfigure}{.34\textwidth}
          \centering
          \includegraphics[width=\linewidth]{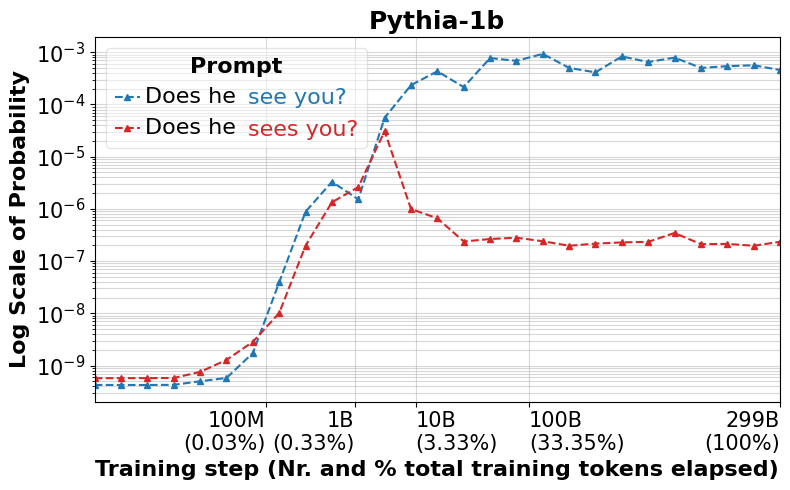}
          \label{fig:appx_morph_eg_pythia_1b_v10}
    \end{subfigure}%
    \begin{subfigure}{.325\textwidth}
          \centering
          \includegraphics[width=\linewidth, trim={1cm 0 0 0}, clip]{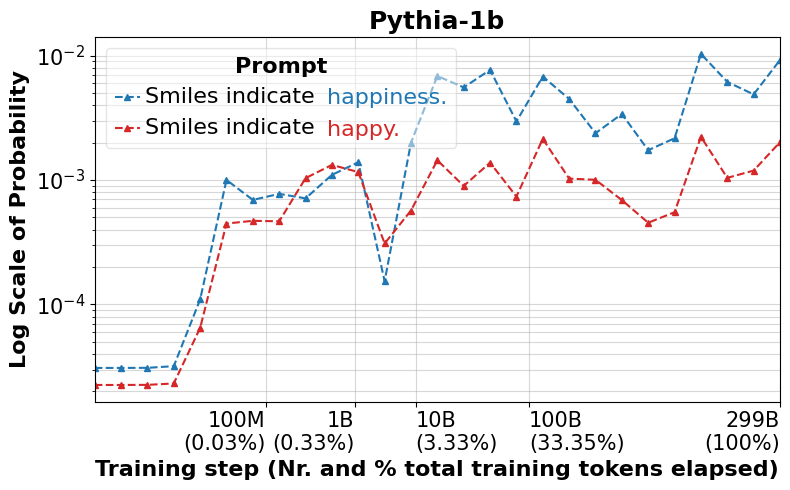}
          \label{fig:appx_morph_eg_pythia_1b_n20}
    \end{subfigure}%
    \begin{subfigure}{.325\textwidth}
          \centering
          \includegraphics[width=\linewidth, trim={1cm 0 0 0}, clip]{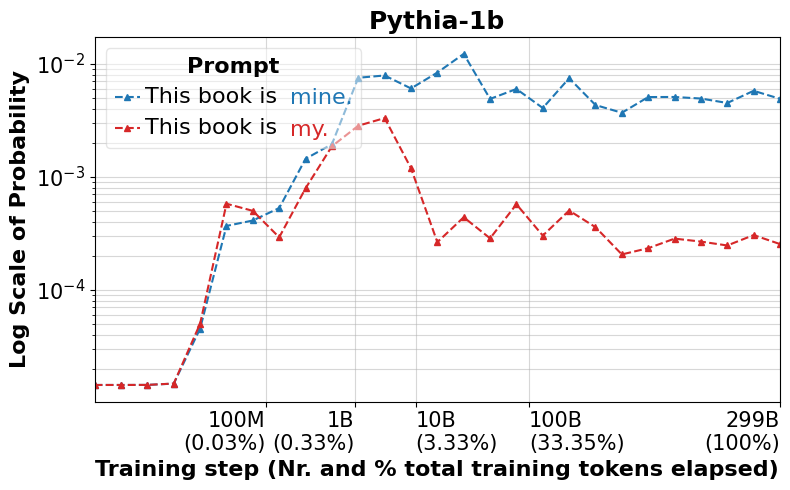}
          \label{fig:appx_morph_eg_pythia_1b_p2}
    \end{subfigure}
    \\
    \vspace{-0.3cm}
    \begin{subfigure}{.34\textwidth}
          \centering
          \includegraphics[width=\linewidth]{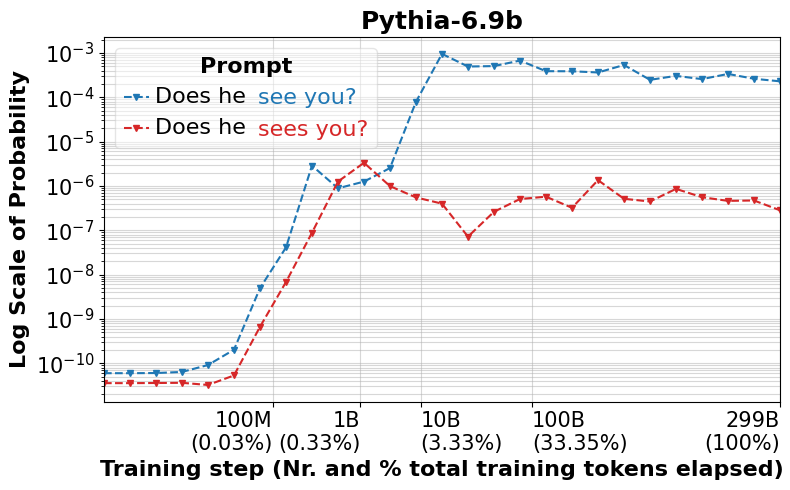}
          \caption{Does he \textcolor{blue}{see you?}}
          \label{fig:appx_morph_eg_pythia_6.9b_v10}
    \end{subfigure}%
    \begin{subfigure}{.325\textwidth}
          \centering
          \includegraphics[width=\linewidth, trim={1cm 0 0 0}, clip]{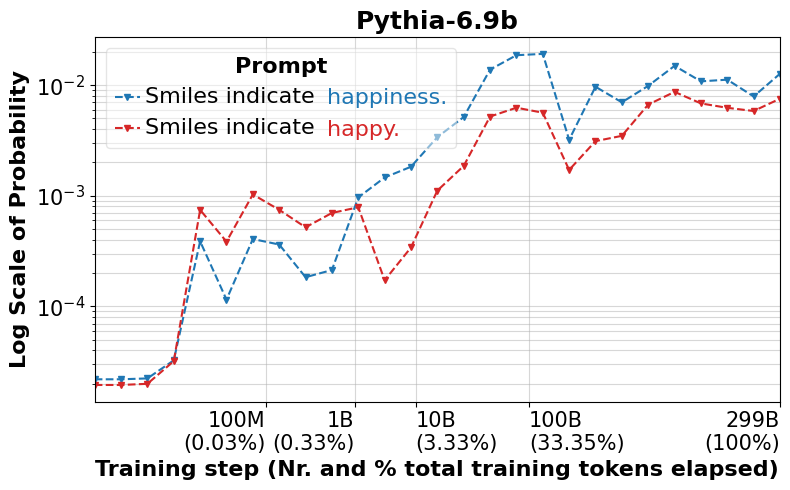}
          \caption{Smiles indicate \textcolor{blue}{happiness.}}
          \label{fig:appx_morph_eg_pythia_6.9b_n20}
    \end{subfigure}%
    \begin{subfigure}{.325\textwidth}
          \centering
          \includegraphics[width=\linewidth, trim={1cm 0 0 0}, clip]{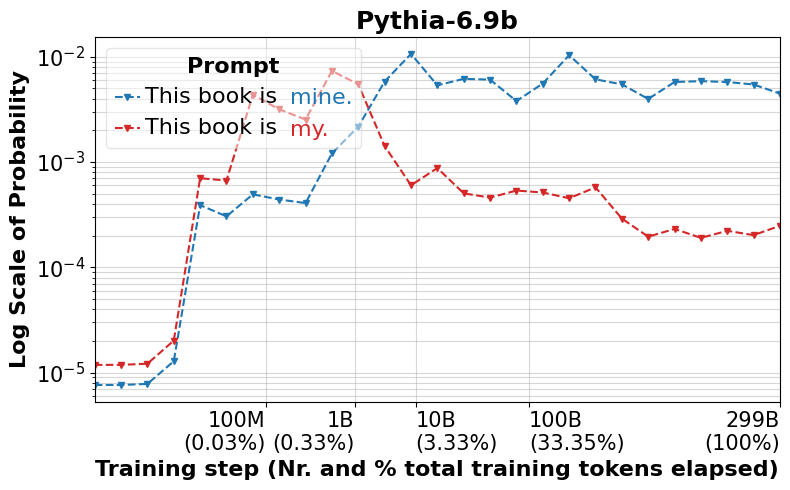}
          \caption{This book is \textcolor{blue}{mine.}}
          \label{fig:appx_morph_eg_pythia_6.9b_p2}
    \end{subfigure}
    
    \caption{Morphological structures across time for Pythia suite: 160M, 410M, 1B, and 6.9B---first to fourth rows, respectively.}
    \label{fig:appx_morph_eg_pythia_suite}
    \vspace{-0.6cm}
\end{figure}
\begin{figure}[H]
    \centering
    \begin{subfigure}{.34\textwidth}
          \centering
          \includegraphics[width=\linewidth]{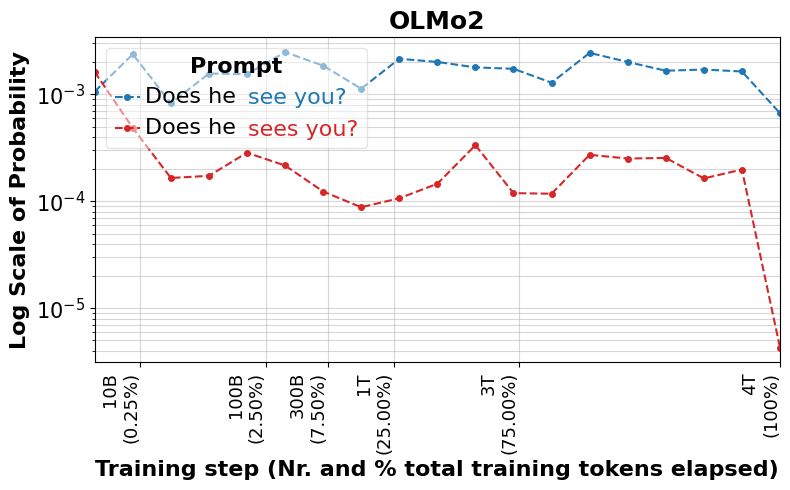}
          \label{fig:appx_morph_eg_olmo2_7B_v10}
    \end{subfigure}%
    \begin{subfigure}{.325\textwidth}
          \centering
          \includegraphics[width=\linewidth, trim={1cm 0 0 0}, clip]{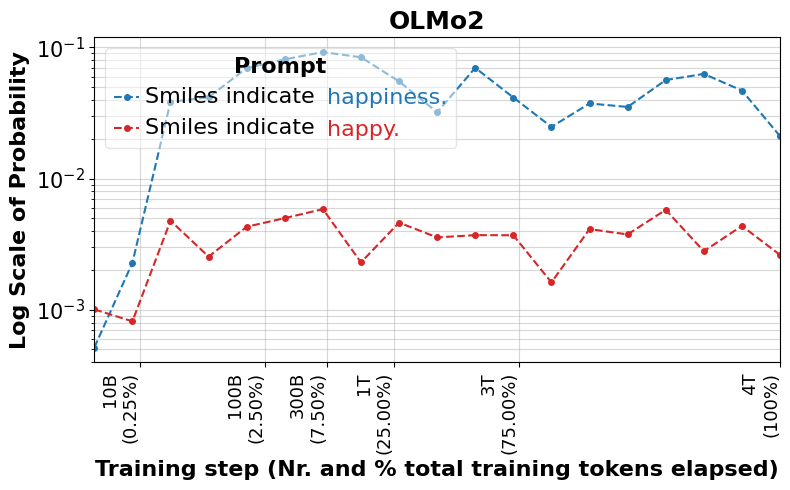}
          \label{fig:appx_morph_eg_olmo2_7B_n20}
    \end{subfigure}%
    \begin{subfigure}{.325\textwidth}
          \centering
          \includegraphics[width=\linewidth, trim={1cm 0 0 0}, clip]{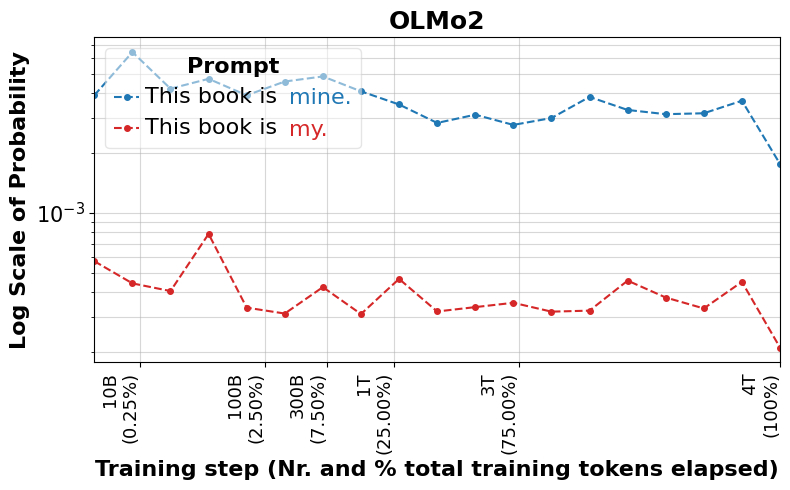}
          \label{fig:appx_morph_eg_olmo2_7B_p2}
    \end{subfigure}
    \\
    \vspace{-0.3cm}
    \begin{subfigure}{.34\textwidth}
          \centering
          \includegraphics[width=\linewidth]{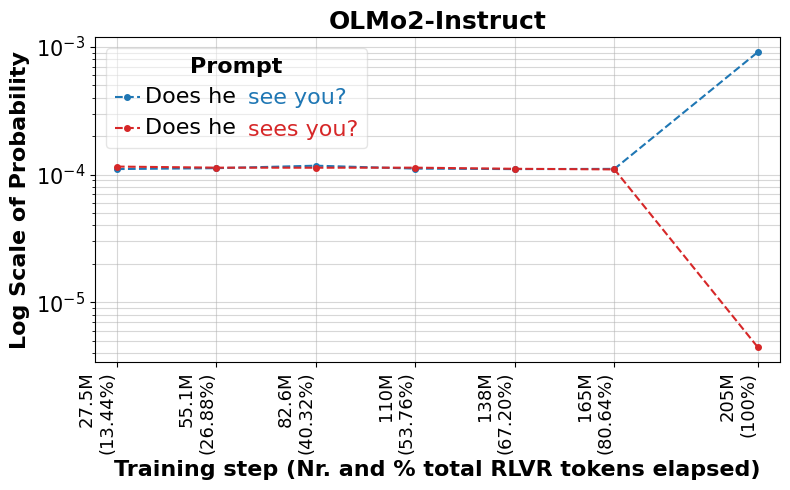}
          \caption{Does he \textcolor{blue}{see you?}}
          \label{fig:appx_morph_eg_olmo2_7B_instruct_v10}
    \end{subfigure}%
    \begin{subfigure}{.325\textwidth}
          \centering
          \includegraphics[width=\linewidth, trim={1cm 0 0 0}, clip]{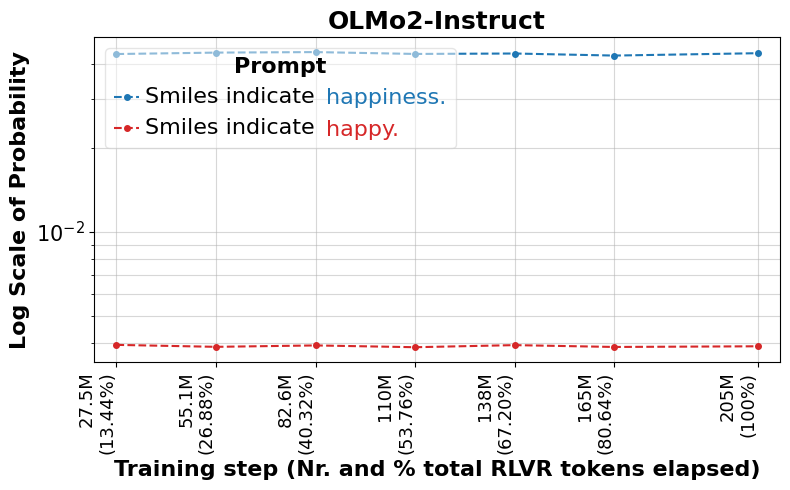}
          \caption{Smiles indicate \textcolor{blue}{happiness.}}
          \label{fig:appx_morph_eg_olmo2_7B_instruct_n20}
    \end{subfigure}%
    \begin{subfigure}{.325\textwidth}
          \centering
          \includegraphics[width=\linewidth, trim={1cm 0 0 0}, clip]{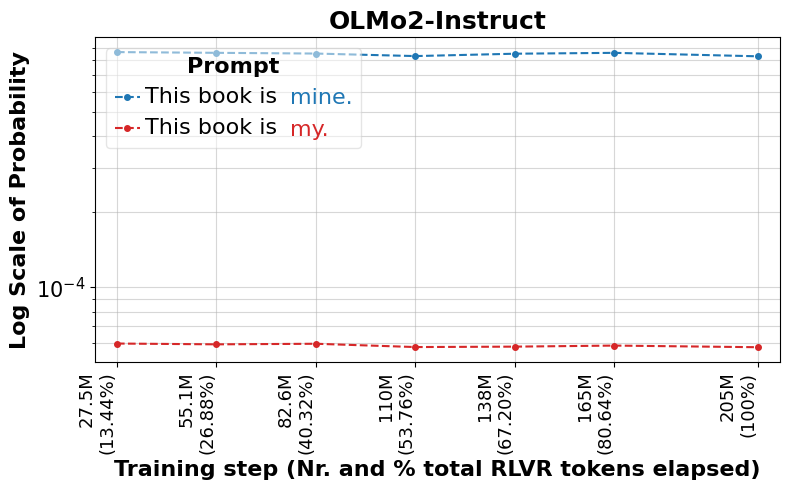}
          \caption{This book is \textcolor{blue}{mine.}}
          \label{fig:appx_morph_eg_olmo2_7B_instruct_p2}
    \end{subfigure}
    \vspace{-0.2cm}
    \caption{Morphological structures across time for \olmoTwo\ (top) and \olmoTwo-Instruct (bottom).}
    \label{fig:appx_morph_eg_olmo2}
    \vspace{-0.5cm}
\end{figure}

\clearpage
\subsection{Tracing Morphological Preferences}\label{sec:appx_morph}
Table \ref{tab:appx_all_morph_eg} summarizes all unit-test cases for morphological structures. In Fig. \ref{fig:appx_morph_main}, we see the three-phase behavior for GPT-2 seeds, where Initialization occurs before 100M tokens; then, the model falls into the Accelerated Learning phase, which continues up to 10B tokens; after that, the model is in the Maturity Plateau phase where the learning is almost converged to its final state. Similar to \olmoThree\ (see Fig. \ref{fig:morph_main_olmo3}), \olmoTwo\ follows a two-phase behavior (a combination of Initialization and Accelerated Learning phases followed by the Maturity Plateau). This is while Fig. \ref{fig:appx_morph_main_olmo2_instruct} states that post-training (instruction) does not play an important role in learning process.

\begin{figure}[H]
    \vspace{-0.6cm}
    \centering
    \vspace{-0.3cm}
    \begin{subfigure}{.34\textwidth}
          \centering
          \includegraphics[width=\linewidth]{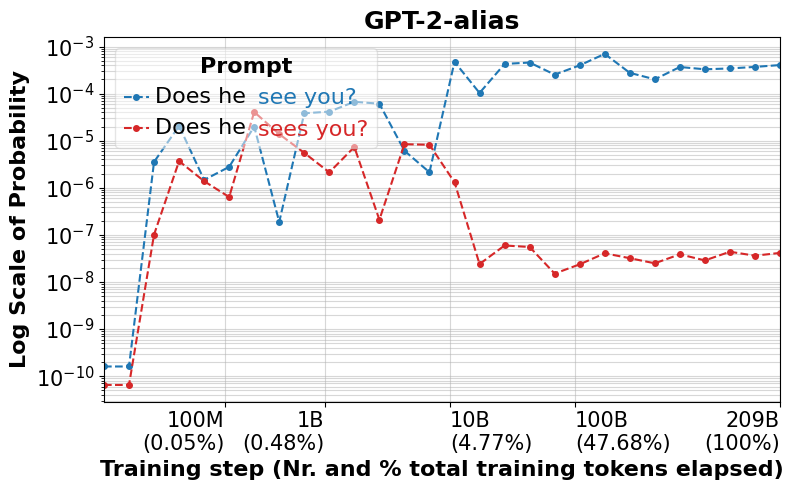}
          \label{fig:appx_morph_eg_gpt2_alias_v10}
    \end{subfigure}%
    \begin{subfigure}{.325\textwidth}
          \centering
          \includegraphics[width=\linewidth, trim={1cm 0 0 0}, clip]{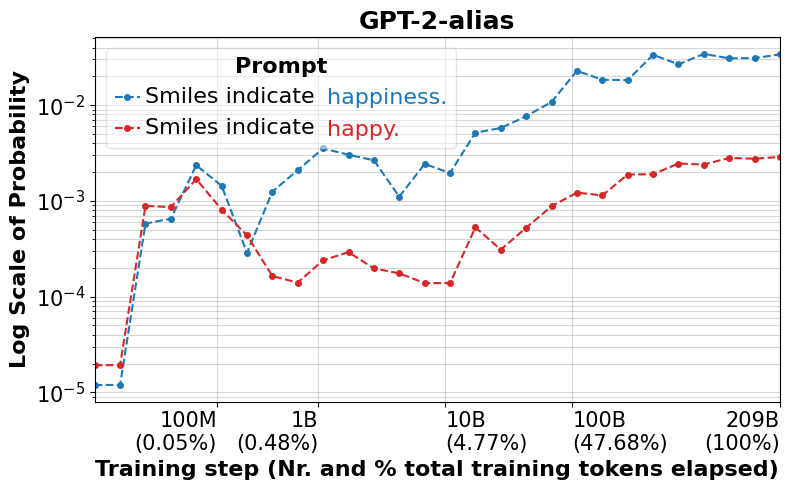}
          \label{fig:appx_morph_eg_gpt2_alias_n20}
    \end{subfigure}%
    \begin{subfigure}{.325\textwidth}
          \centering
          \includegraphics[width=\linewidth, trim={1cm 0 0 0}, clip]{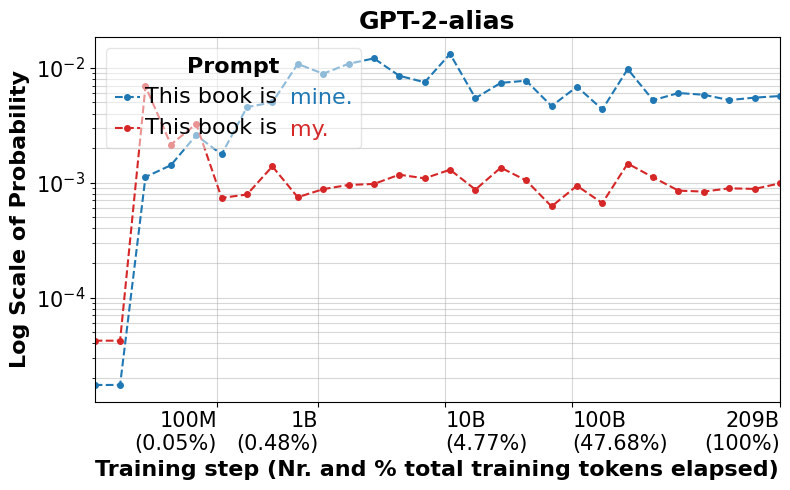}
          \label{fig:appx_morph_eg_gpt2_alias_p2}
    \end{subfigure}
    \\
    \vspace{-0.3cm}
    \begin{subfigure}{.34\textwidth}
          \centering
          \includegraphics[width=\linewidth]{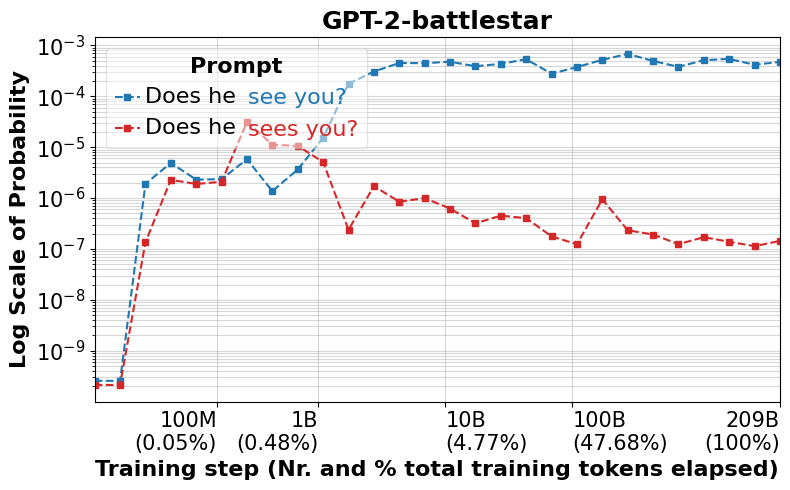}
          \label{fig:appx_morph_eg_gpt2_battlestar_v10}
    \end{subfigure}%
    \begin{subfigure}{.325\textwidth}
          \centering
          \includegraphics[width=\linewidth, trim={1cm 0 0 0}, clip]{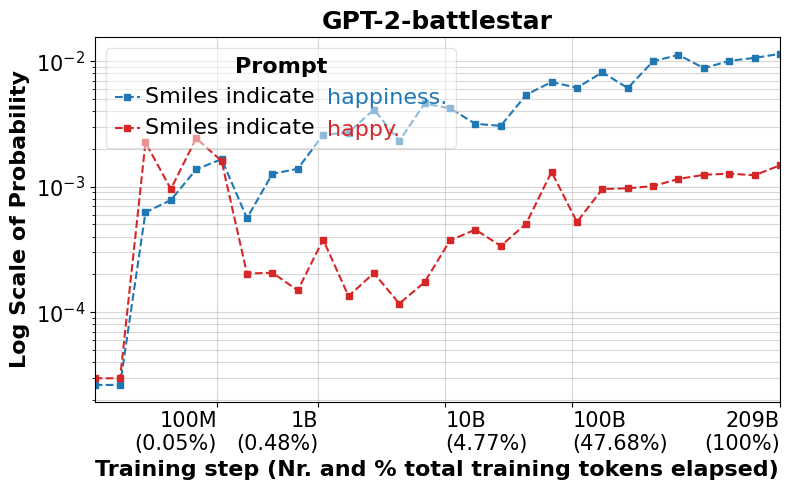}
          \label{fig:appx_morph_eg_gpt2_battlestar_n20}
    \end{subfigure}%
    \begin{subfigure}{.325\textwidth}
          \centering
          \includegraphics[width=\linewidth, trim={1cm 0 0 0}, clip]{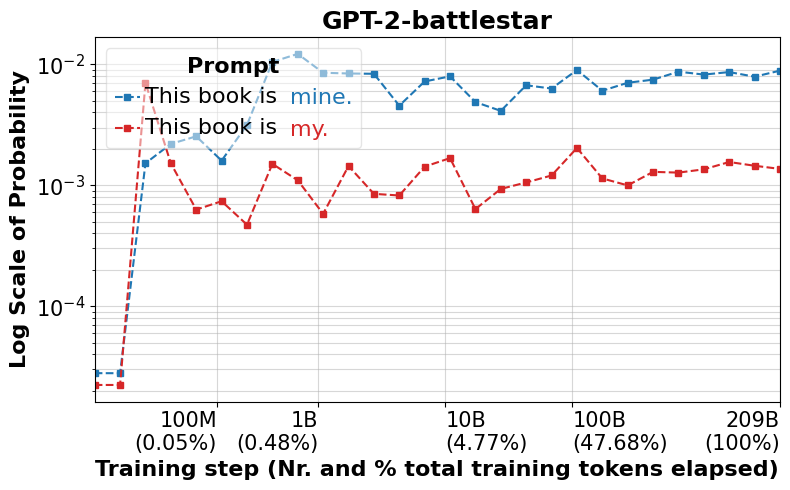}
          \label{fig:appx_morph_eg_gpt2_battlestar_p2}
    \end{subfigure}
    \\
    \vspace{-0.3cm}
    \begin{subfigure}{.34\textwidth}
          \centering
          \includegraphics[width=\linewidth]{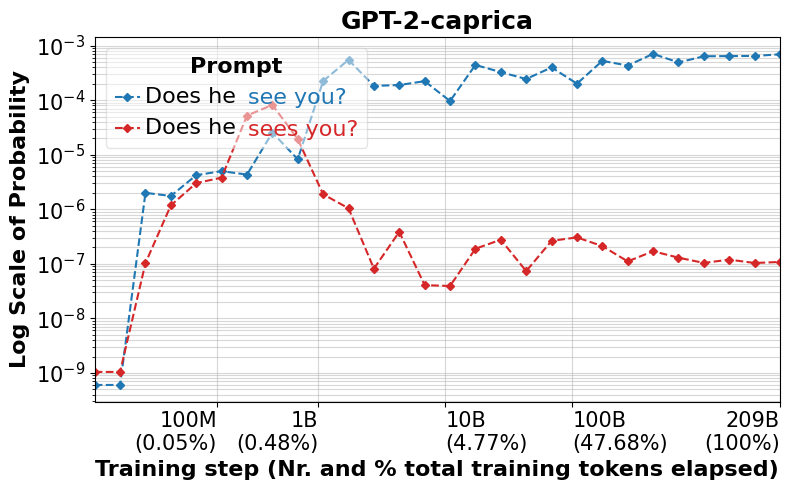}
          \label{fig:appx_morph_eg_gpt2_caprica_v10}
    \end{subfigure}%
    \begin{subfigure}{.325\textwidth}
          \centering
          \includegraphics[width=\linewidth, trim={1cm 0 0 0}, clip]{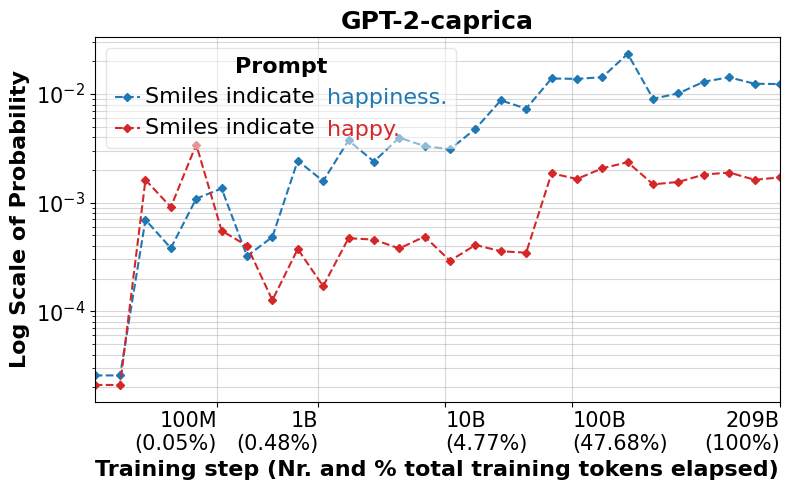}
          \label{fig:appx_morph_eg_gpt2_caprica_n20}
    \end{subfigure}%
    \begin{subfigure}{.325\textwidth}
          \centering
          \includegraphics[width=\linewidth, trim={1cm 0 0 0}, clip]{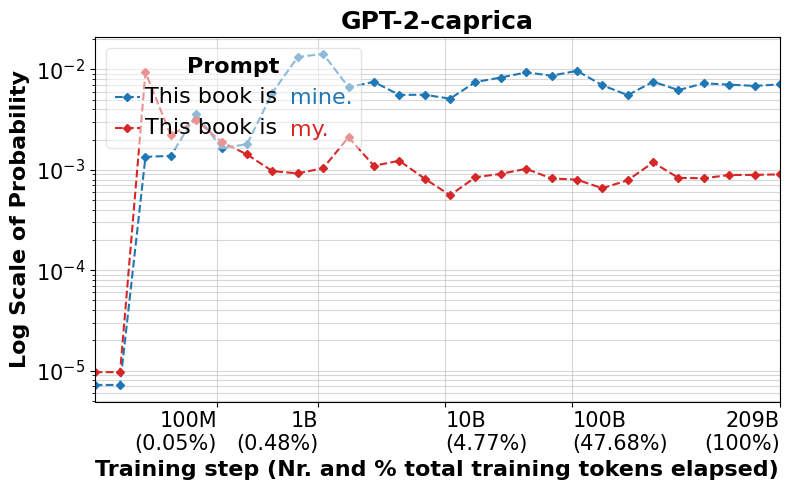}
          \label{fig:appx_morph_eg_gpt2_caprica_p2}
    \end{subfigure}
    \\
    \vspace{-0.3cm}
    \begin{subfigure}{.34\textwidth}
          \centering
          \includegraphics[width=\linewidth]{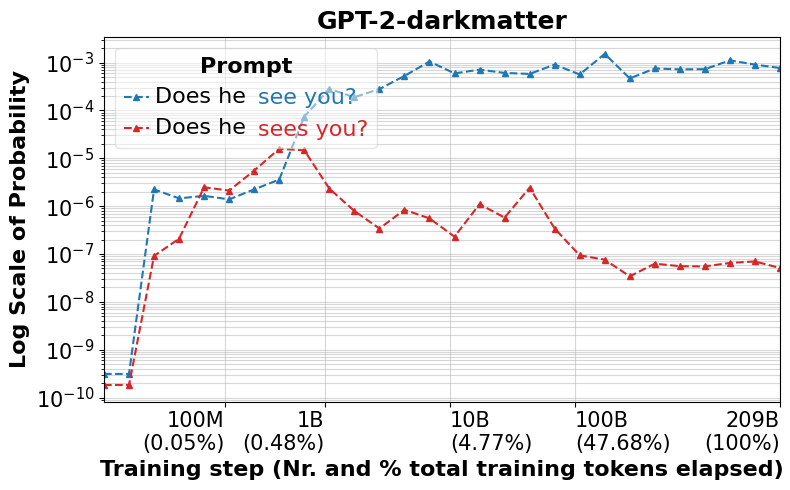}
          \label{fig:appx_morph_eg_gpt2_darkmatter_v10}
    \end{subfigure}%
    \begin{subfigure}{.325\textwidth}
          \centering
          \includegraphics[width=\linewidth, trim={1cm 0 0 0}, clip]{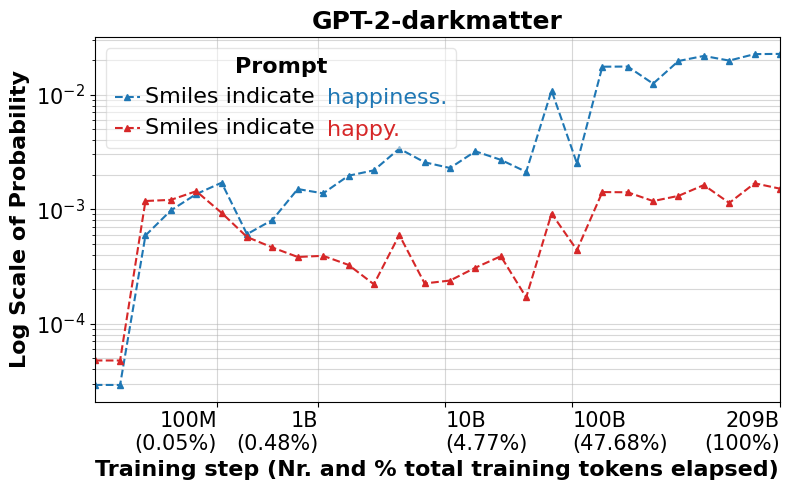}
          \label{fig:appx_morph_eg_gpt2_darkmatter_n20}
    \end{subfigure}%
    \begin{subfigure}{.325\textwidth}
          \centering
          \includegraphics[width=\linewidth, trim={1cm 0 0 0}, clip]{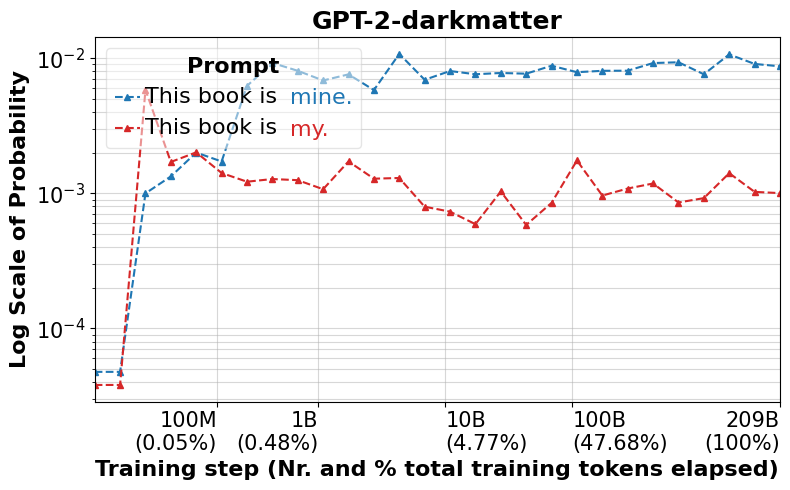}
          \label{fig:appx_morph_eg_gpt2_darkmatter_p2}
    \end{subfigure}
    \\
    \vspace{-0.3cm}
    \begin{subfigure}{.34\textwidth}
          \centering
          \includegraphics[width=\linewidth]{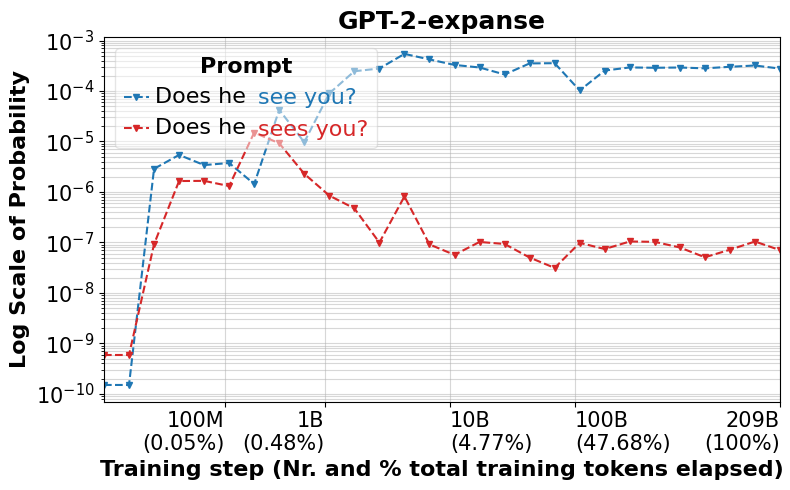}
          \caption{Does he \textcolor{blue}{see you?}}
          \label{fig:appx_morph_eg_gpt2_expanse_v10}
    \end{subfigure}%
    \begin{subfigure}{.325\textwidth}
          \centering
          \includegraphics[width=\linewidth, trim={1cm 0 0 0}, clip]{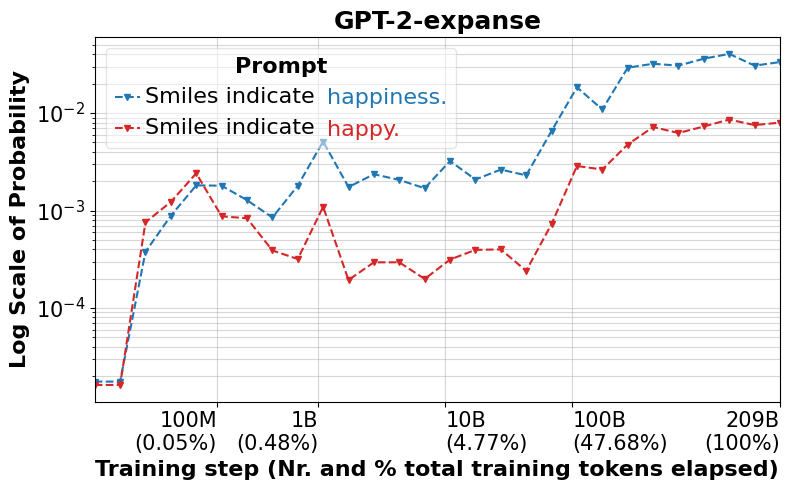}
          \caption{Smiles indicate \textcolor{blue}{happiness.}}
          \label{fig:appx_morph_eg_gpt2_expanse_n20}
    \end{subfigure}%
    \begin{subfigure}{.325\textwidth}
          \centering
          \includegraphics[width=\linewidth, trim={1cm 0 0 0}, clip]{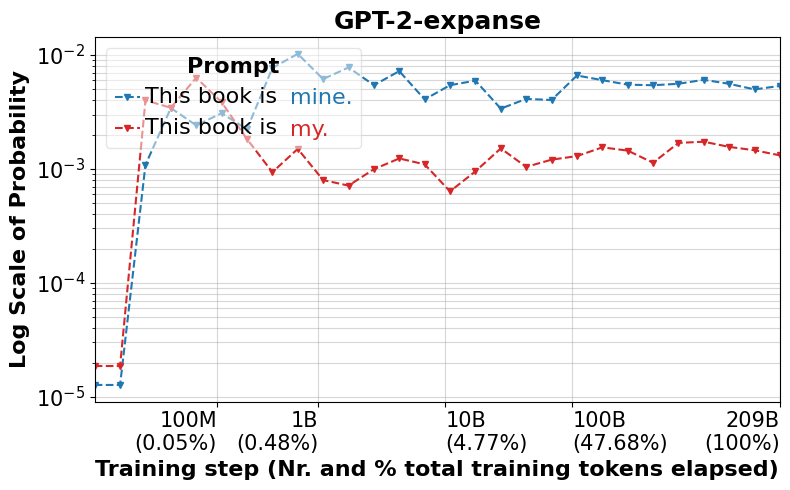}
          \caption{This book is \textcolor{blue}{mine.}}
          \label{fig:appx_morph_eg_gpt2_expanse_p2}
    \end{subfigure}
    \caption{Morphological structures across time for GPT-2 variations: Alias, Battlestar, Caprica, Darkmatter, and Expanse---first to fifth rows, respectively.}
    \label{fig:appx_morph_eg_gpt2}
    \vspace{-0.5cm}
\end{figure}

We further consider unit-test cases mentioned in Fig. \ref{fig:morph_eg}.
Figures \ref{fig:appx_morph_eg_pythia_suite}-\ref{fig:appx_morph_eg_gpt2} show the performance of the Pythia suite, the GPT-2 seeds, and the \olmoTwo\ suite for the cases of morphological structures mentioned above, respectively.

Similar to Pythia-70M, Pythia-160M results in higher probabilities for unfavorable targets for cases in Figs. \ref{fig:appx_morph_eg_pythia_6.9b_v10} and \ref{fig:appx_morph_eg_pythia_6.9b_n20}. As expected, this trend changes for larger models (410M, 1B, and 6.9B), even the gap between favorable and unfavorable targets increases following enlarging the scales; however, Fig. \ref{fig:appx_morph_eg_pythia_6.9b_n20} shows that Pythia-6.9B is not successful in prioritizing the favorable target, ``happiness'', over ``happy''. Comparing Pythia-6.9B and Pythia-12B (see Fig. \ref{fig:morph_eg}), the results highlight the architectural mastery of Pythia-12B that leads to a more rigid internal grasp of English syntax than Pythia-6.9B. Furthermore, the results underscore a fascinating aspect of LLM training: \textbf{emergence, scale, and training trajectories are non-linear}. This phenomenon is known as \textbf{U-shaped learning} \cite{ushape} in neural networks: good performance followed by bad performance followed by good performance once again. This is translated to: the smaller models (410M and 1B) perform better than Pythia-6.9B. However, the performance is recovered by Pythia-12B. Smaller models heavily weight literal pattern matching found in dictionary definitions or formal encyclopedia text within the training data, whereas medium-sized models (Pythia-6.9B) tend to absorb broader web data, colloquial expressions, and association heuristics. As a result, the massive co-occurrence of ``smiles'' and ``happy'' across standard web text might distract or mislead the middle-sized models. When a model is large enough (Pythia-12B), it reconciles these forces, by synthesizing both the formal grammatical rule and the raw data correlation to results in the correct answer. Figure \ref{fig:appx_morph_eg_pythia_6.9b_p2} includes the results for the incipit ``This book is''. Following the assessments mentioned in Sec. \ref{sec:morph_per_lex}, that emphasizes the impact of frequency on LLMs performance, all models prioritize the highly repeated combination ``is my'' over rare phrase ``is mine'' in their training data; thus, they fail to learn the morphological structure.

Figures \ref{fig:appx_morph_eg_olmo2} and \ref{fig:appx_morph_eg_gpt2} illustrate the corresponding results for different GPT-2 and \olmoTwo\ variations, respectively. While all GPT-2 variations are successful in learning morphological structures of Nouns (Fig. \ref{fig:appx_morph_eg_gpt2_expanse_n20}), some (Battlestar, Caprica, and Darkmatter) fail the test (Fig. \ref{fig:appx_morph_eg_gpt2_expanse_v10}). \olmoTwo\ variations are also successful in Verbs and Nouns cases, except \olmoTwo-Instruct that is misled in the Verbs case. Similar to Pythia suite, all GPT-2 and \olmoTwo\ variations underperform the Pronouns case (Fig. \ref{fig:appx_morph_eg_gpt2_expanse_p2}). Indeed, \olmoTwo\ has been trained on 4.05-T tokens of raw web text, books, academic papers, and code, whereas \olmoTwo-Instruct is derived from \olmoTwo\ through several post-training stages using the T\"ulu-3 recipe. Unlike \olmoTwo that sees the third-person verbs frequently in its training data, \olmoTwo-Instruct is a chat-based model where rarely feature isolated narrative fragments starting with the third-person or a third person pronoun is followed by auxiliary verbs.

\begin{figure}[H]
    \centering
    \begin{subfigure}{.48\textwidth}
          \centering
          \includegraphics[width=\linewidth]{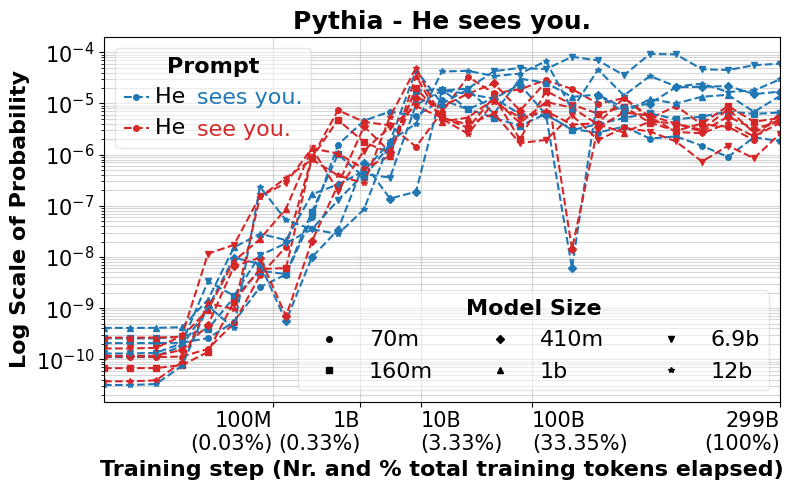}
          \caption{He \textcolor{blue}{sees you?}}
          \label{fig:appx_morph_except_v12}
    \end{subfigure}%
    \begin{subfigure}{.46\textwidth}
          \centering
          \includegraphics[width=\linewidth, trim={1cm 0 0 0}, clip]{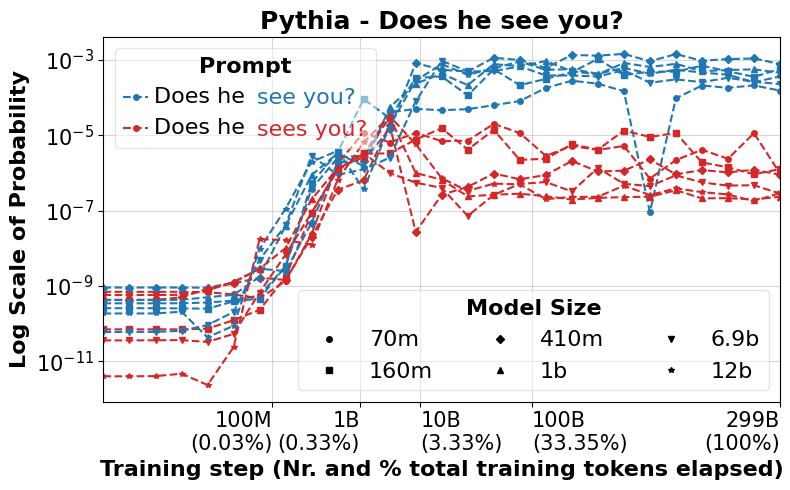}
          \caption{Does he \textcolor{blue}{see you?}}
          \label{fig:appx_morph_except_v10}
    \end{subfigure}\\
    
    \caption{Training corpus strongly affects morphological structure.}
    \label{fig:appx_morph_except_pythia}
\end{figure}

\textbf{Training data shape the morphological structure.} Figure \ref{fig:appx_morph_except_pythia} indicates two correct morphological structures, ``He sees you.'' and ``Does he see you?''. Small Pythia models (70M and 160M) fail to learn the correct structure for the first case. This failure can happen because ``Does he see you?'' appears more often in the training data than ``He sees you.''.

\clearpage


\begin{figure}[H]
    \vspace{-0.5cm}
    \centering
    \setlength{\tabcolsep}{0pt}
    \begin{tabular}{cl}
        \centering
        {
        \renewcommand{\arraystretch}{0.6} 
        \begin{tabular}{@{}l@{}}
            \vspace{-0.4cm}
             \begin{subfigure}{.75\textwidth}
                  \centering
                  \includegraphics[width=\linewidth, trim={0 9.5cm 0 0}, clip]{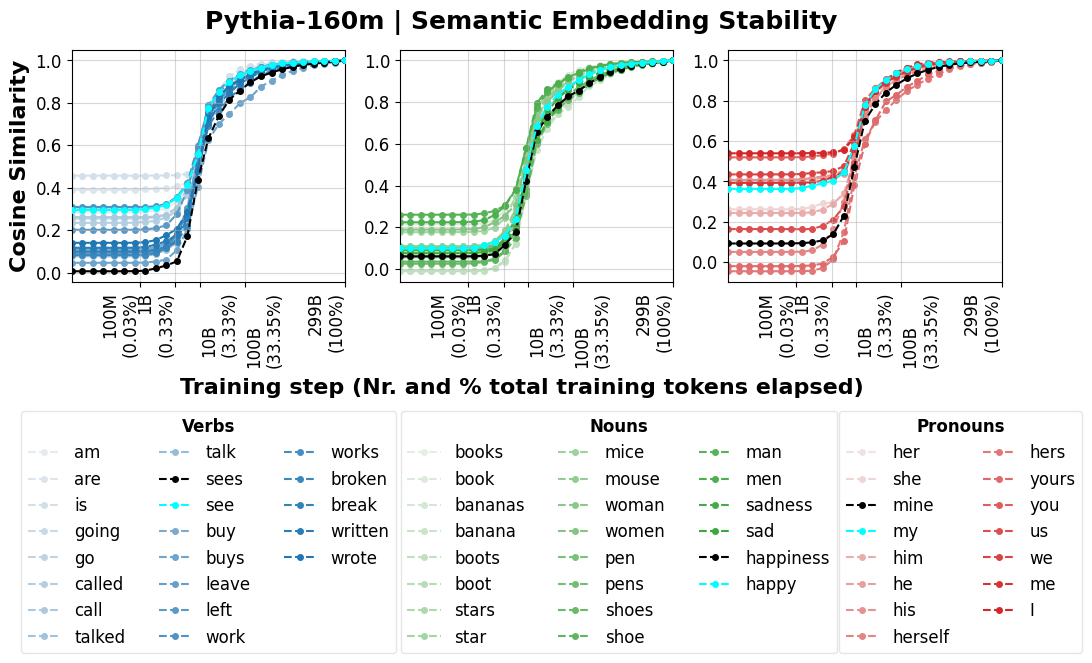}
                  \label{fig:embedding_pythia_160m}
            \end{subfigure}\\
            \vspace{-0.4cm}
             \begin{subfigure}{.75\textwidth}
                  \centering
                  \includegraphics[width=\linewidth, trim={0 9.5cm 0 0.2cm}, clip]{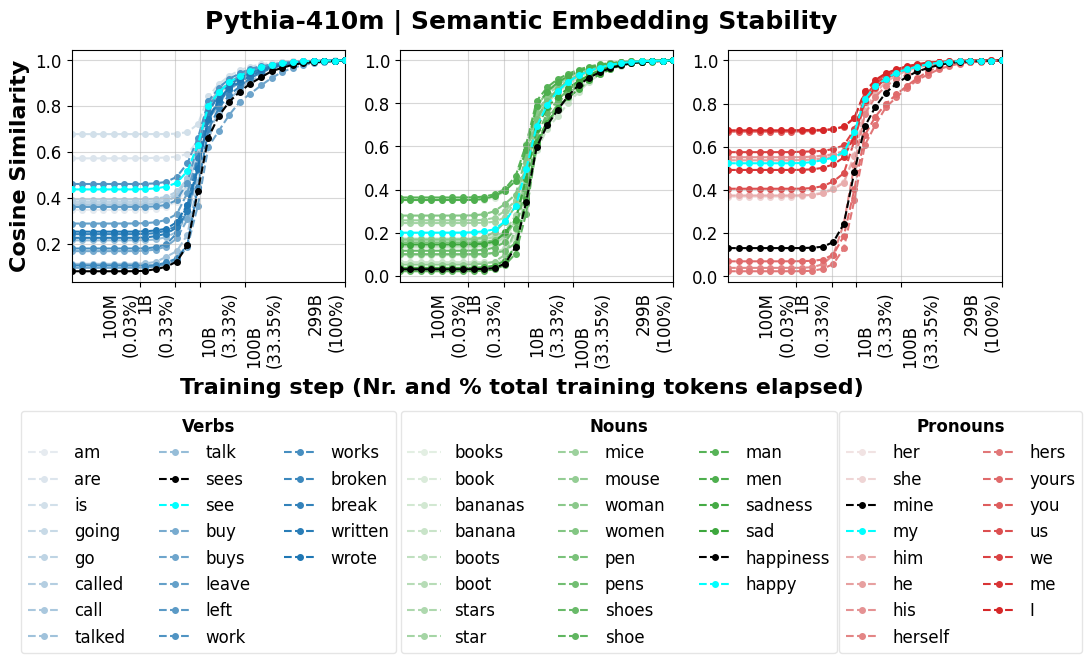}
                  \label{fig:embedding_pythia_410m}
            \end{subfigure}\\
            \vspace{-0.4cm}
            \hspace{-.04cm}
            \begin{subfigure}{.75\textwidth}
                  \centering
                  \includegraphics[width=\linewidth, , trim={0 9.5cm 0 0.2cm}, clip]{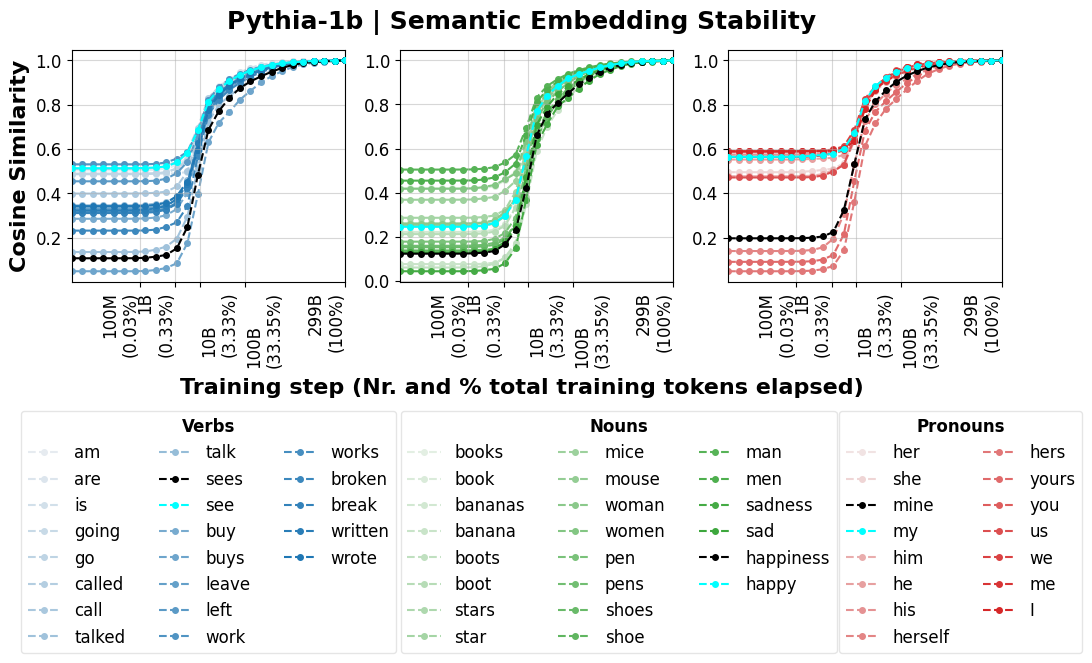}
                  \label{fig:embedding_pythia_1b}
            \end{subfigure}\\
            \hspace{-.04cm}
            \begin{subfigure}{.75\textwidth}
                  \centering
                  \includegraphics[width=\linewidth, , trim={0 6.5cm 0 0.2cm}, clip]{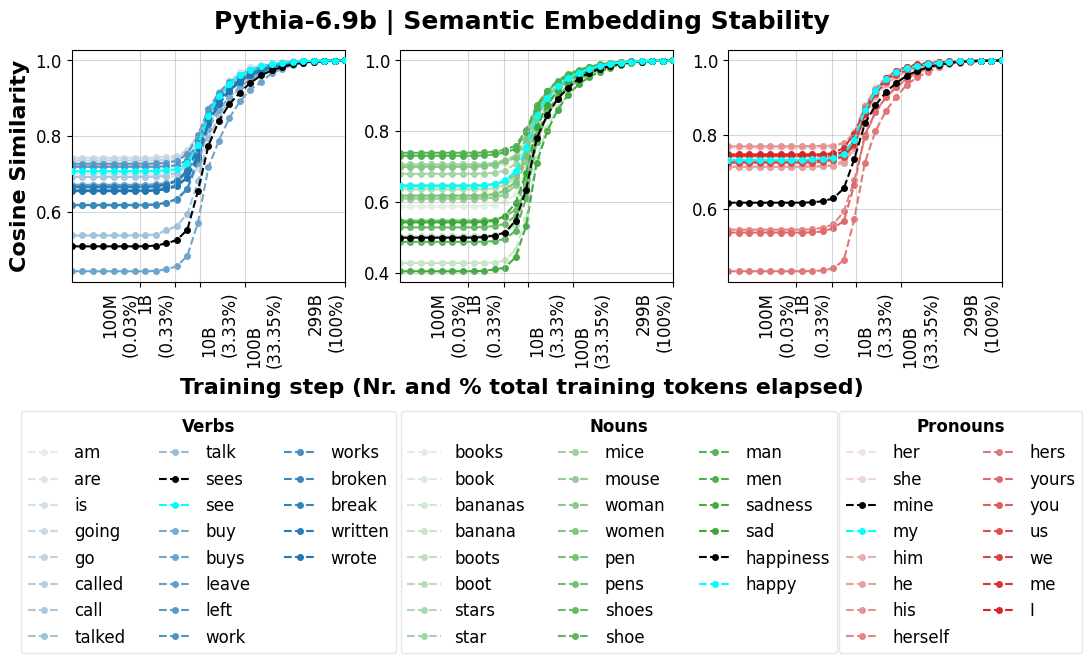}
                  \label{fig:embedding_pythia_6.9b}
            \end{subfigure}
        \end{tabular}
        }
        &
        {
        \renewcommand{\arraystretch}{0.6} 
        \begin{tabular}{@{}l@{}}
            \hspace{-1.1cm}
            \begin{subfigure}{.27\textwidth}
                  \centering
                  \includegraphics[width=\linewidth, trim={0 0cm 17.6cm 10.4cm}, clip]{images/embedding/pythia_70m_embed_sim.png}
            \end{subfigure} \\
            \hspace{-0.9cm}
            \begin{subfigure}{.27\textwidth}
                  \centering
                  \includegraphics[width=\linewidth, trim={10.15cm 0cm 6.5cm 10.4cm}, clip]{images/embedding/pythia_70m_embed_sim.png}
            \end{subfigure}\\
            \hspace{-0.9cm}
            \begin{subfigure}{.2\textwidth}
                  \centering
                  \includegraphics[width=\linewidth, trim={21.25cm 0cm 0cm 10.4cm}, clip]{images/embedding/pythia_70m_embed_sim.png}
            \end{subfigure}
        \end{tabular}
        }
    \vspace{-0.5cm}
    \end{tabular}
    \caption{Semantic embedding stability for Pythia suite: 160M, 410M, 1B, and 6.9B---first to fourth rows, respectively; left (Verbs), middle (Nouns), and right (Pronouns).}
    \label{fig:appx_embedding_pythia}
    \vspace{-0.6cm}
\end{figure}
\begin{figure}[H]
    \vspace{-0.3cm}
    \centering
    \setlength{\tabcolsep}{0pt}
    \begin{tabular}{cl}
        \centering
        {
        \renewcommand{\arraystretch}{0.6} 
        \begin{tabular}{@{}l@{}}
            \vspace{-0.4cm}
             \begin{subfigure}{.75\textwidth}
                  \centering
                  \includegraphics[width=\linewidth, trim={0 7.5cm 0 0}, clip]{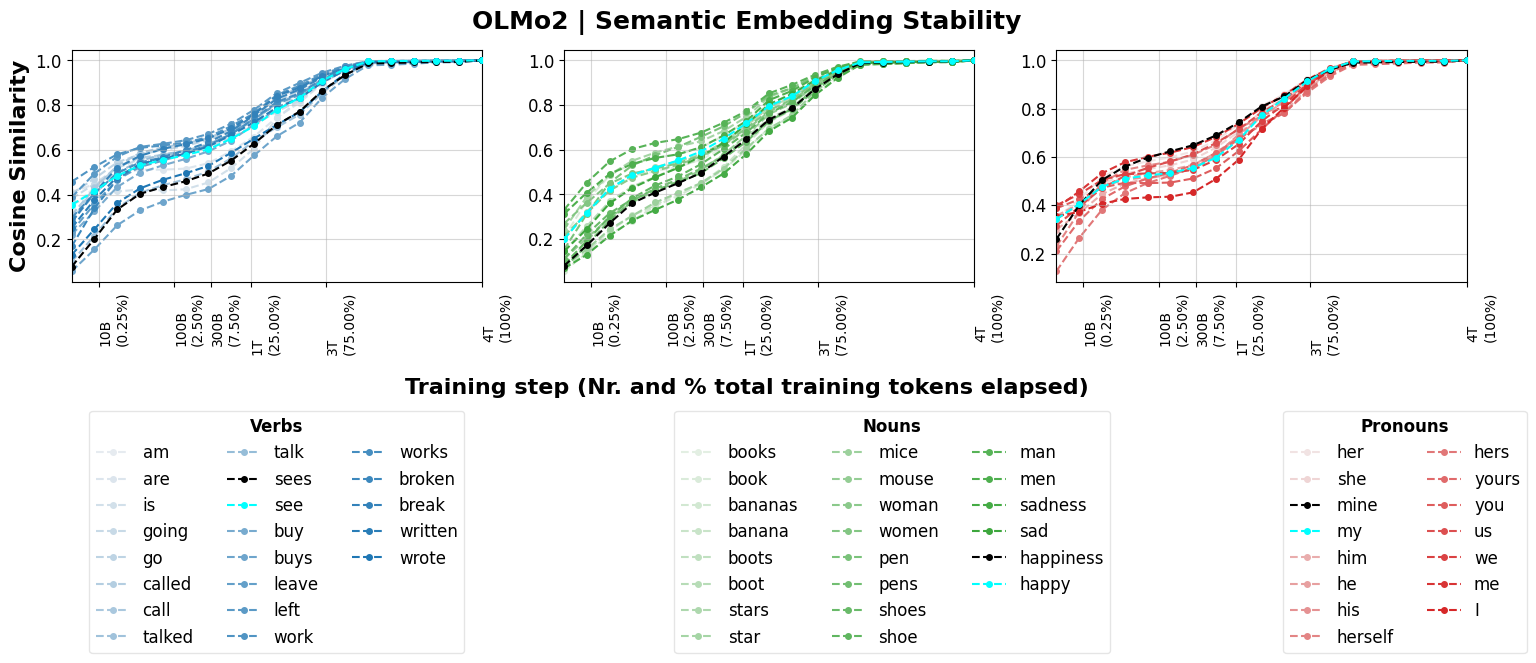}
                  \label{fig:embedding_olmo2}
            \end{subfigure}\\
            \hspace{-.04cm}
            \begin{subfigure}{.75\textwidth}
                  \centering
                  \includegraphics[width=\linewidth, , trim={0 6.5cm 0 0.2cm}, clip]{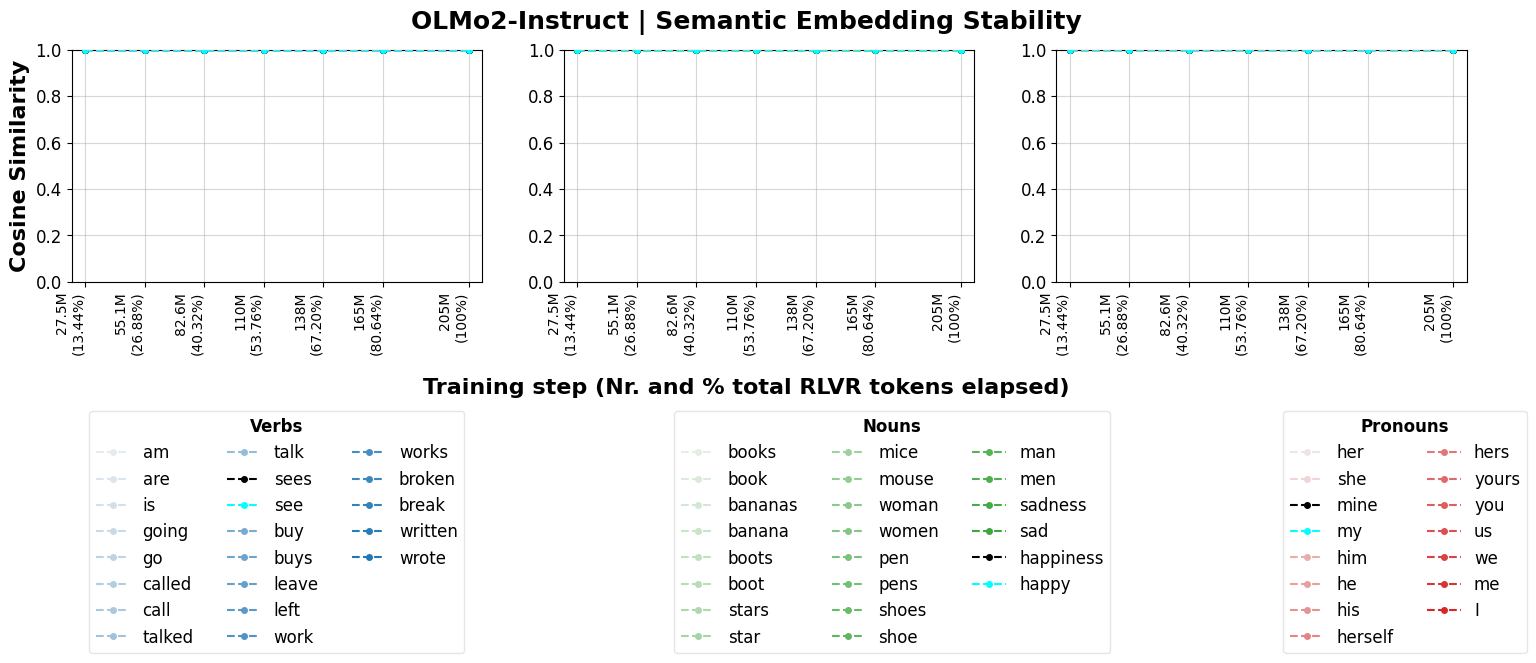}
                  \label{fig:embedding_olmo2_instruct}
            \end{subfigure}
        \end{tabular}
        }
        &
        {
        \renewcommand{\arraystretch}{0.6} 
        \begin{tabular}{@{}l@{}}
            \hspace{-0.75cm}
            \begin{subfigure}{.27\textwidth}
                  \centering
                  \includegraphics[width=\linewidth, trim={0 0cm 17.6cm 10.4cm}, clip]{images/embedding/pythia_70m_embed_sim.png}
            \end{subfigure} \\
            \hspace{-0.55cm}
            \begin{subfigure}{.27\textwidth}
                  \centering
                  \includegraphics[width=\linewidth, trim={10.15cm 0cm 6.5cm 10.4cm}, clip]{images/embedding/pythia_70m_embed_sim.png}
            \end{subfigure}\\
            \hspace{-0.55cm}
            \begin{subfigure}{.2\textwidth}
                  \centering
                  \includegraphics[width=\linewidth, trim={21.25cm 0cm 0cm 10.4cm}, clip]{images/embedding/pythia_70m_embed_sim.png}
            \end{subfigure}
        \end{tabular}
        }
    \vspace{-0.3cm}
    \end{tabular}
    \caption{Semantic embedding stability for \olmoTwo\ and \olmoTwo-Instruct: left (Verbs), middle (Nouns), and right (Pronouns).}
    \label{fig:appx_embedding_olmo2}
\end{figure}
\begin{figure}[H]
    \vspace{-0.5cm}
    \centering
    \setlength{\tabcolsep}{0pt}
    \begin{tabular}{cl}
        \centering
        {
        \renewcommand{\arraystretch}{0.6} 
        \begin{tabular}{@{}l@{}}
            \vspace{-0.4cm}
             \begin{subfigure}{.75\textwidth}
                  \centering
                  \includegraphics[width=\linewidth, trim={0 9.5cm 0 0}, clip]{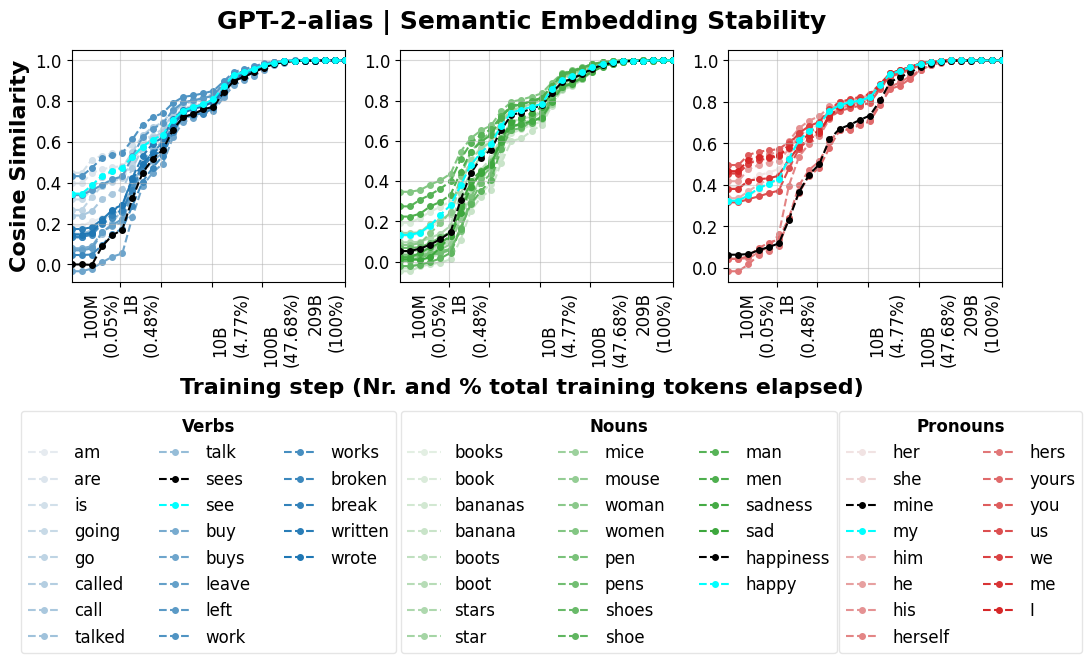}
                  \label{fig:embedding_gpt2_alias}
            \end{subfigure}\\
            \vspace{-0.4cm}
             \begin{subfigure}{.75\textwidth}
                  \centering
                  \includegraphics[width=\linewidth, trim={0 9.5cm 0 0.2cm}, clip]{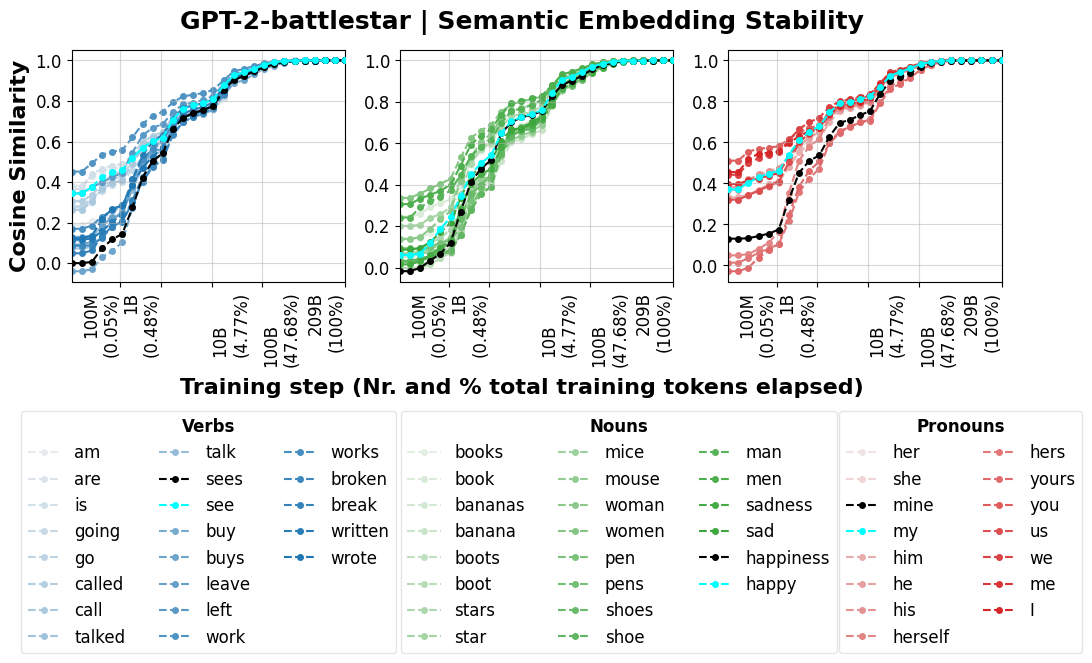}
                  \label{fig:embedding_gpt2_battlestar}
            \end{subfigure}\\
            \vspace{-0.4cm}
            \hspace{-.04cm}
            \begin{subfigure}{.75\textwidth}
                  \centering
                  \includegraphics[width=\linewidth, , trim={0 9.5cm 0 0.2cm}, clip]{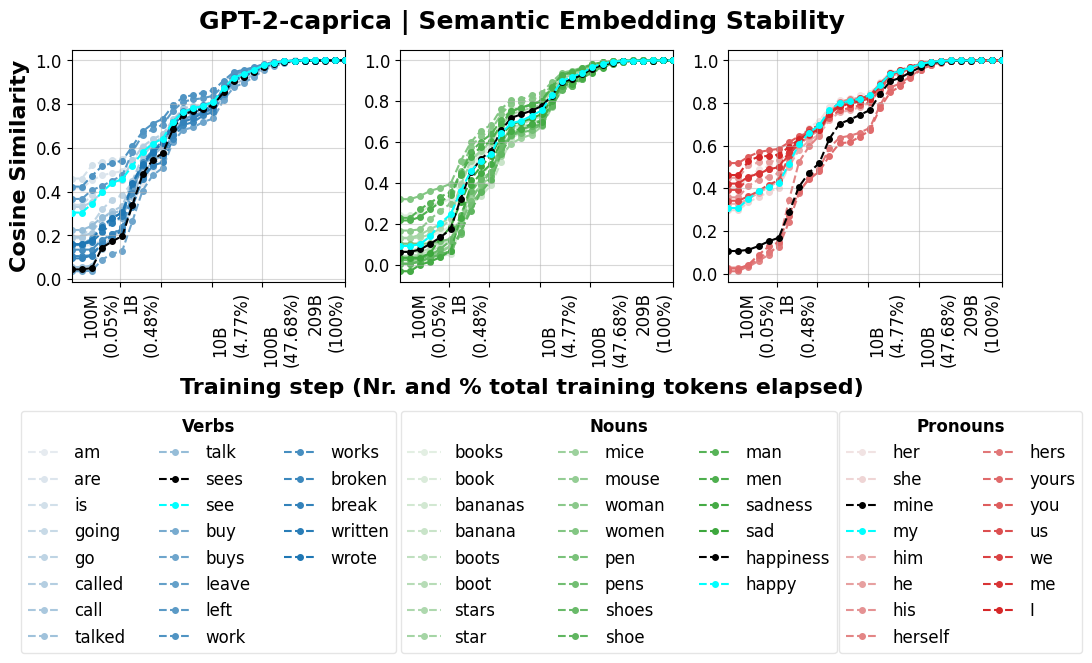}
                  \label{fig:embedding_gpt2_caprica}
            \end{subfigure}\\
            \vspace{-0.4cm}
            \hspace{-.04cm}
            \begin{subfigure}{.75\textwidth}
                  \centering
                  \includegraphics[width=\linewidth, , trim={0 9.5cm 0 0.2cm}, clip]{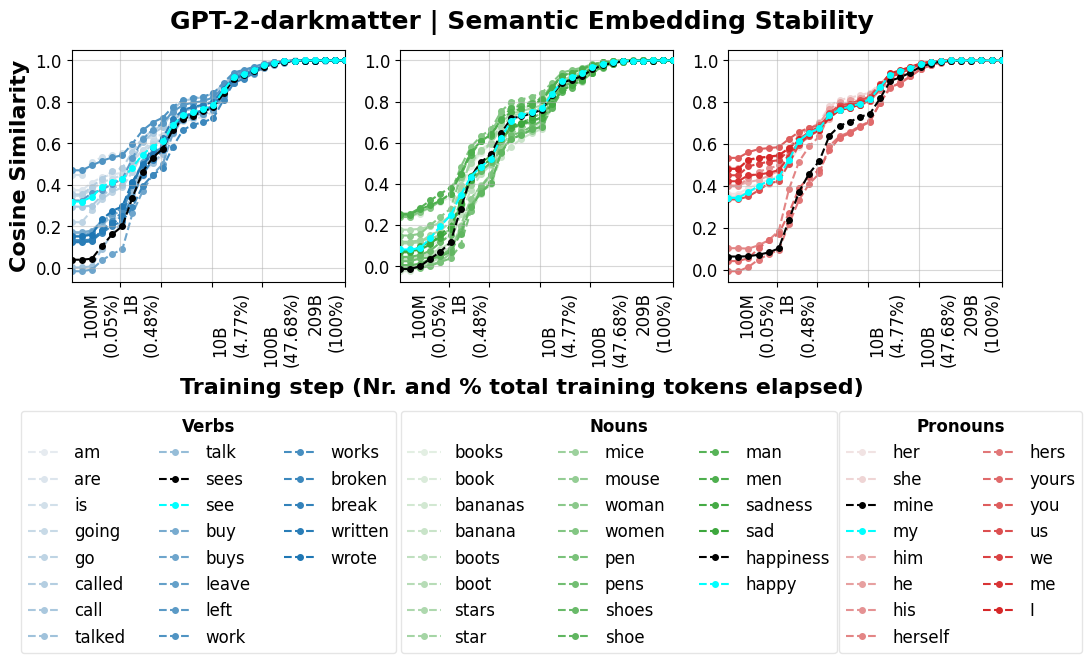}
                  \label{fig:embedding_darkmatter}
            \end{subfigure}\\
            \hspace{-.04cm}
            \begin{subfigure}{.75\textwidth}
                  \centering
                  \includegraphics[width=\linewidth, , trim={0 6.5cm 0 0.2cm}, clip]{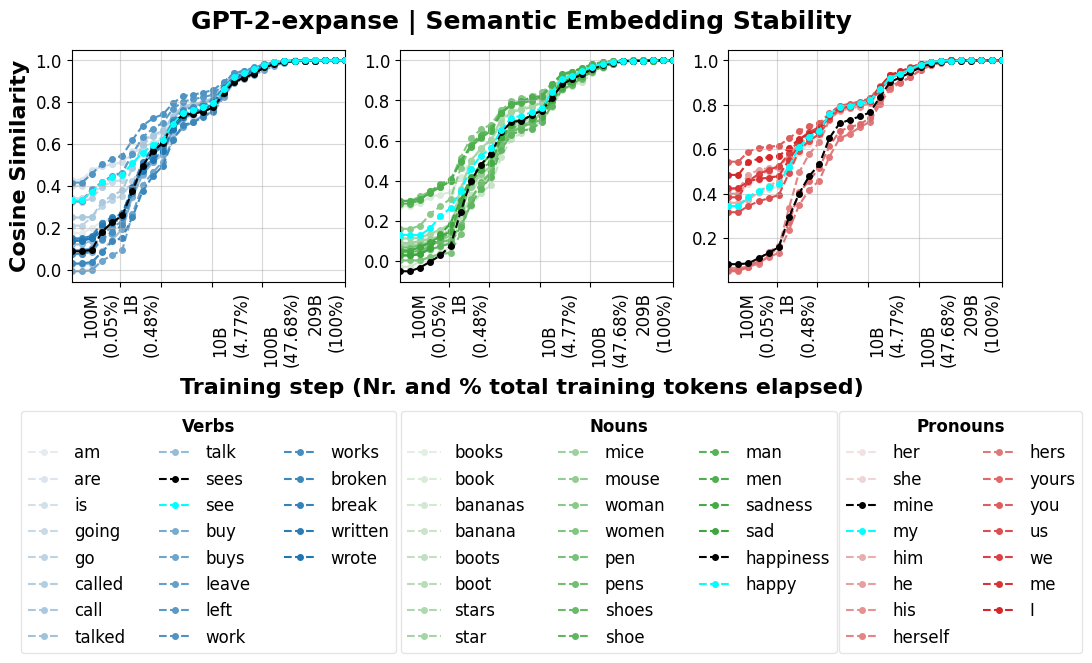}
                  \label{fig:embedding_expanse}
            \end{subfigure}
        \end{tabular}
        }
        &
        {
        \renewcommand{\arraystretch}{0.6} 
        \begin{tabular}{@{}l@{}}
            \hspace{-1.1cm}
            \begin{subfigure}{.27\textwidth}
                  \centering
                  \includegraphics[width=\linewidth, trim={0 0cm 17.6cm 10.4cm}, clip]{images/embedding/pythia_70m_embed_sim.png}
            \end{subfigure} \\
            \hspace{-0.9cm}
            \begin{subfigure}{.27\textwidth}
                  \centering
                  \includegraphics[width=\linewidth, trim={10.15cm 0cm 6.5cm 10.4cm}, clip]{images/embedding/pythia_70m_embed_sim.png}
            \end{subfigure}\\
            \hspace{-0.9cm}
            \begin{subfigure}{.2\textwidth}
                  \centering
                  \includegraphics[width=\linewidth, trim={21.25cm 0cm 0cm 10.4cm}, clip]{images/embedding/pythia_70m_embed_sim.png}
            \end{subfigure}
        \end{tabular}
        }
    \vspace{-0.5cm}
    \end{tabular}
    \caption{Semantic embedding stability for GPT-2 variations: Alias, Battlestar, Caprica, Darkmatter, and Expanse---first to fifth rows, respectively; left (Verbs), middle (Nouns), and right (Pronouns).}
    \label{fig:appx_embedding_gpt2}
    \vspace{-0.5cm}
\end{figure}

\clearpage
\begin{figure}[H]
    \centering
    \includegraphics[width=\linewidth]{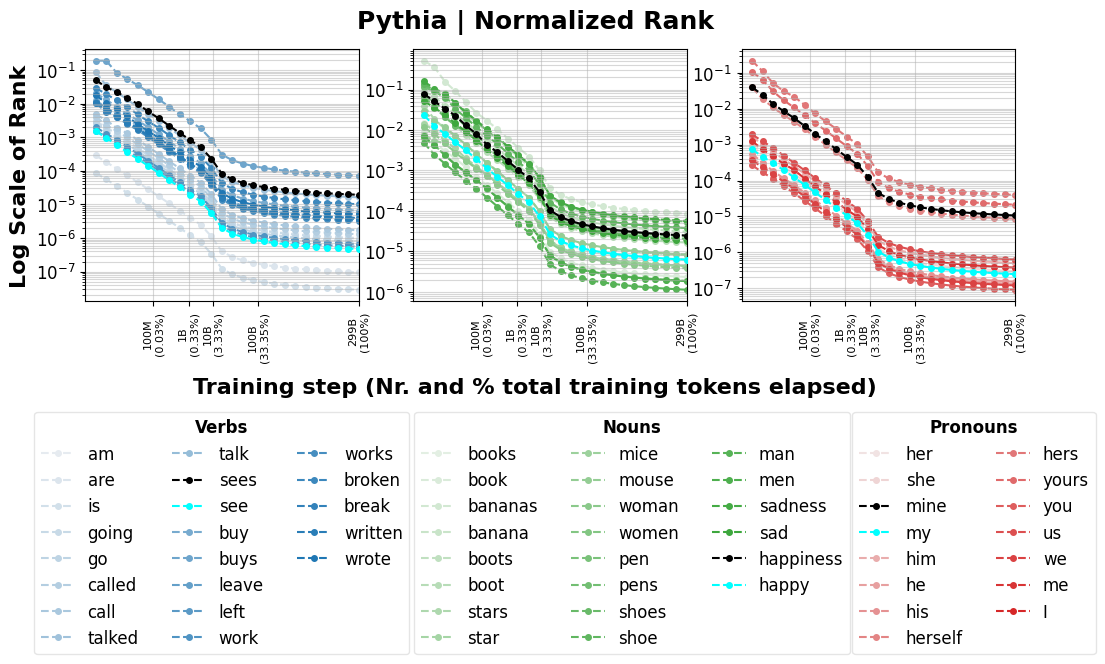}
    \caption{Rank evolution of of words in The Pile dataset over checkpoints.}
    \label{fig:appx_rank_pythia}
\end{figure}

\subsection{Tracing Semantic Stabilization}\label{sec:appx_embedd}
Figures \ref{fig:appx_embedding_pythia}-\ref{fig:appx_embedding_gpt2} represent embedding drifts for Pythia suit, GPT-2 variations, and \olmoTwo\ variations, respectively. As expected, all models follow their three-phase behavioral mentioned in Sec. \ref{sec:appx_morph}. Again, \olmoTwo-Instruct behavior is fixed across time, stating that post-training lacks a significant impact on semantic learning.

\subsection{Tracing Frequency of Words}\label{sec:appx_rank}
Figure \ref{fig:appx_rank_pythia} shows the evolution of the frequency (rank) of words in the Pile data set over training of Pythia scales, where a lower rank indicates higher popularity of words. Comparing these results with the semantic learning shown in Figs. \ref{fig:embedding_all} and \ref{fig:appx_embedding_pythia} indicates that the models stabilize the frequency of the words once the semantic learning plateaus.

\subsection{Divergence Recognition}\label{sec:appx_div}
In Sec. \ref{sec:merc_div}, we assessed the performance of Pythia suit and \olmoThree\ for the rDAT test with 16 lists, where we measured Development Ratio (DR) and Maturity Ratio (MR) for top-10 triplets. Here, repeat our experiments for top-$k$ triplets, where $k\in\{1, 5\}$. Also, we extend our evaluation to top-10 triplets of 32 lists for all Pythia suit, GPT-2 variations. \olmoTwo\ variations, and \olmoThree.

\clearpage

\begin{figure}[!th]
    \centering
    \begin{subfigure}{.305\textwidth}
          \centering
          \includegraphics[width=\linewidth]{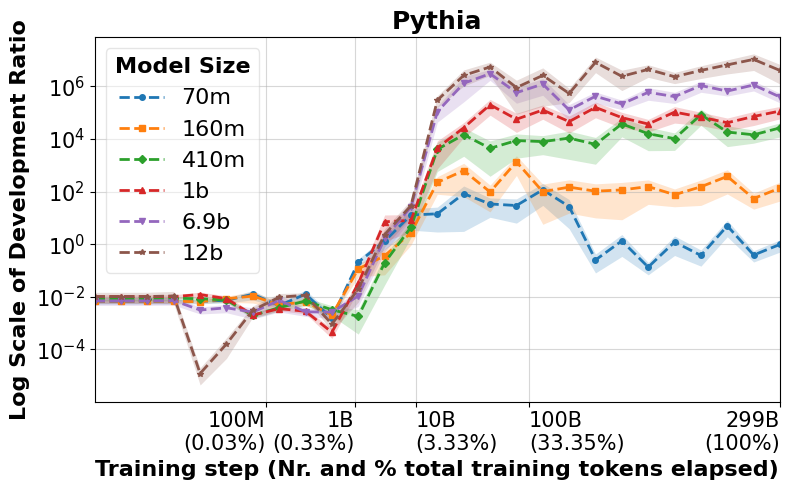}
          \caption{Pythia\textemdash Development Ratio}
          \label{fig:appx_rept_dat_creat_pythia_dr_t1}
    \end{subfigure}%
    \begin{subfigure}{.305\textwidth}
          \centering
          \includegraphics[width=\linewidth]{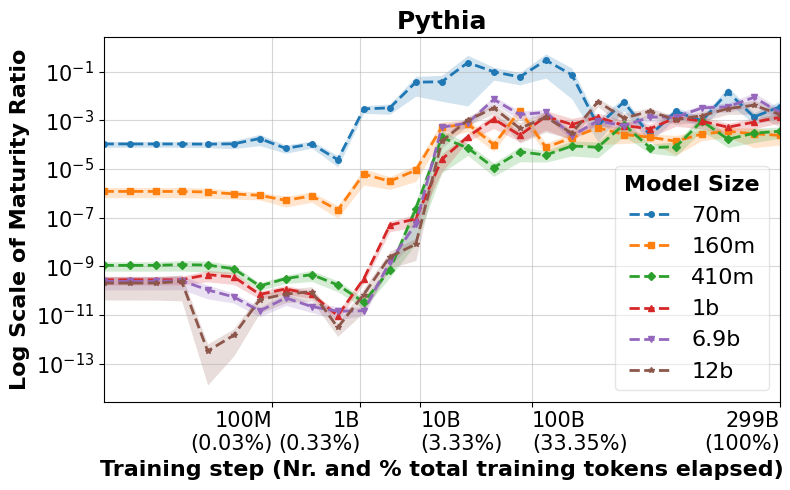}
          \caption{Pythia\textemdash Maturity Ratio}
          \label{fig:appx_rept_dat_creat_pythia_mr_t1}
    \end{subfigure}%
    \begin{subfigure}{.36\textwidth}
          \centering
          \includegraphics[width=\linewidth]{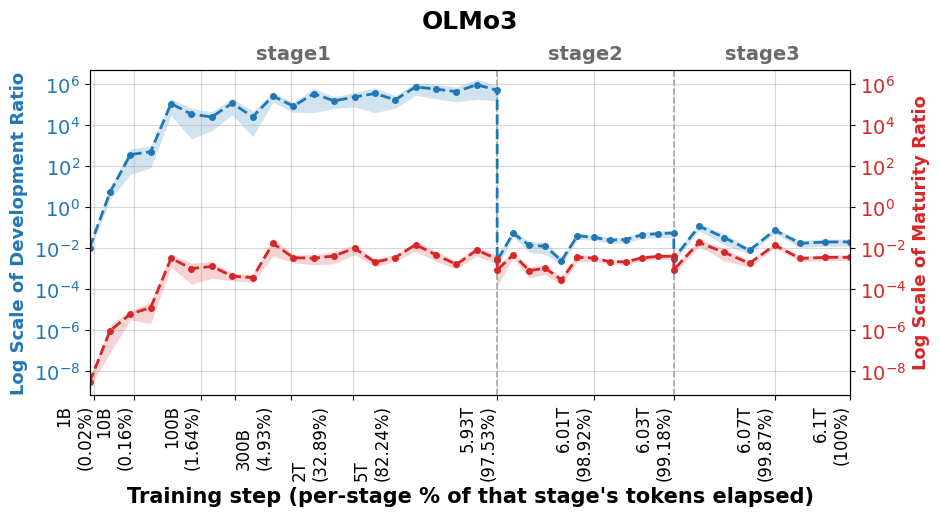}
          \caption{\olmoThree}
          \label{fig:appx_rept_dat_creat_olmo3_t1}
    \end{subfigure}\\
    \begin{subfigure}{.305\textwidth}
          \centering
          \includegraphics[width=\linewidth]{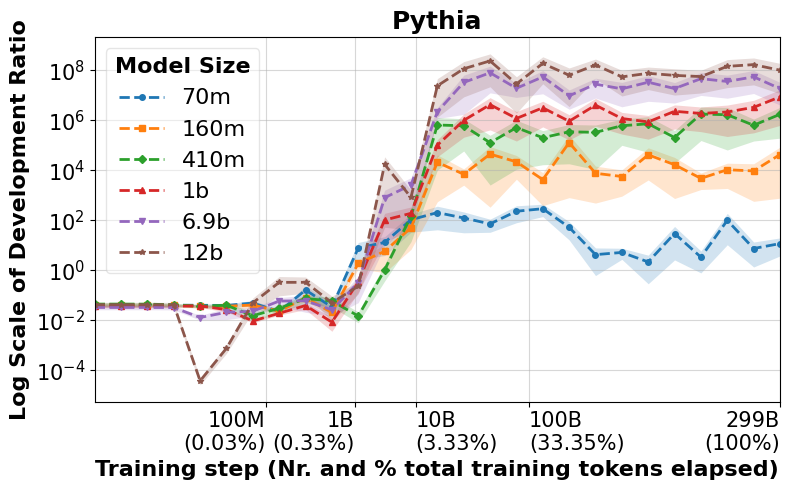}
          \caption{Pythia\textemdash Development Ratio}
          \label{fig:appx_rept_dat_creat_pythia_dr_t5}
    \end{subfigure}%
    \begin{subfigure}{.305\textwidth}
          \centering
          \includegraphics[width=\linewidth]{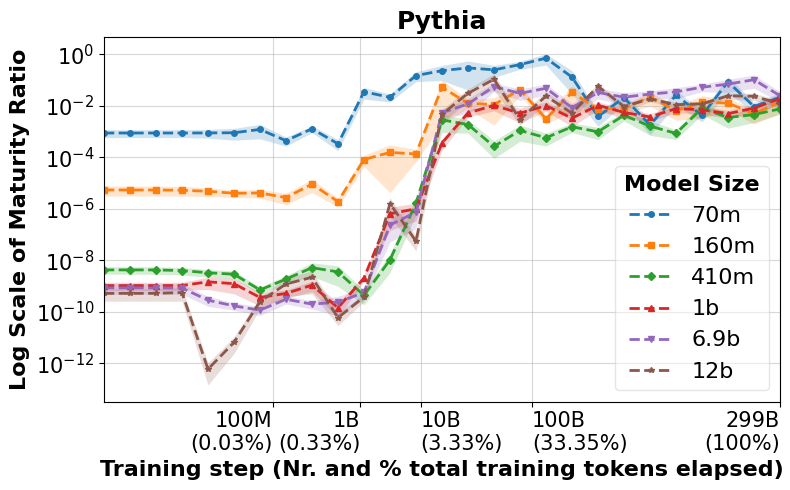}
          \caption{Pythia\textemdash Maturity Ratio}
          \label{fig:appx_rept_dat_creat_pythia_mr_t5}
    \end{subfigure}%
    \begin{subfigure}{.36\textwidth}
          \centering
          \includegraphics[width=\linewidth]{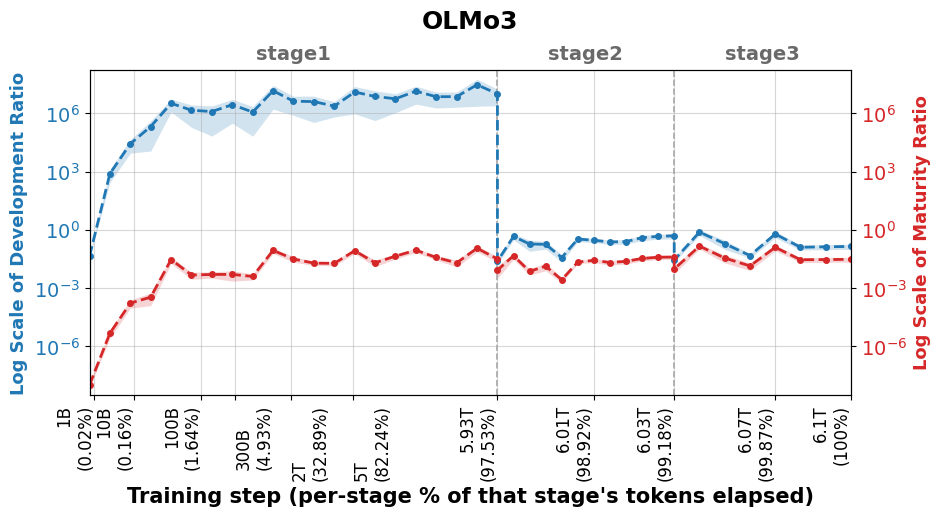}
          \caption{\olmoThree}
          \label{fig:appx_rept_dat_creat_olmo3_t5}
    \end{subfigure}%
    
    \caption{Performance of Pythia suite and \olmoThree: baseline and maturity likelihoods for 16 lists in rDAT test---top row (a-c), top-1 triplets; bottom row (d-f), top-5 triplets.}
    \label{fig:appx_rept_dat_creat}
    \vspace{-0.3cm}
\end{figure}

\begin{figure}[!th]
    \centering
    \begin{subfigure}{.45\textwidth}
          \centering
          \includegraphics[width=\linewidth]{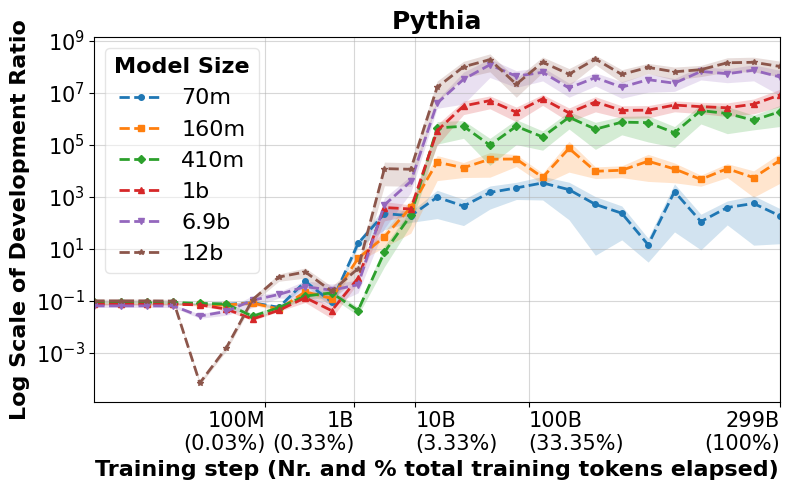}
          \caption{Pythia\textemdash Development Ratio}
          \label{fig:appx_rept_dat_creat_32_pythia_dr_t10}
    \end{subfigure}%
    \begin{subfigure}{.45\textwidth}
          \centering
          \includegraphics[width=\linewidth]{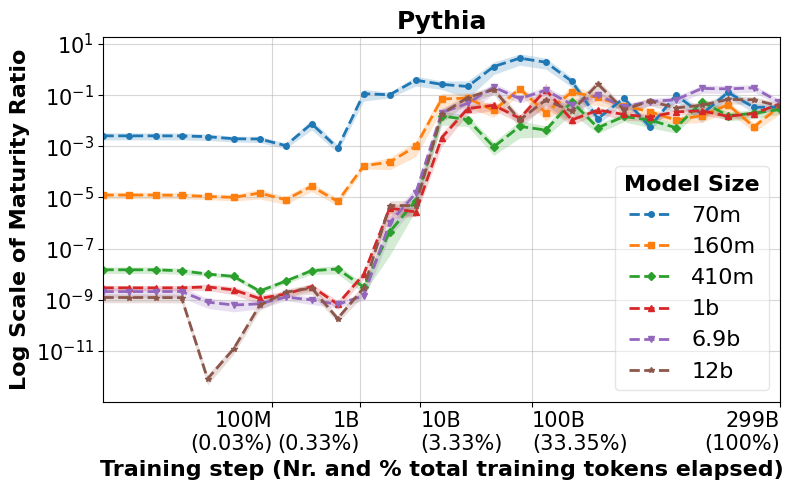}
          \caption{Pythia\textemdash Maturity Ratio}
          \label{fig:appx_rept_dat_creat_32_pythia_mr_t10}
    \end{subfigure}\\
    \begin{subfigure}{.45\textwidth}
          \centering
          \includegraphics[width=\linewidth]{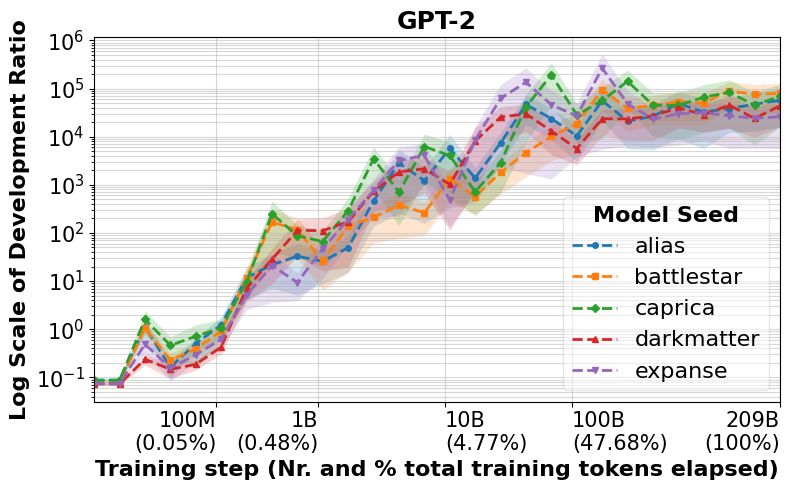}
          \caption{GPT-2\textemdash Development Ratio}
          \label{fig:appx_rept_dat_creat_32_gpt2_dr_t10}
    \end{subfigure}%
    \begin{subfigure}{.45\textwidth}
          \centering
          \includegraphics[width=\linewidth]{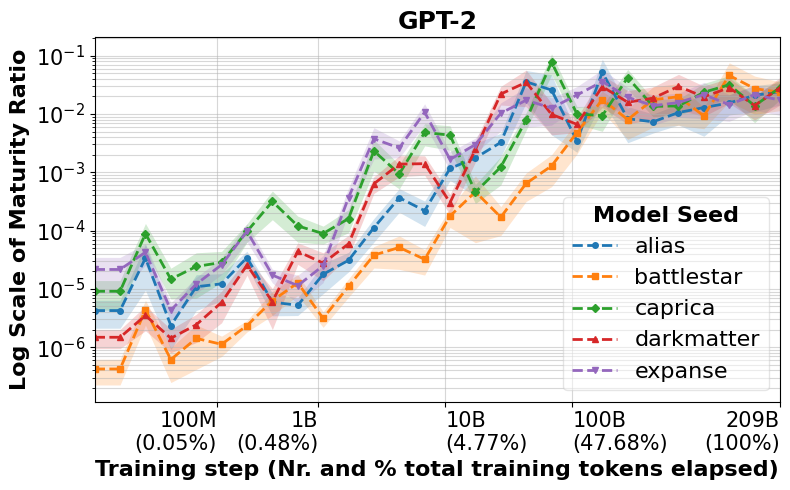}
          \caption{GPT-2\textemdash Maturity Ratio}
          \label{fig:appx_rept_dat_creat_32_gpt2_mr_t10}
    \end{subfigure}\\
    %
    \begin{subfigure}{.33\textwidth}
          \centering
          \includegraphics[width=\linewidth]{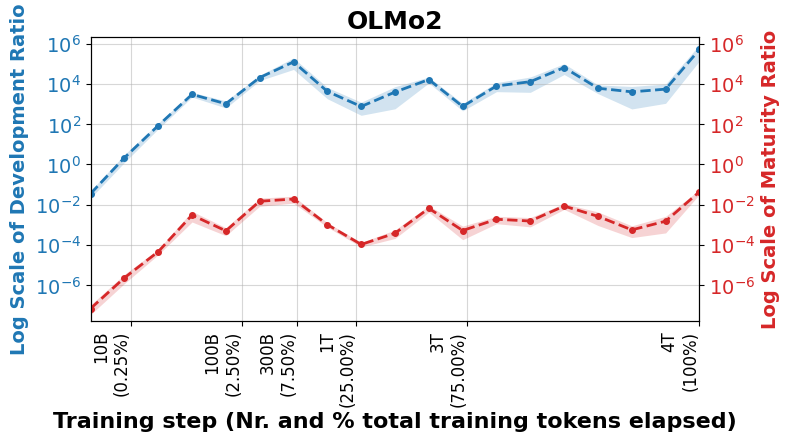}
          \caption{\olmoTwo}
          \label{fig:appx_rept_dat_creat_32_olmo2_t10}
    \end{subfigure}%
    \begin{subfigure}{.33\textwidth}
          \centering
          \includegraphics[width=\linewidth]{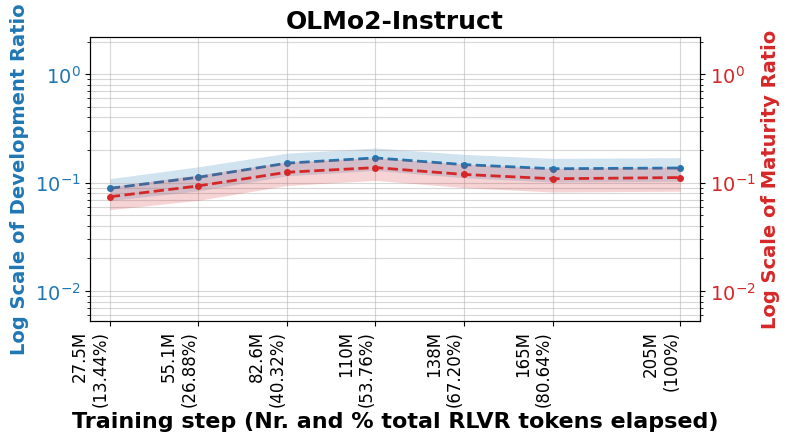}
          \caption{\olmoTwo-Instruct}
          \label{fig:appx_rept_dat_creat_32_olmo2_instruct_t10}
    \end{subfigure}%
    \begin{subfigure}{.33\textwidth}
          \centering
          \includegraphics[width=\linewidth]{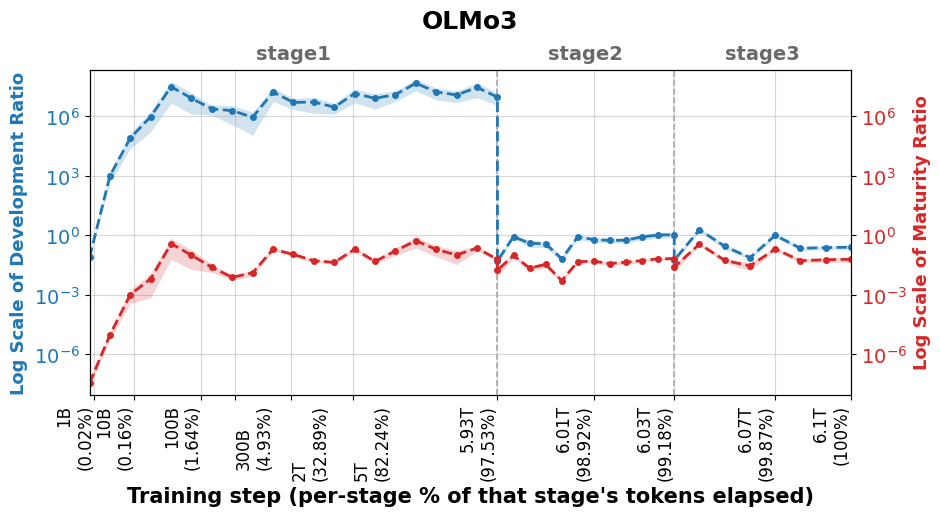}
          \caption{\olmoThree}
          \label{fig:appx_rept_dat_creat_32_olmo3_t10}
    \end{subfigure}\\
    
    \caption{Performance of Pythia suite, GPT-2 variations, \olmoTwo\ variations, and \olmoThree: baseline and maturity likelihoods for top-10 triplets of 32 lists in rDAT test.}
    \label{fig:appx_rept_dat_creat_32}
    \vspace{-0.3cm}
\end{figure}

\clearpage

The top and bottom rows in Fig. \ref{fig:appx_rept_dat_creat} show the results for top-1 and top-5 triplets, where all results follow the same trend as top-10 triplet's in Sec. \ref{sec:merc_div}. Also, the results are shown to be proportional to $k$; more $k$, the probability sum of the top triplets $k$ increases (a higher DR), and the triplets converge to their final values (a higher MR). Moreover, as discussed in Sec. \ref{sec:merc_div}, the initial DR and the final MR values are comparable to $k/\binom{10}{3}$. Figure \ref{fig:appx_rept_dat_creat_32} shows the corresponding results for 32 lists (including 16 lists in Fig. \ref{fig:appx_rept_dat_creat}). As expected, all models follow the three-phase behavior defined in Sec. \ref{sec:embedding}; models' performance improves over time and scale, except \olmoTwo-Instruct whose evolution does not vary significantly; and both the initial DR and final MR values are comparable to $10/\binom{10}{3}$.

\subsection{Association Recognition}\label{sec:appx_assoc}
\begin{figure}[H]
    \centering
    \vspace{-0.3cm}
    \begin{subfigure}{.34\textwidth}
          \centering
          \includegraphics[width=\linewidth]{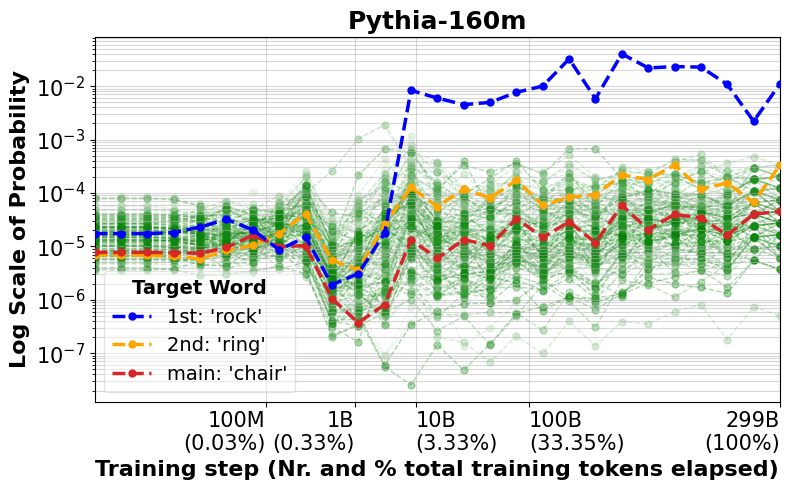}
          \label{fig:appx_rept_rat_pythia_160m_easy_chair}
    \end{subfigure}%
    \begin{subfigure}{.325\textwidth}
          \centering
          \includegraphics[width=\linewidth, trim={1cm 0 0 0}, clip]{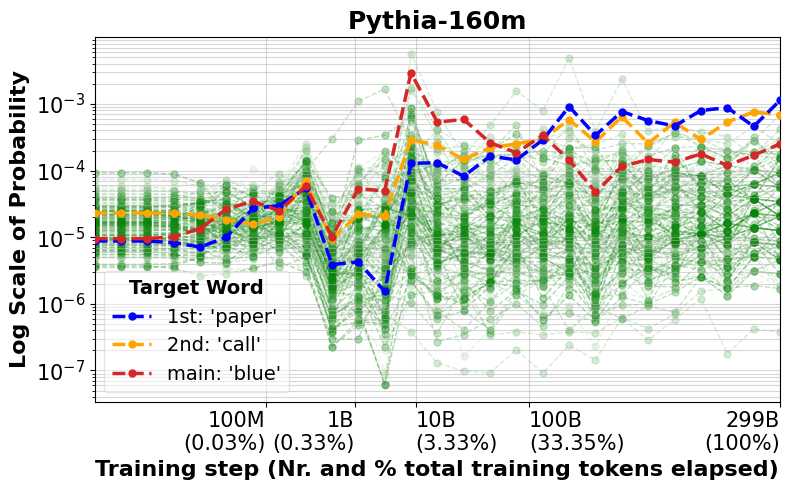}
          \label{fig:appx_rept_rat_pythia_160m_medium_blue}
    \end{subfigure}%
    \begin{subfigure}{.325\textwidth}
          \centering
          \includegraphics[width=\linewidth, trim={1cm 0 0 0}, clip]{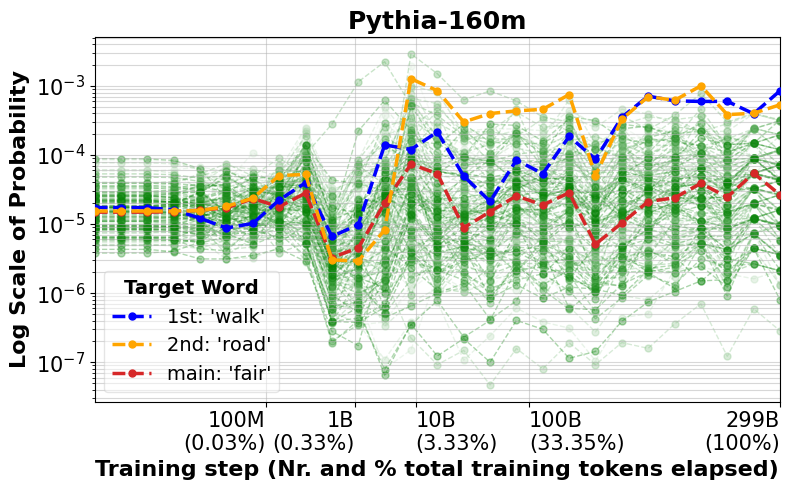}
          \label{fig:appx_rept_rat_pythia_160m_hard_fair}
    \end{subfigure}
    \\
    \vspace{-0.3cm}
    \begin{subfigure}{.34\textwidth}
          \centering
          \includegraphics[width=\linewidth]{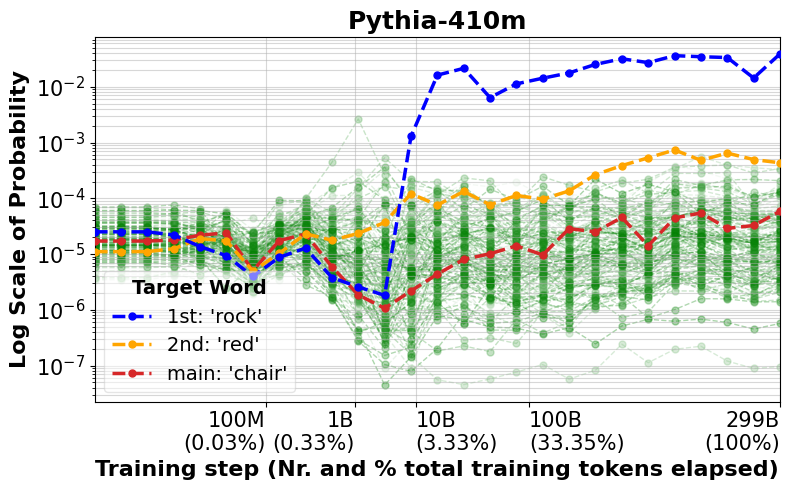}
          \label{fig:appx_rept_rat_pythia_410m_easy_chair}
    \end{subfigure}%
    \begin{subfigure}{.325\textwidth}
          \centering
          \includegraphics[width=\linewidth, trim={1cm 0 0 0}, clip]{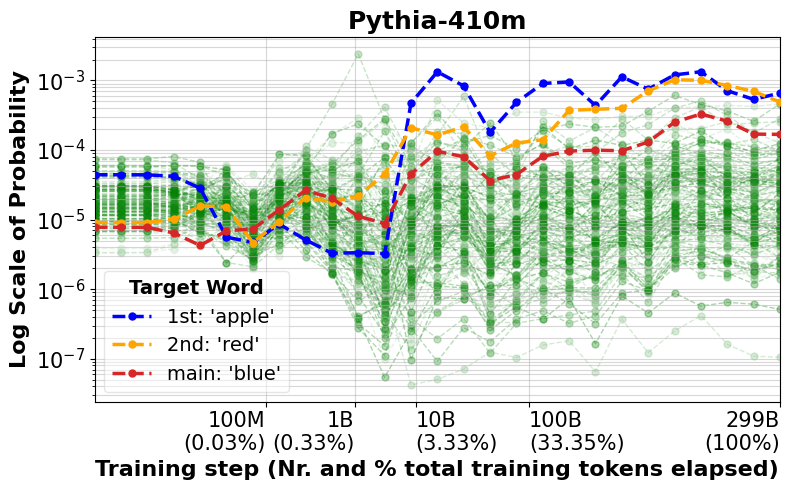}
          \label{fig:appx_rept_rat_pythia_410m_medium_blue}
    \end{subfigure}%
    \begin{subfigure}{.325\textwidth}
          \centering
          \includegraphics[width=\linewidth, trim={1cm 0 0 0}, clip]{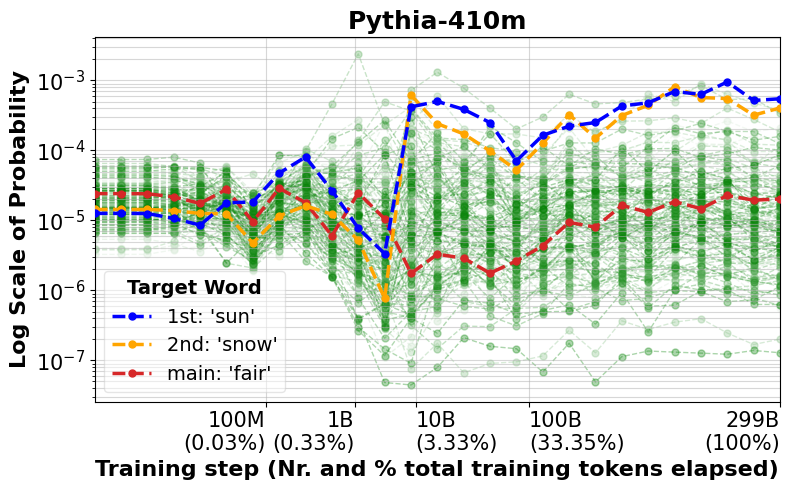}
          \label{fig:appx_rept_rat_pythia_410m_hard_fair}
    \end{subfigure}
    \\
    \vspace{-0.3cm}
    \begin{subfigure}{.34\textwidth}
          \centering
          \includegraphics[width=\linewidth, height=1in]{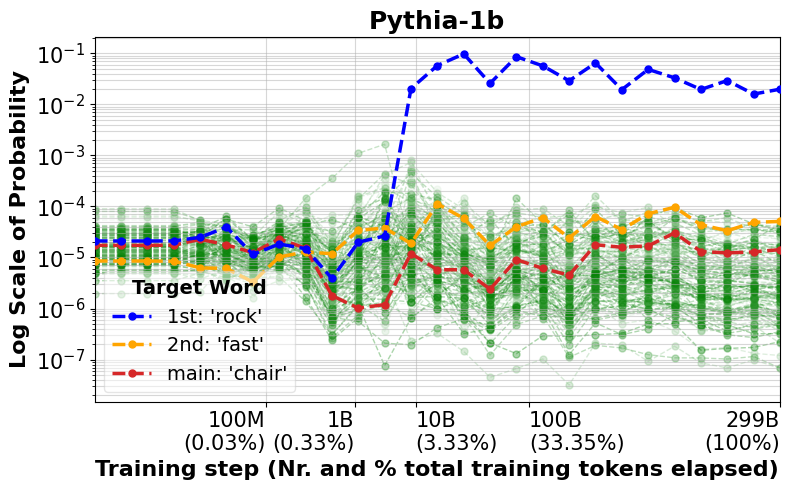}
          \label{fig:appx_rept_rat_pythia_1b_easy_chair}
    \end{subfigure}%
    \begin{subfigure}{.325\textwidth}
          \centering
          \includegraphics[width=\linewidth, height=1in, trim={1cm 0 0 0}, clip]{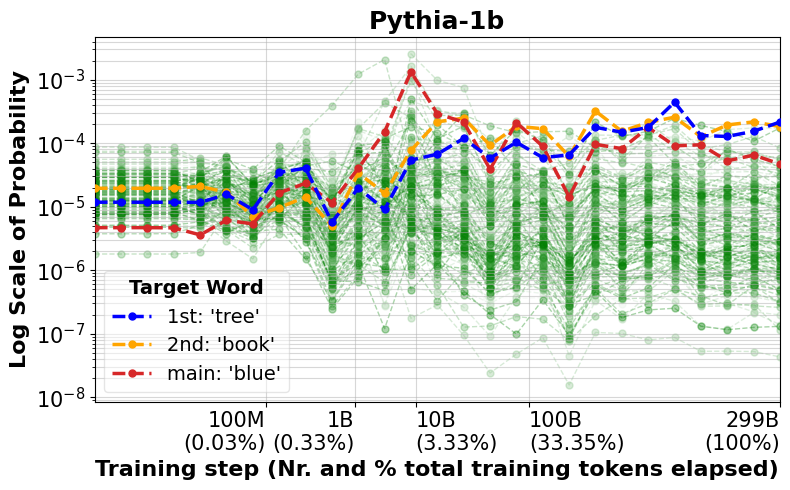}
          \label{fig:appx_rept_rat_pythia_1b_medium_blue}
    \end{subfigure}%
    \begin{subfigure}{.325\textwidth}
          \centering
          \includegraphics[width=\linewidth, height=1in, trim={1cm 0 0 0}, clip]{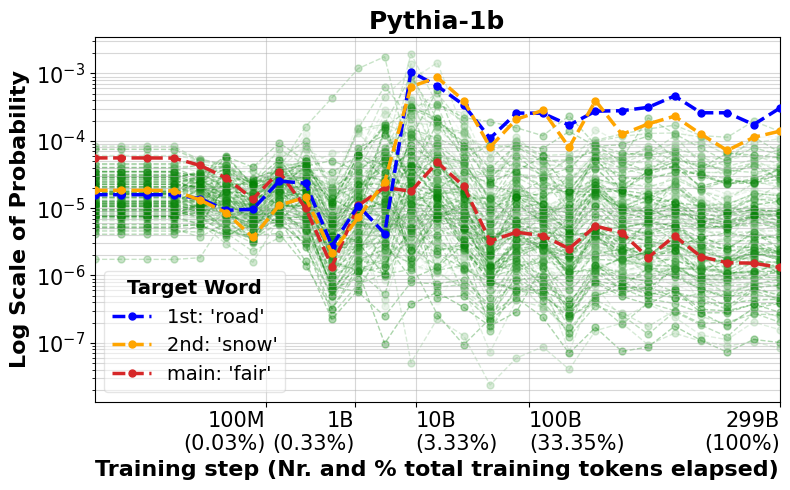}
          \label{fig:appx_rept_rat_pythia_1b_hard_fair}
    \end{subfigure}
    \\
    \vspace{-0.3cm}
    \begin{subfigure}{.34\textwidth}
          \centering
          \includegraphics[width=\linewidth, height=1in]{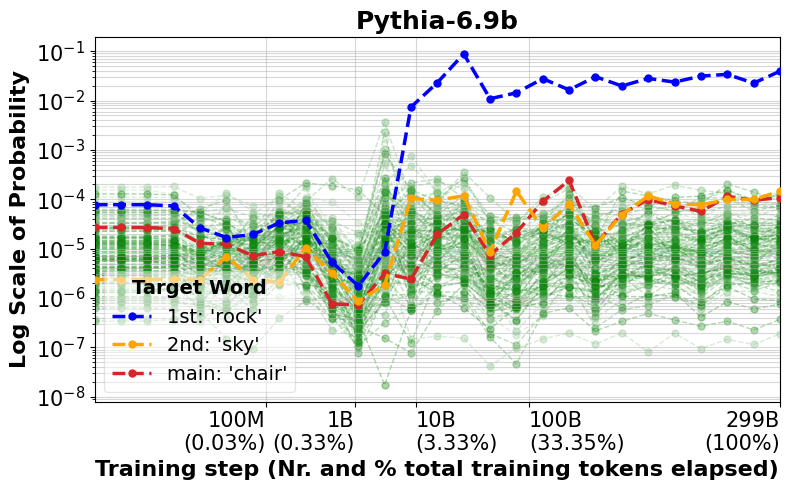}
          \caption{``rocking, wheel, high'' $\rightarrow$ ``chair''}
          \label{fig:appx_rept_rat_pythia_6.9b_easy_chair}
    \end{subfigure}%
    \begin{subfigure}{.325\textwidth}
          \centering
          \includegraphics[width=\linewidth, height=1in, trim={1cm 0 0 0}, clip]{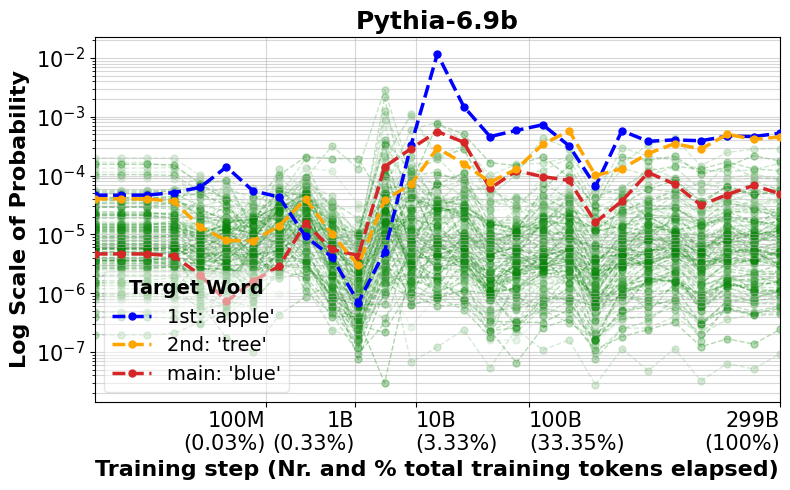}
          \caption{``print, berry, bird'' $\rightarrow$ ``blue''}
          \label{fig:appx_rept_rat_pythia_6.9b_medium_blue}
    \end{subfigure}%
    \begin{subfigure}{.325\textwidth}
          \centering
          \includegraphics[width=\linewidth, height=1in, trim={1cm 0 0 0}, clip]{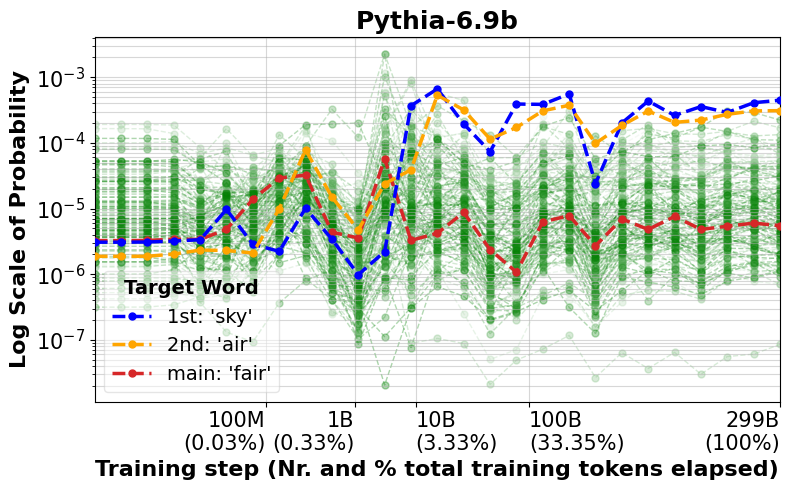}
          \caption{``way, ground, weather'' $\rightarrow$ ``fair''}
          \label{fig:appx_rept_rat_pythia_6.9b_hard_fair}
    \end{subfigure}
    
    \caption{RAT results across time for Pythia suit: 160M, 410M, 1B, and 6.9B---first to fourth rows, respectively; (a) Easy; (b) Medium; and (c) Hard cases.}
    \label{fig:appx_rept_rat_pythia}
    \vspace{-0.5cm}
\end{figure}

\begin{figure}[H]
    \centering
    \vspace{-0.3cm}
    \begin{subfigure}{.34\textwidth}
          \centering
          \includegraphics[width=\linewidth]{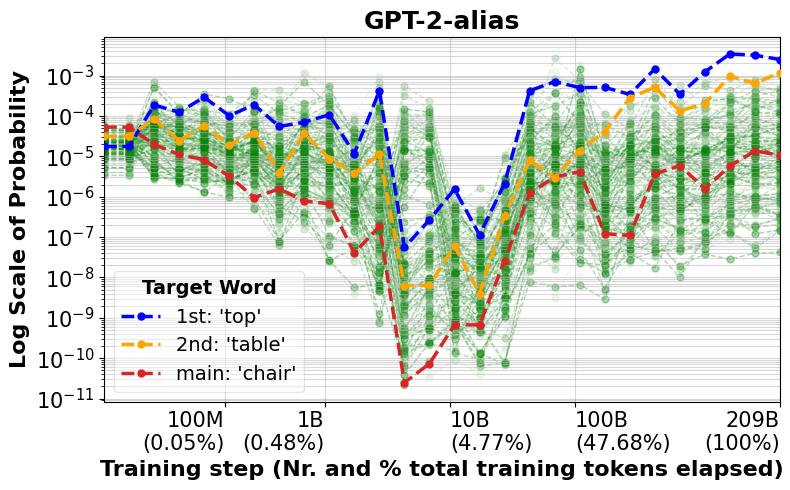}
          \label{fig:appx_rept_rat_gpt2_alias_easy_chair}
    \end{subfigure}%
    \begin{subfigure}{.325\textwidth}
          \centering
          \includegraphics[width=\linewidth, trim={1cm 0 0 0}, clip]{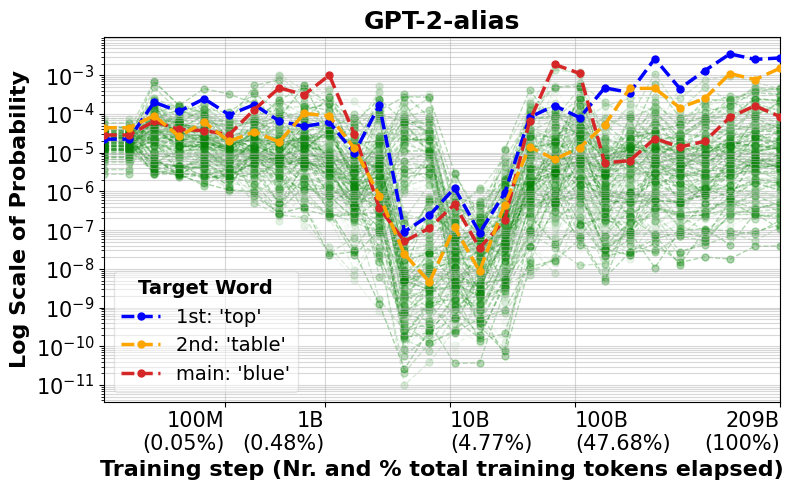}
          \label{fig:appx_rept_rat_gpt2_alias_medium_blue}
    \end{subfigure}%
    \begin{subfigure}{.325\textwidth}
          \centering
          \includegraphics[width=\linewidth, trim={1cm 0 0 0}, clip]{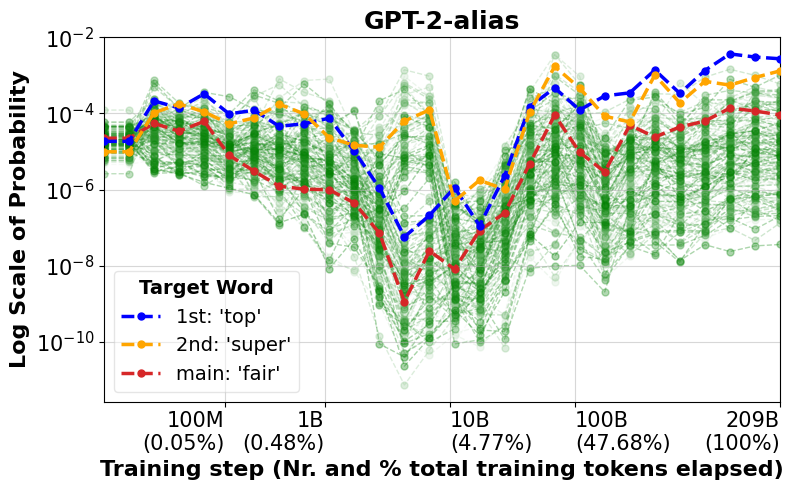}
          \label{fig:appx_rept_rat_gpt2_alias_hard_fair}
    \end{subfigure}
    \\
    \vspace{-0.3cm}
    \begin{subfigure}{.34\textwidth}
          \centering
          \includegraphics[width=\linewidth]{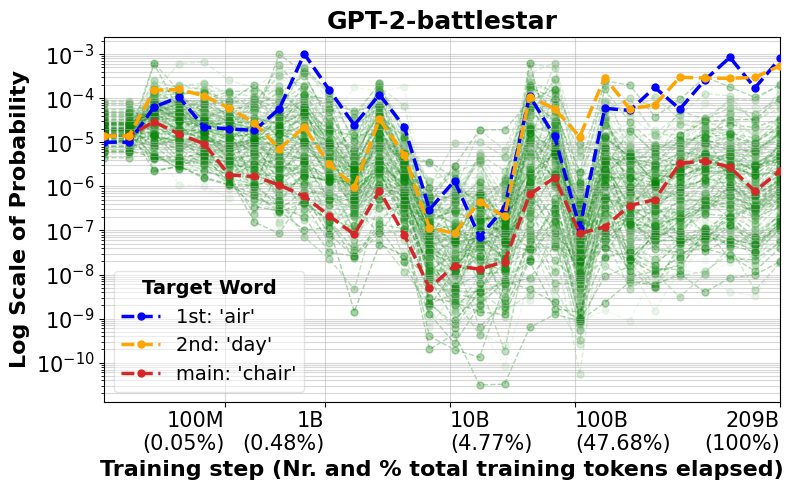}
          \label{fig:appx_rept_rat_gpt2_battlestar_easy_chair}
    \end{subfigure}%
    \begin{subfigure}{.325\textwidth}
          \centering
          \includegraphics[width=\linewidth, trim={1cm 0 0 0}, clip]{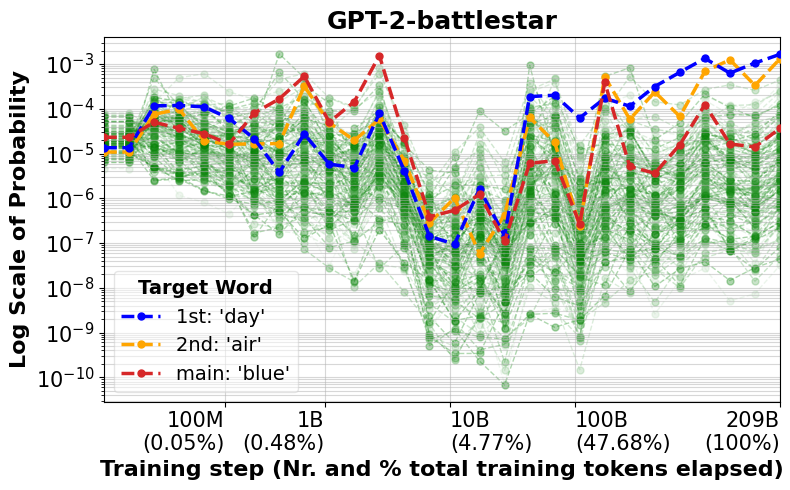}
          \label{fig:appx_rept_rat_gpt2_battlestar_medium_blue}
    \end{subfigure}%
    \begin{subfigure}{.325\textwidth}
          \centering
          \includegraphics[width=\linewidth, trim={1cm 0 0 0}, clip]{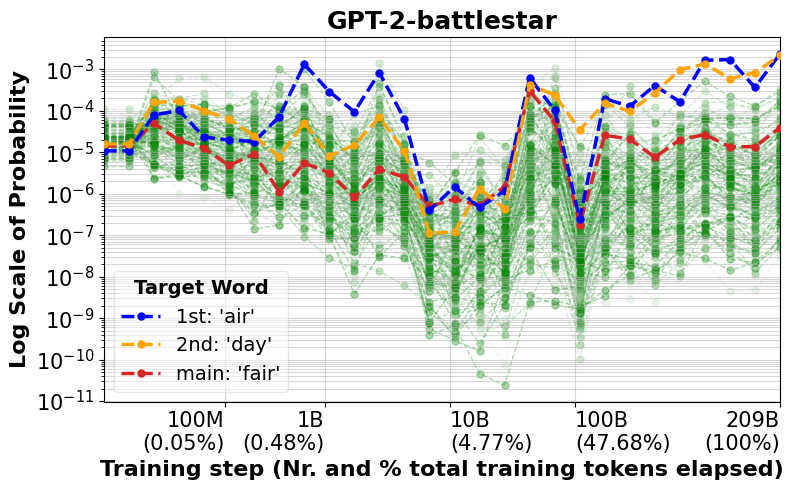}
          \label{fig:appx_rept_rat_gpt2_battlestar_hard_fair}
    \end{subfigure}
    \\
    \vspace{-0.3cm}
    \begin{subfigure}{.34\textwidth}
          \centering
          \includegraphics[width=\linewidth, height=1in]{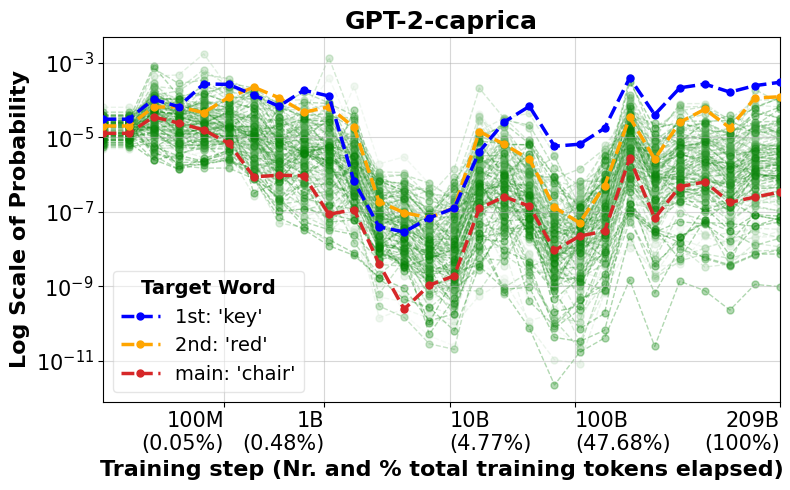}
          \label{fig:appx_rept_rat_gpt2_caprica_easy_chair}
    \end{subfigure}%
    \begin{subfigure}{.325\textwidth}
          \centering
          \includegraphics[width=\linewidth, height=1in, trim={1cm 0 0 0}, clip]{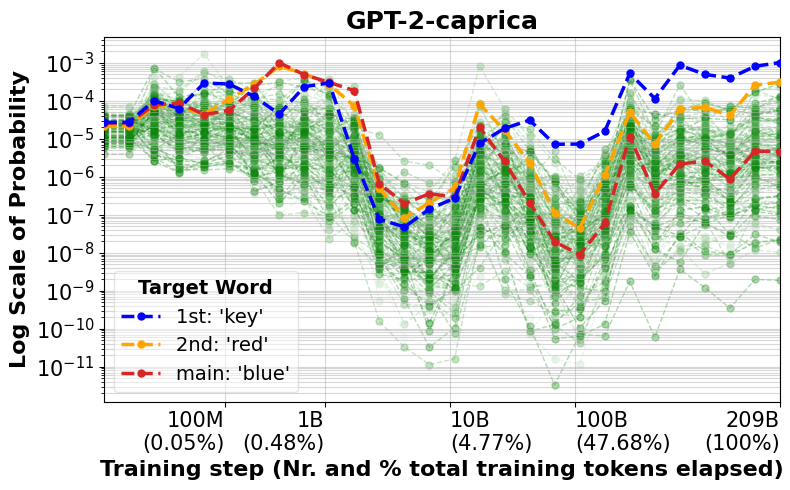}
          \label{fig:appx_rept_rat_gpt2_caprica_medium_blue}
    \end{subfigure}%
    \begin{subfigure}{.325\textwidth}
          \centering
          \includegraphics[width=\linewidth, height=1in, trim={1cm 0 0 0}, clip]{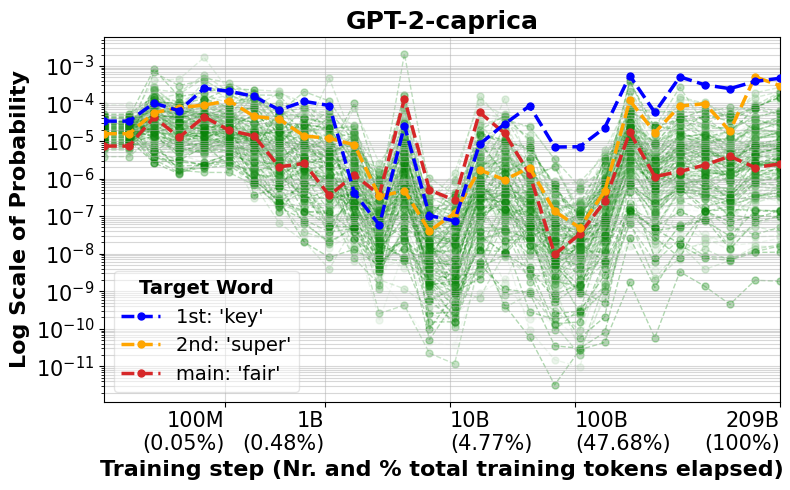}
          \label{fig:appx_rept_rat_gpt2_caprica_hard_fair}
    \end{subfigure}
    \\
    \vspace{-0.3cm}
    \begin{subfigure}{.34\textwidth}
          \centering
          \includegraphics[width=\linewidth, height=1in]{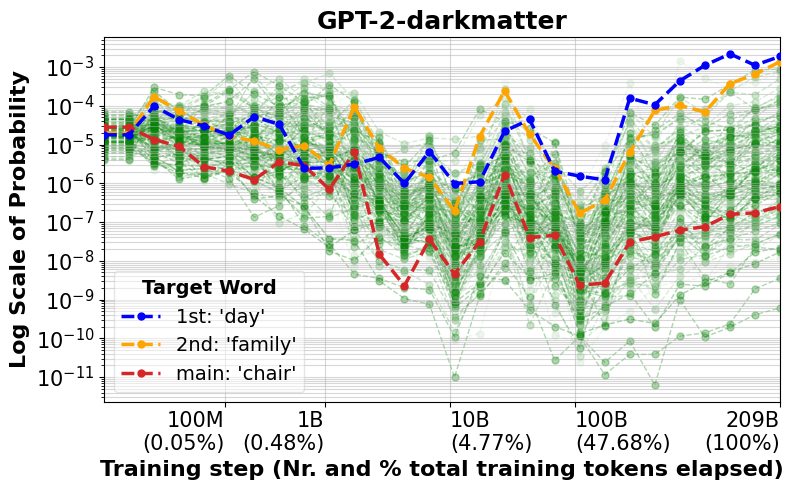}
          \label{fig:appx_rept_rat_gpt2_darkmatter_easy_chair}
    \end{subfigure}%
    \begin{subfigure}{.325\textwidth}
          \centering
          \includegraphics[width=\linewidth, height=1in, trim={1cm 0 0 0}, clip]{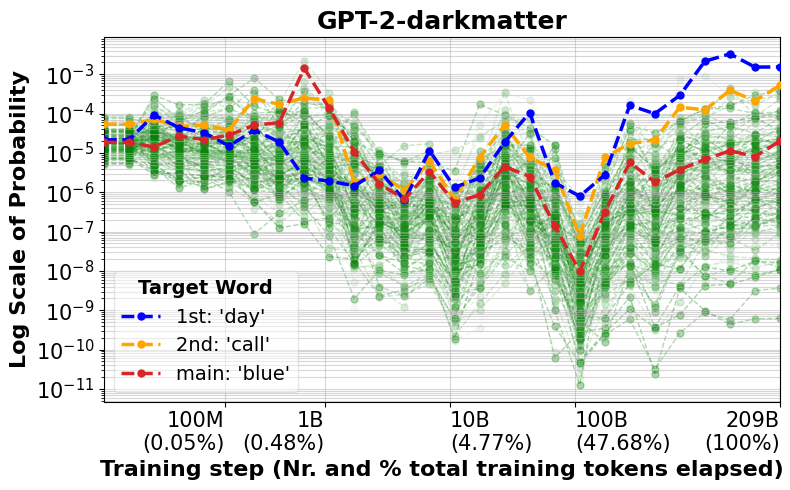}
          \label{fig:appx_rept_rat_gpt2_darkmatter_medium_blue}
    \end{subfigure}%
    \begin{subfigure}{.325\textwidth}
          \centering
          \includegraphics[width=\linewidth, height=1in, trim={1cm 0 0 0}, clip]{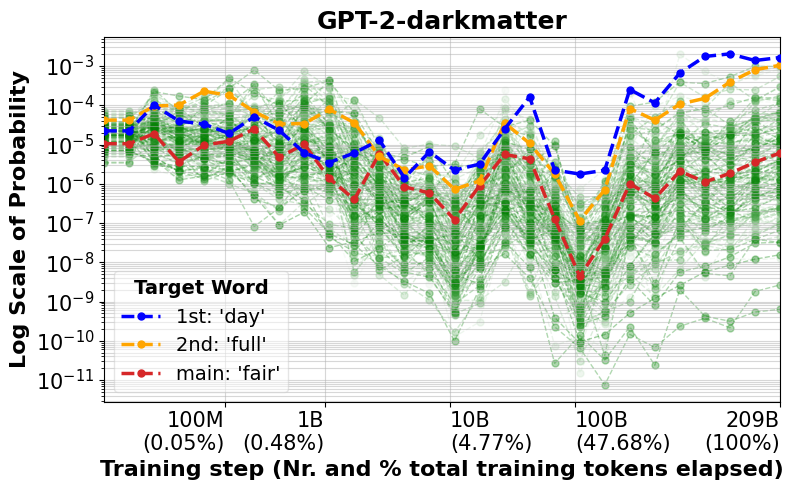}
          \label{fig:appx_rept_rat_gpt2_darkmatter_hard_fair}
    \end{subfigure}
    \\
    \vspace{-0.3cm}
    \begin{subfigure}{.34\textwidth}
          \centering
          \includegraphics[width=\linewidth, height=1in]{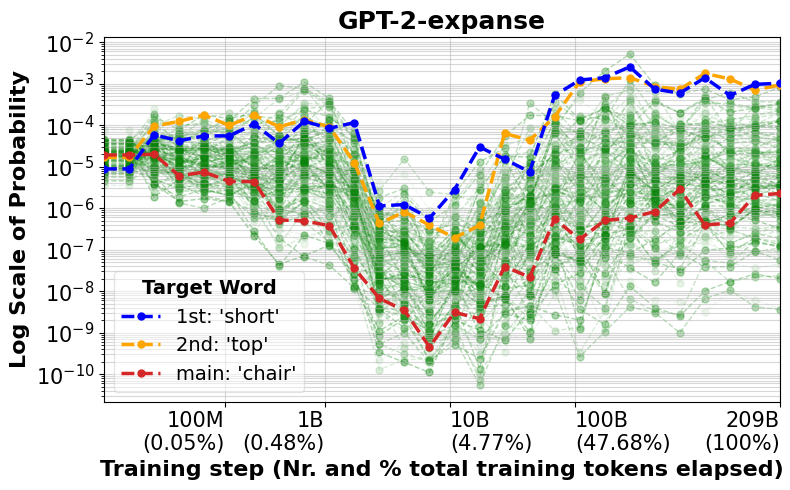}
          \caption{``rocking, wheel, high'' $\rightarrow$ ``chair''}
          \label{fig:appx_rept_rat_gpt2_expanse_easy_chair}
    \end{subfigure}%
    \begin{subfigure}{.325\textwidth}
          \centering
          \includegraphics[width=\linewidth, height=1in, trim={1cm 0 0 0}, clip]{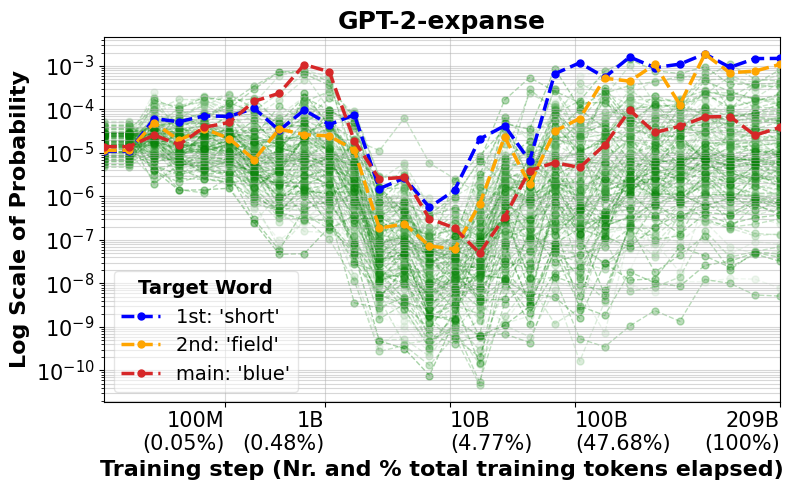}
          \caption{``print, berry, bird'' $\rightarrow$ ``blue''}
          \label{fig:appx_rept_rat_gpt2_expanse_medium_blue}
    \end{subfigure}%
    \begin{subfigure}{.325\textwidth}
          \centering
          \includegraphics[width=\linewidth, height=1in, trim={1cm 0 0 0}, clip]{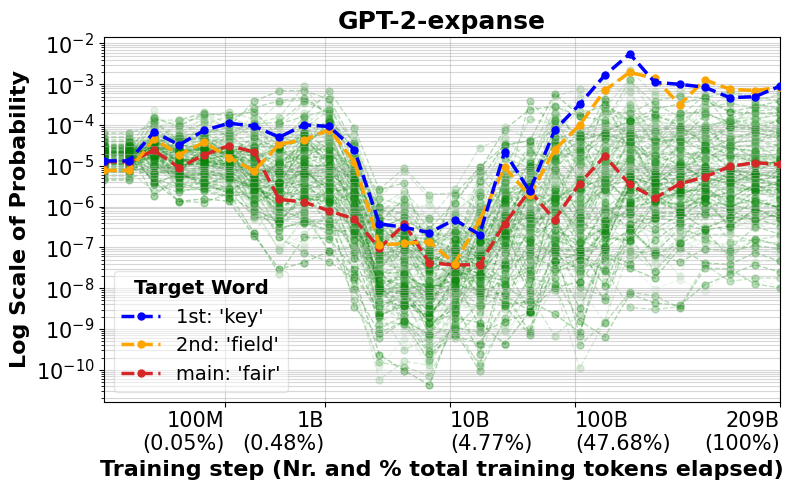}
          \caption{``way, ground, weather'' $\rightarrow$ ``fair''}
          \label{fig:appx_rept_rat_gpt2_expanse_hard_fair}
    \end{subfigure}
    
    \caption{RAT results across time for GPT-2 seeds: Alias, Battlestar, Caprica, Darkmatter, and Expanse---first to fifth rows, respectively; (a) Easy; (b) Medium; and (c) Hard cases.}
    \label{fig:appx_rept_gpt2}
    \vspace{-0.5cm}
\end{figure}

\begin{figure}[H]
    \centering
    \vspace{-0.3cm}
    \begin{subfigure}{.34\textwidth}
          \centering
          \includegraphics[width=\linewidth]{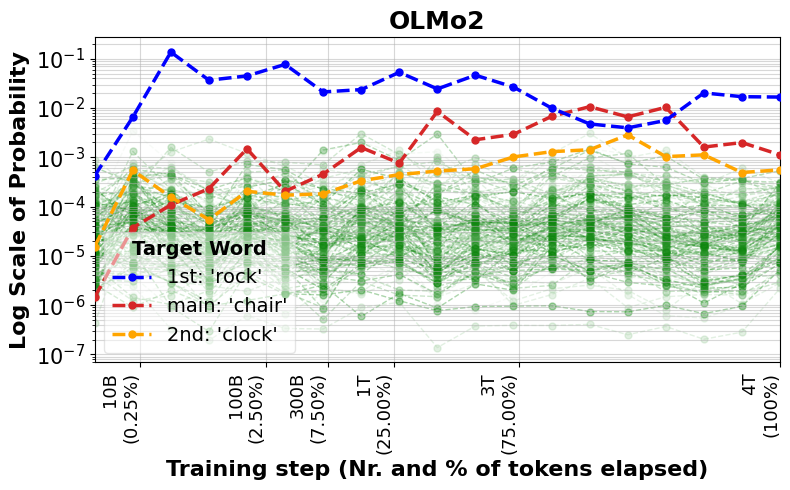}
          \label{fig:appx_rept_rat_olmo2_7B_easy_chair}
    \end{subfigure}%
    \begin{subfigure}{.325\textwidth}
          \centering
          \includegraphics[width=\linewidth, trim={1cm 0 0 0}, clip]{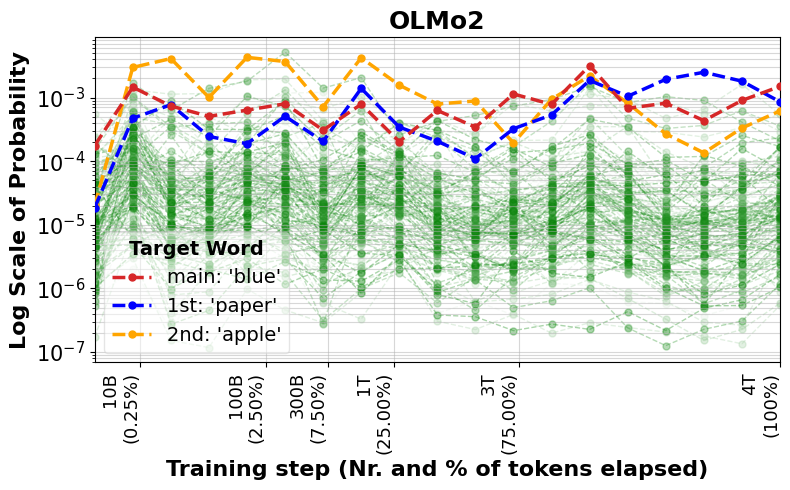}
          \label{fig:appx_rept_rat_olmo2_7B_medium_blue}
    \end{subfigure}%
    \begin{subfigure}{.325\textwidth}
          \centering
          \includegraphics[width=\linewidth, trim={1cm 0 0 0}, clip]{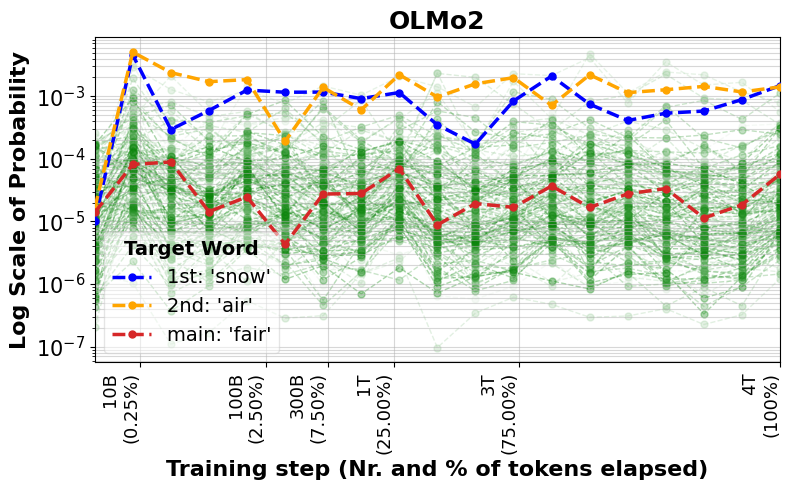}
          \label{fig:appx_rept_rat_olmo2_7B_hard_fair}
    \end{subfigure}
    \\
    \vspace{-0.3cm}
    \begin{subfigure}{.34\textwidth}
          \centering
          \includegraphics[width=\linewidth, height=1in]{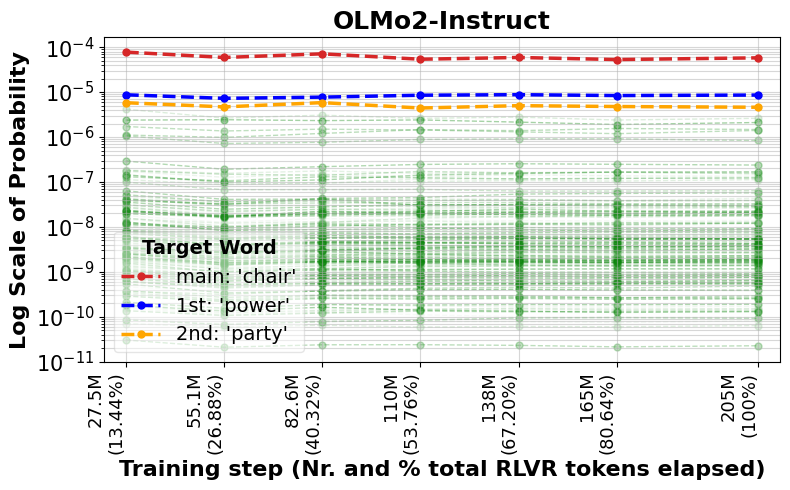}
          \caption{``rocking, wheel, high'' $\rightarrow$ ``chair''}
          \label{fig:appx_rept_rat_olmo2_7B_instruct_easy_chair}
    \end{subfigure}%
    \begin{subfigure}{.325\textwidth}
          \centering
          \includegraphics[width=\linewidth, height=1in, trim={1cm 0 0 0}, clip]{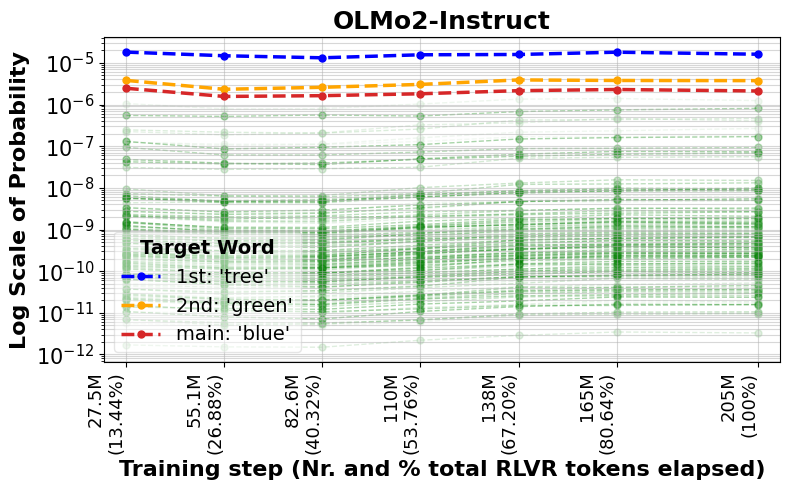}
          \caption{``print, berry, bird'' $\rightarrow$ ``blue''}
          \label{fig:appx_rept_rat_olmo2_7B_instruct_medium_blue}
    \end{subfigure}%
    \begin{subfigure}{.325\textwidth}
          \centering
          \includegraphics[width=\linewidth, height=1in, trim={1cm 0 0 0}, clip]{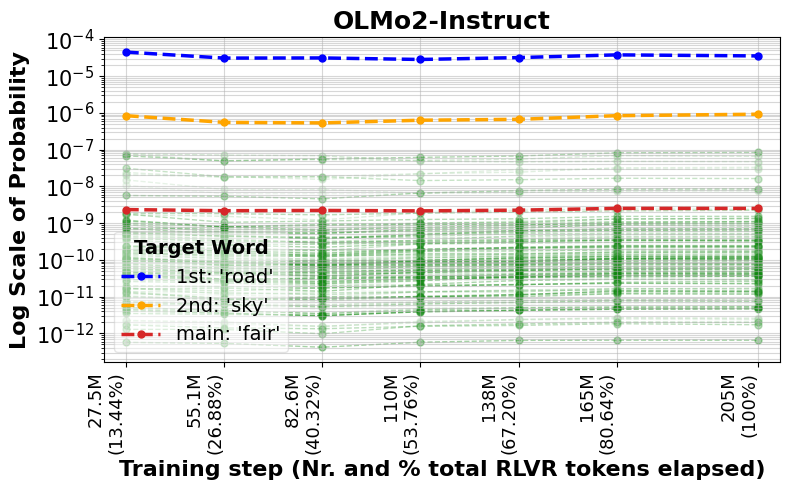}
          \caption{``way, ground, weather'' $\rightarrow$ ``fair''}
          \label{fig:appx_rept_rat_olmo2_7B_instruct_hard_fair}
    \end{subfigure}
    
    \caption{RAT results across time for \olmoTwo\ (top row) and \olmoTwo-Instruct (bottom row): (a) Easy; (b) Medium; and (c) Hard cases.}
    \label{fig:appx_rept_rat_olmo2}
    \vspace{-0.5cm}
\end{figure}

Following experiments in Sec. \ref{sec:merc_assoc}, we evaluate models' performance for the RAT test w.r.t. the correct answer (labeled ``main'' and colored ``red'') and the two targets with the highest probabilities at the last checkpoint in Figs. \ref{fig:appx_rept_rat_pythia}-\ref{fig:appx_rept_rat_olmo2} for Pythia suite, GPT-2 seeds, an \olmoTwo\ suite, respectively. Compared to Fig. \ref{fig:rept_rat}, \olmoTwo\ variations can do better than other models for Easy and Medium cases. According to Fig. \ref{fig:appx_rept_rat_olmo2_7B_instruct_easy_chair}, \olmoTwo\ increases the probability of the correct answer, ``chair'', over training time. However, ``rock'' owns the highest probability among all target words, emphasizing the impact of repetition on the model's performance. Also, Fig. \ref{fig:appx_rept_rat_olmo2_7B_instruct_medium_blue} indicates that \olmoTwo\ increases the probability of ``blue'' over ``tree'' most of the times, particularly at the last checkpoint. On the other hand, according to Fig. \ref{fig:appx_rept_rat_olmo2_7B_instruct_easy_chair}, \olmoTwo-Instruct gives a higher probability to ``chair'', showing the influence of instruction on the model's performance. However, our experiments on the other unit-test cases reveal that this behavior is not consistent with \textit{all} unit-test cases in RAT, yet for \textit{most} cases; not only Easy cases but also Medium ones.


\clearpage


\newpage

\end{document}